\documentclass{article}

\usepackage{multirow}
\usepackage{geometry}
\usepackage{cite,hyperref}
\usepackage{amsmath,amssymb,amsfonts,bbm,amsthm}
\usepackage{algorithm}
\usepackage{MnSymbol}
\usepackage{graphicx,tikz}
\usepackage{enumitem}
\usepackage{flushend}
\usepackage{textcomp,multicol}
\usepackage{xcolor,url,comment}
\usepackage{stfloats,comment,lscape}
\usepackage{float,flushend}
\usepackage{graphics,subcaption,nameref,tikz}

\usepackage{algpseudocode}

\let\oldReturn\Return
\renewcommand{\Return}{\State\oldReturn}

\usepackage[figuresright]{rotating}

\newcommand\btab{\begin{tabular}}
\newcommand\etab{\end{tabular}}
\newcommand\bfig{\begin{figure}\centering}
\newcommand\efig{\end{figure}}
\newcommand\bfigs{\begin{figure*}\centering}
\newcommand\efigs{\end{figure*}}

\newcommand\Idle{\mathrm{Idle}}
\newcommand\idle{\mathrm{idle}}

\usepackage{authblk}

\title{A Survey on Task Allocation and Scheduling in Robotic Network Systems}
\author[1]{Saeid Alirezazadeh\thanks{The work was done when Saeid Alirezazadeh was with C4-Cloud Computing Competence Center, Universidade da Beira Interior, C4-Estrada Municipal, 506, 6200-284, Covilh\~{a}, Portugal. This work has been accepted for publication in the IEEE Internet of Things Journal \url{https://ieeexplore.ieee.org/document/10742644} with DOI: 10.1109/JIOT.2024.3491944. IEEE allows its authors to follow mandates of agencies that fund the author's research by posting the peer-reviewed accepted manuscript versions of their articles in the agencies' publicly accessible repositories. No third party (other than authors and employers) may post IEEE- copyrighted material without obtaining the necessary licenses or permissions from the IEEE Intellectual Property Rights Office or other authorized representatives of the IEEE.}}
\author[2]{Lu\'{i}s A. Alexandre}
\affil[1]{Computer Science and Communication Research Centre-(CIIC), Escola Superior de Tecnologia e Gest\~{a}o, Instituto Polit\'{e}cnico de Leiria, Portugal, Building C – Campus 2, Morro do Lena – Alto do Vieiro, 2411-901, Leiria, Portugal.}
\affil[2]{NOVA LINCS, Universidade da Beira Interior, Covilh\~{a}, Portugal.}
\affil[1]{Email id: saeid.alirezazadeh@gmail.com}
\affil[2]{Email id: lfbaa@di.ubi.pt}
\date{}

\providecommand{\keywords}[1]{\textbf{\textit{Keywords---}} #1}
\begin{document}

\maketitle

\begin{abstract}
Robotic networks are increasingly relied upon to perform complex tasks that require efficient scheduling and task allocation to optimize processing power, resource management, and energy use. The primary goal in these systems is to enhance performance by minimizing completion time, energy consumption, and delays, while maximizing resource utilization and task throughput. Numerous studies have examined different aspects of task allocation and scheduling, from static approaches to dynamic models that adapt to real-time conditions. This paper presents a comprehensive survey of the methods and strategies used in robotic network systems, considering not only traditional approaches but also the role of emerging technologies such as cloud, fog, and edge computing. We categorize the literature from three perspectives: Architectures and Applications, Methods, and Parameters. Furthermore, we analyze the limitations of each approach and propose directions for future research, with a particular focus on scalability, real-world applicability, and the integration of these technologies in dynamic environments.
\end{abstract}


%

\section{Introduction}
Robots are increasingly taking on dangerous tasks and acting as companions that are integrated into various aspects of human life \cite{Chatterjee:2021, Amritha:2022}. Some tasks like carrying a very heavy object, monitoring a wide area or dealing with a disaster cannot be performed by a single robot. To overcome this limitation, several studies, such as \cite{osumi:2014, michael:2012, Hu:2012, Mckee:2000, McKee:2008, Tenorth:2013, Kamei:2012}, propose that multiple robots can be used to perform tasks cooperatively, in a configuration called a robotic network. The capacity of a robotic network is limited by the collective capacity of all robots, \cite{Hu:2012}. Increasing the number of robots can increase the capacity, but it also increases the complexity of the model. Moreover, most tasks related to human-robot collaboration, such as speech, face, and object recognition, are very computationally intensive.

\begin{figure}[th!]
\begin{minipage}{.48\textwidth}
\centering
\includegraphics[width=1\linewidth]{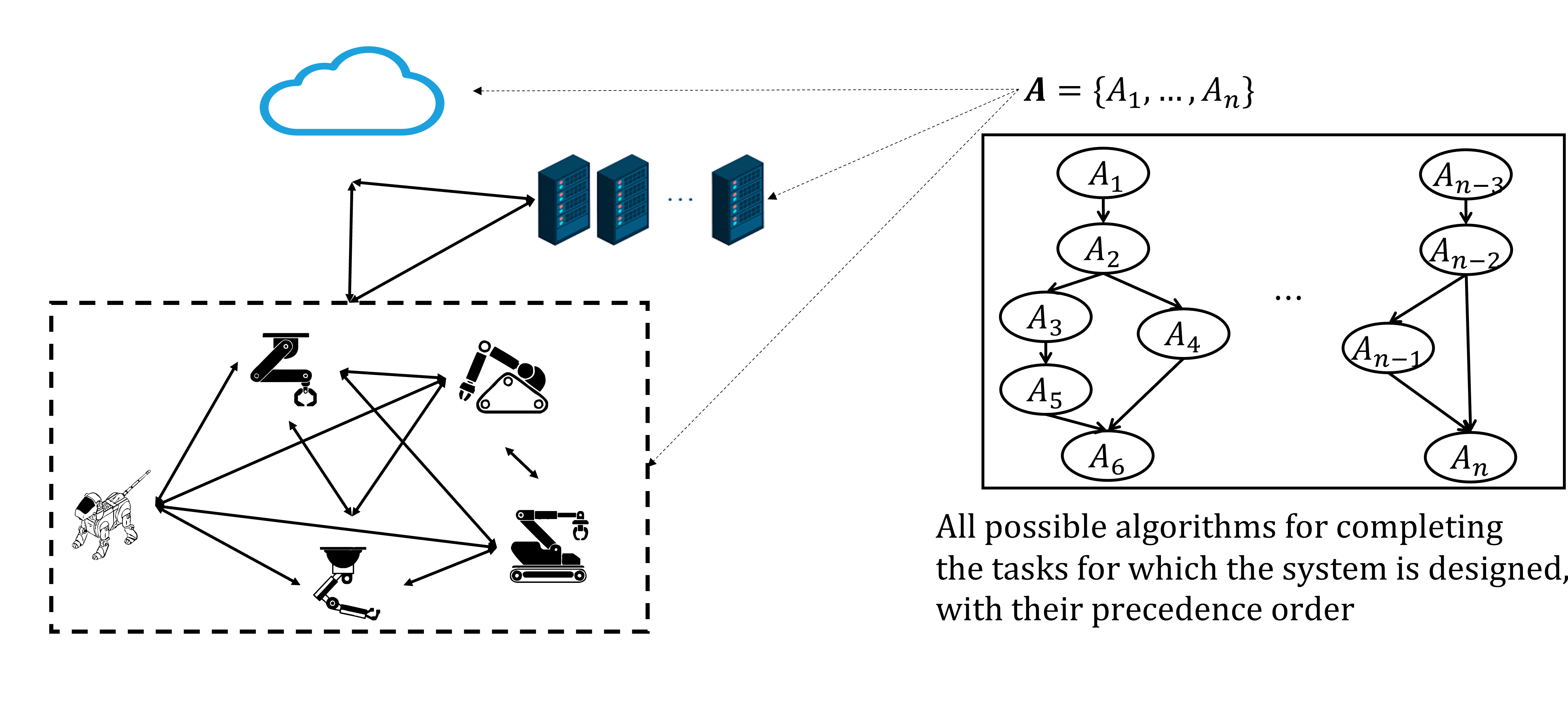}
\caption{In static task allocation, we determine the set of all algorithms required to perform any task for which the system is designed. The goal is to determine which unit should be assigned to each algorithm for execution so that each individual task can be optimally executed by each unit. $\mathbf{A}$ is the set of all algorithms required by the system to perform all the tasks for which the system is designed, and each of the tasks is denoted by a DAG with precedence order. $A_i$ denotes a single algorithm.}
\label{figu01}
\end{minipage}\hfill
\begin{minipage}{.48\textwidth}
\centering
\includegraphics[width=1\linewidth]{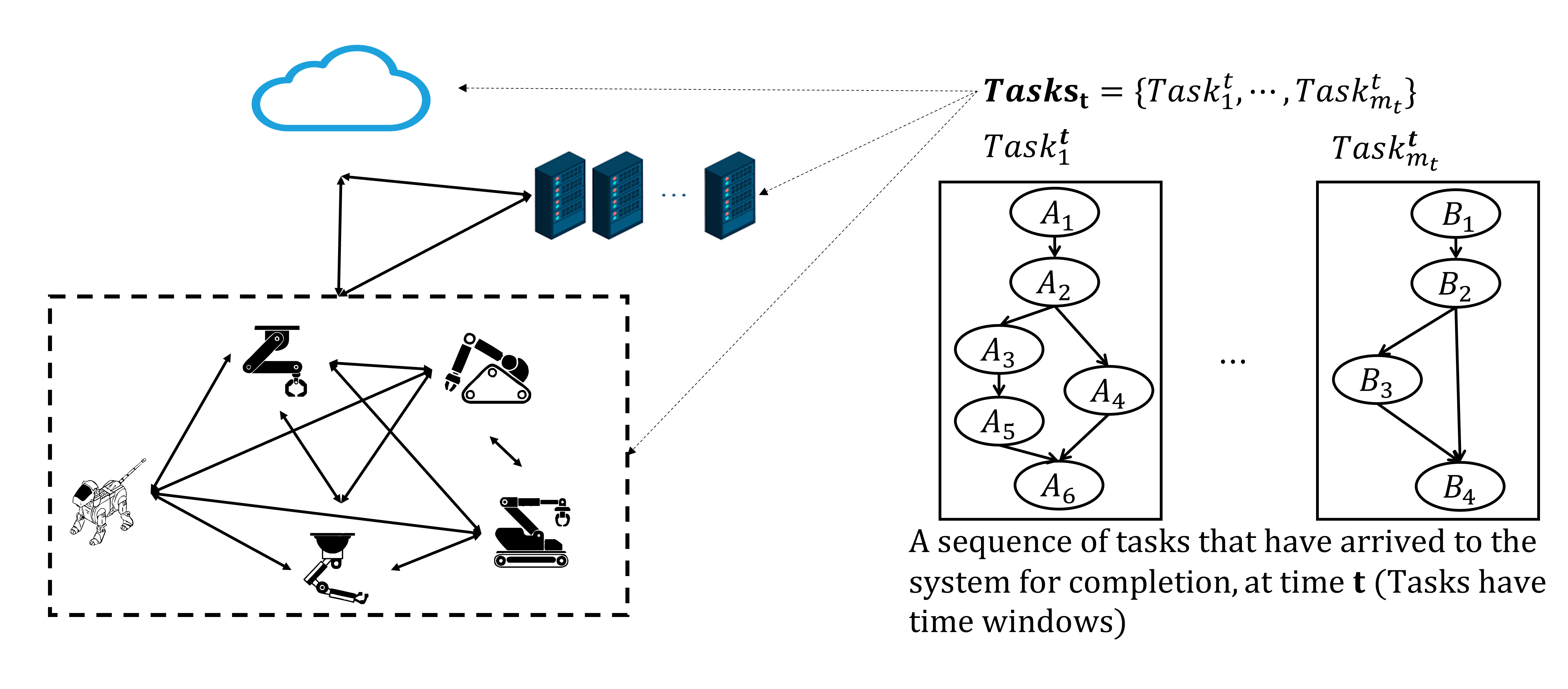}
\caption{In dynamic task allocation (task scheduling), a set of tasks arrives to the system at time $t$ to be executed. The goal is to determine to which unit each task should be assigned for execution in order to optimally complete all requested tasks. $\mathbf{Task_t}$ is the set of tasks that have arrived to the system at time $t$, and each of the newly arrived tasks in $\mathbf{Task_t}$ is denoted by a DAG with precedence order. $A_i$ and $B_i$ are used to denote algorithms of tasks.}
\label{figu02}
\end{minipage}
\end{figure}

Cloud robotics overcomes computational limitations by using the internet to assign tasks and share data in real time \cite{kehoe:2015}. An important factor in determining the performance of cloud-based robotic systems is deciding whether to upload a newly arrived task to the cloud, process it on a server (fog computing \cite{bonomi:2012}), or execute it on one of the robots (edge computing \cite{shi:2016}), which is called the allocation problem, see Figure \ref{figu01}. When the allocation problem is solved, the result is a set of tasks assigned to each processing unit. The scheduling problem deals with the problem of arranging (scheduling) this set of tasks, given the priority of the tasks, their time constraints, and the precedence order among the tasks, to answer the question of which task, from the set of tasks assigned to each processing unit, should be executed first, see Figure \ref{figu02}.

Robotic systems often need to be reconfigured to cope with new tasks. To do this, the tasks must be divided into smaller subtasks that must be executed in a predefined sequence.


A robotic system is usually designed to perform a few specific tasks. To optimize the execution of the tasks within the system, it is important to determine the responsible components for each task. The answer to this question lies in \textbf{static task allocation}. Static task allocation is the process of assigning tasks to different robots or processing units within a network based on the capabilities of the robots and processing units and the requirements of the tasks that enable their execution. In this process, the tasks are viewed as sets of basic subtasks (algorithms) that must be executed in a specific order to accomplish the task at hand. This process is usually carried out statically, i.e. the assignment of the algorithms is fixed at the beginning and does not change dynamically during execution, as shown in Figure \ref{figu01}.

In dynamic scenarios, tasks arrive sequentially over time and require real-time decision making to ensure optimal performance. \textbf{Dynamic scheduling} determines which part of the system is responsible for performing each task and when each task should be performed. In dynamic scheduling, tasks are assigned and reassigned to robots based on the current state of the system, such as the availability of robots, task priorities and changes in the environment. This approach is illustrated in Figure \ref{figu02}. The system continuously adapts to ensure optimal performance.

Static task allocation can be seen as the first step, determining which robots should perform which tasks based on predefined criteria such as the precedence relationship between the algorithms required to perform each task. Scheduling then takes over the dynamic management of these tasks and ensures efficient execution when conditions change.

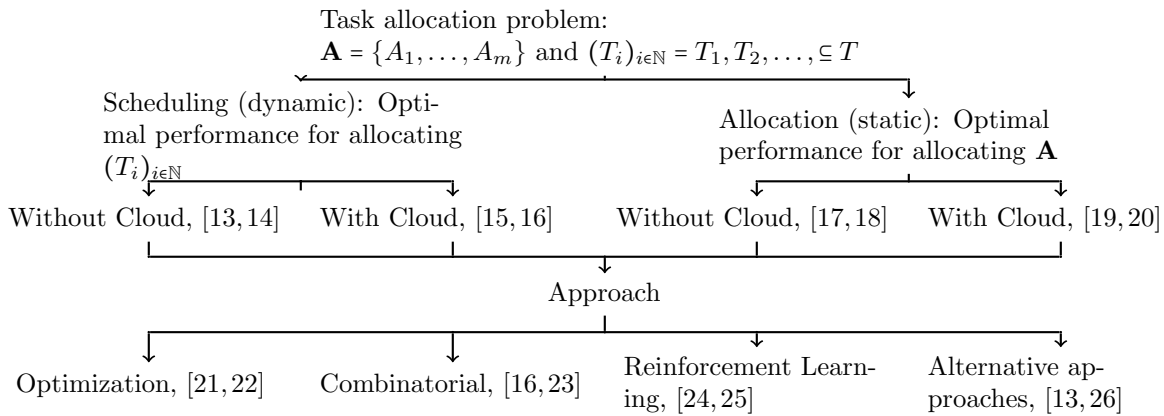
\begin{figure*}[tb]\centering
\begin{tikzpicture}
\begin{scope}
    \node (A) at (0,0.1) [text width=7.5cm] {Task allocation problem: \\$\mathbf{A}=\{A_1,\ldots,A_m\}$ and  $(T_i)_{i\in\mathbb{N}}=T_1,T_2,\ldots, \subseteq T$};
    \node[fill,circle,scale=0.1]  (B) at (0,-0.45) {};
    \node[fill,circle,scale=0.1]  (C) at (-4,-0.45) {};
    \node[fill,circle,scale=0.1]  (D) at (4,-0.45) {};
    \node (E) at (-4,-1.2)  [text width=5.2cm] {Scheduling (dynamic): Optimal performance for allocating $(T_i)_{i\in\mathbb{N}}$};
    \node (F) at (4,-1.2)  [text width=5cm] {Allocation (static): Optimal performance for allocating $\mathbf{A}$};
    \node[fill,circle,scale=0.1]  (G) at (-4,-1.8) {};
    \node[fill,circle,scale=0.1]  (H) at (-6,-1.8) {};
    \node[fill,circle,scale=0.1]  (I) at (-2,-1.8) {};
    \node[fill,circle,scale=0.1]  (J) at (4,-1.8) {};
    \node[fill,circle,scale=0.1]  (K) at (2,-1.8) {};
    \node[fill,circle,scale=0.1]  (L) at (6,-1.8) {};
    \node (M) at (-6,-2.3) [text width=3.7cm] {Without Cloud, \cite{Ours:2022ral, Sa:2021}};
    \node (N) at (-2,-2.3) [text width=3.5cm] {With Cloud, \cite{Shafiq:2021, Bharti:2022}};
    \node (O) at (2,-2.3) [text width=3.7cm] {Without Cloud, \cite{Fu:2021,Orr:2020}};
    \node (P) at (6,-2.3) [text width=3.5cm] {With Cloud, \cite{Minjia:2021,Pu:2021}};
    \node[fill,circle,scale=0.1]  (Q) at (-6,-2.8) {};
    \node[fill,circle,scale=0.1]  (R) at (-2,-2.8) {};
    \node[fill,circle,scale=0.1]  (S) at (2,-2.8) {};
    \node[fill,circle,scale=0.1]  (T) at (6,-2.8) {};
    \node[fill,circle,scale=0.1]  (U) at (0,-2.8) {};
    \node (V) at (0,-3.3) [text width=1.5cm] {Approach};
    \node[fill,circle,scale=0.1]  (W) at (0,-3.8) {};
    \node[fill,circle,scale=0.1]  (X) at (-6,-3.8) {};
    \node[fill,circle,scale=0.1]  (Y) at (-2,-3.8) {};
    \node[fill,circle,scale=0.1]  (Z) at (2,-3.8) {};
    \node[fill,circle,scale=0.1]  (a) at (6,-3.8) {};

    \node (b) at (-6,-4.5) [text width=3.5cm] {Optimization, \cite{Bai:2022,Ours:2022net}};
    \node (c) at (-2,-4.5) [text width=3.5cm] {Combinatorial, \cite{Jin:2022,Bharti:2022}};
    \node (d) at (2,-4.5) [text width=3.5cm] {Reinforcement Learning, \cite{Bian:2019,Ding:2020}};
    \node (e) at (6,-4.5) [text width=3.5cm] {Alternative approaches, \cite{Ours:2022ral,Ours:2020h}};

\end{scope}

    \path [->] (A) edge[thick,-] (B);
    \path [->] (C) edge[thick,-] (D);
    \path [->] (E) edge[thick,-] (G);
    \path [->] (F) edge[thick,-] (J);
    \path [->] (H) edge[thick,-] (I);
    \path [->] (K) edge[thick,-] (L);
    \path [->] (M) edge[thick,-] (Q);
    \path [->] (N) edge[thick,-] (R);
    \path [->] (O) edge[thick,-] (S);
    \path [->] (P) edge[thick,-] (T);
    \path [->] (Q) edge[thick,-] (T);
    \path [->] (V) edge[thick,-] (W);
    \path [->] (X) edge[thick,-] (a);

    \path [->](C) edge [thick,->] (E);
    \path [->](D) edge [thick,->] (F);
    \path [->](H) edge [thick,->] (M);
    \path [->](I) edge [thick,->] (N);
    \path [->](K) edge [thick,->] (O);
    \path [->](L) edge [thick,->] (P);
    \path [->](U) edge [thick,->] (V);
    \path [->](X) edge [thick,->] (b);
    \path [->](Y) edge [thick,->] (c);
    \path [->](Z) edge [thick,->] (d);
    \path [->](a) edge [thick,->] (e);

\end{tikzpicture}
\caption{Overview of approaches to task allocation and scheduling. We exemplify each with two references. $T$ is a finite set of tasks, $\mathbf{A}$ is the set of all algorithms that the system needs to perform all tasks in $T$ for which the system is designed. All tasks in $T$ can be completed by executing some algorithms from $\mathbf{A}$ with a certain order of precedence. $T_i\subseteq T$ is the set of tasks arriving in the system at time step $t_i$, where $t_1 < t_2 < \ldots$ is the time at which a new set of tasks arrives in the system.}
\label{fig0}
\end{figure*}

We assume that the system can only perform a finite set of tasks $T$ and that over time a set of tasks $T_i$, a subset of $T$, arrives that must be performed by the system. Figure \ref{fig0} shows an overview of the approaches to task allocation.

Allocation and scheduling problems usually involve optimizing one or more parameters such as time, total energy consumption, total resource consumption, average memory usage, etc., for completing all tasks. If the problem is limited only to the use of a cloud server, parameters such as load balancing and minimizing the makespan should be considered. 

Task allocation and scheduling are fundamental components of real-world system optimization problems. One of the most important applications for task allocation and scheduling is manufacturing, where the prioritization of resources such as labor, raw materials, and equipment must be optimized based on the urgency of production schedules \cite{Faster:2024}. A common scenario is the execution of a series of jobs by robotic systems. A key challenge in robotic networks is determining the optimal type and number of robots required and efficiently distributing computational tasks across cloud, fog and edge computing resources. Task allocation and scheduling aim to answer these questions. Despite advances in these areas, there are still major challenges, especially for robotic networks operating in dynamic and unpredictable environments. Existing methods often struggle with issues of scalability, real-time processing and the integration of cloud, fog and edge computing to effectively optimize robotic systems.

In the context of robotic network systems, cloud computing refers to the use of centralized data centers for processing tasks. Fog computing involves intermediate nodes (fog nodes) that process data closer to the source than the cloud, but further away than edge devices. Edge computing refers to the execution of tasks directly on local devices such as robots or sensors, which can reduce latency and bandwidth usage as processing takes place close to the data source.

These infrastructures offer varying levels of computing power, latency and bandwidth utilization, which affects task scheduling strategies. For example, cloud computing offers high computing capacity but can lead to delays due to the distance to the data source, while edge computing minimizes latency by processing tasks locally.

With the increasing use of robotic networks in industries such as manufacturing, healthcare and disaster management, the need for efficient, scalable and adaptable methods of task allocation and scheduling is becoming more pressing. This is especially important given the emergence of distributed computing paradigms such as cloud, fog and edge computing, which present new opportunities and challenges for robotic systems.

This paper makes the following contributions: (1) it provides an overview of methods for task allocation and scheduling in robotic networks and categorizes them, focusing on the main approaches, including optimization, combinatorics, and reinforcement learning. The review highlights how these methods take into account different system characteristics, such as static vs. dynamic task allocation and load balancing; (2) it provides an evaluation of the strengths and limitations of these methods, focusing in particular on their applicability to robotic systems of different sizes and types (e.g., static, dynamic, human-robot collaboration). It also considers the importance of cloud, fog and edge computing in optimizing task scheduling; (3) critical challenges for existing methods are identified, including scalability and adaptability in dynamic environments, and potential avenues for future research are highlighted, e.g., the development of hybrid models that combine the strengths of multiple approaches to improve performance in real-time and resource-constrained settings.

In this paper, we categorize the methods of task allocation and scheduling according to their underlying mathematical approaches, such as optimization, combinatorics and reinforcement learning. This categorization approach was chosen because task allocation methods are usually independent of the type of robot used. Instead, the focus is on optimizing resource allocation and system efficiency, regardless of whether the robot is a drone, a humanoid or a small robot. These mathematical methods offer generalizable frameworks that are applicable to robotic systems with different types of robots.

By providing a comprehensive overview of the state-of-the-art task allocation and scheduling methods, this paper aims to provide a solid theoretical foundation for researchers looking to explore more specialized areas of robotic network optimization and bridge the gap between theoretical advances and practical, real-world implementations.

We have reviewed high-level studies on "scheduling" and "allocation" from 2017 to 2024. Studies were selected that address the problem in robotic networks or in the cloud. We tried to cover most of the methods used in recent studies. This paper provides a comprehensive overview of methods for task allocation and scheduling in robotic networks. It categorizes the existing solutions according to their underlying approach (optimization, combinatorics and reinforcement learning) and provides a detailed analysis of the performance, scalability and applicability of these methods. In addition, this survey highlights open research challenges, especially in the integration of cloud, fog and edge computing, and suggests directions for future research.

\section{Related Work}
In the article \cite{Arunarani:2019}, articles related to scheduling in cloud computing were examined. The authors examined all articles with the word ''scheduling'' in the title or keyword published from January 2005 to March 2018. The article explains the importance of task scheduling, and that it cannot be done manually. The main benefits of proper task scheduling are: (for the user) spending less money on using virtual machines, getting the result of task execution faster, among others, and (for the cloud provider) processing a large number of incoming requests, spending less on service maintenance, providing the best quality of service, and so forth. They classify existing task scheduling techniques into ten categories and briefly explain each category with its advantages and limitations. They categorize the papers according to the year of publication, the techniques and parameters they measure, and state their limitations and highlighted time complexity. Studies on scheduling and task allocation, both with and without cloud infrastructure, provide complementary methods for broader applications. The paper \cite{Arunarani:2019} focuses mainly on works on cloud computing and works that apply scheduling methods to robotic networks are not included. Scheduling methods are also evolving rapidly, and several new methods have been developed in the last six years.

The authors of the article \cite{Rizk:2019} examined studies on heterogeneous multiagent systems and focused mainly on robot agents. They considered studies on task decomposition, coalition formation, task allocation, and multiagent scheduling. Most studies dealing with cloud infrastructures have not considered optimal task allocation in the cloud. The automation of the task allocation problem is divided into three stages depending on the usage of human agents to find an optimal task allocation. Finally, task decomposition, coalition formation, and task allocation are considered separately. Several challenges such as using Big Data, considering task complexity, applying machine learning, human-robot collaboration, and communication instability are identified for further exploration. Most of these challenges have been explored in recent years and several results have been proposed.

The review article \cite{Dawarka:2022} examined the studies on the cloud robotic architecture developments without considering task allocation and scheduling, which is beneficial for further optimizing the performance of a cloud robotic system.

\section{Categorization}

\begin{figure*}
     \centering
     \begin{subfigure}[b]{0.24\linewidth}
         \centering
         \includegraphics[width=1\linewidth]{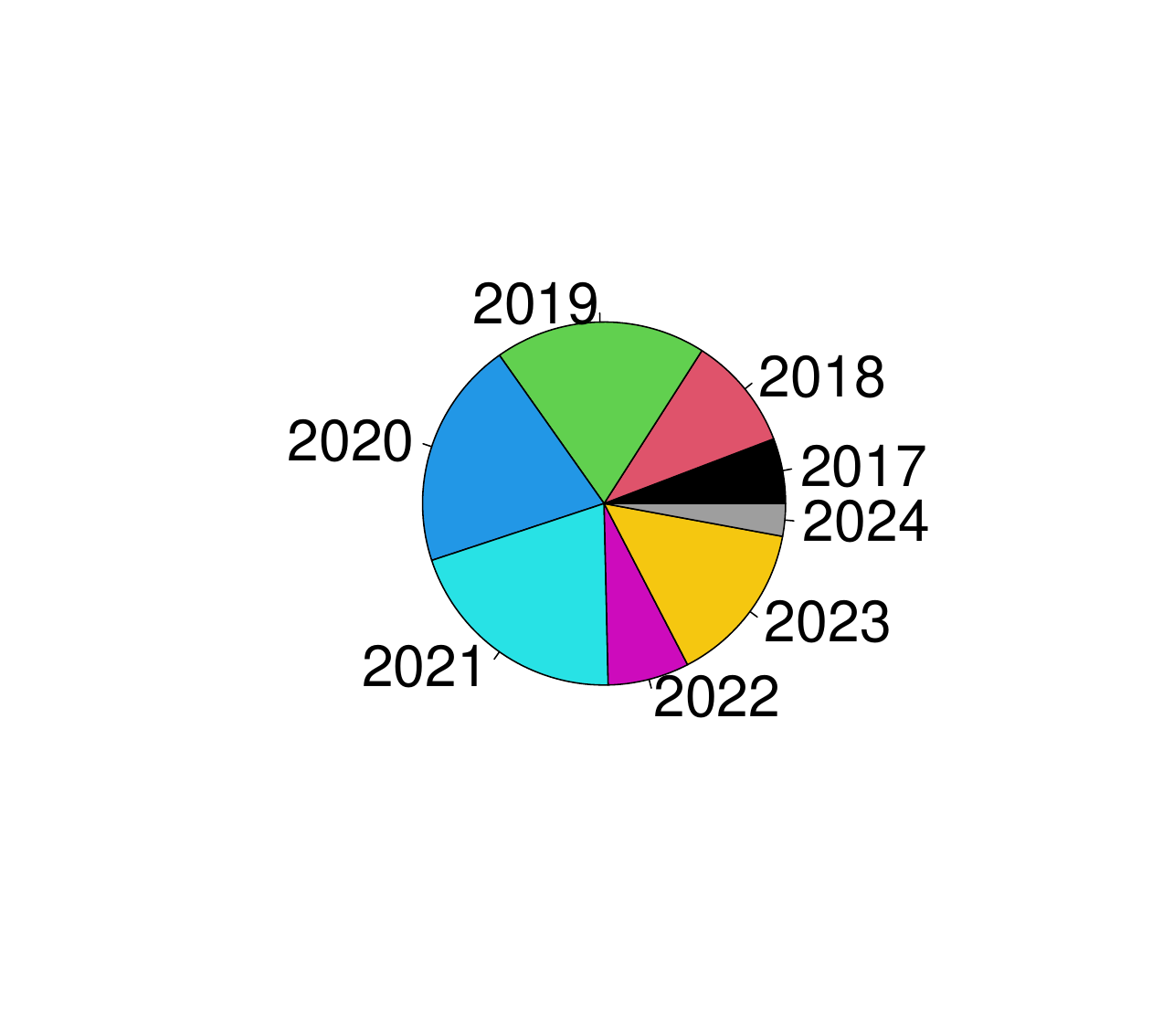}
         \caption{\footnotesize Categorizing by year.}
     \end{subfigure}
     \hfill
     \begin{subfigure}[b]{0.24\linewidth}
         \centering
         \includegraphics[width=1\linewidth]{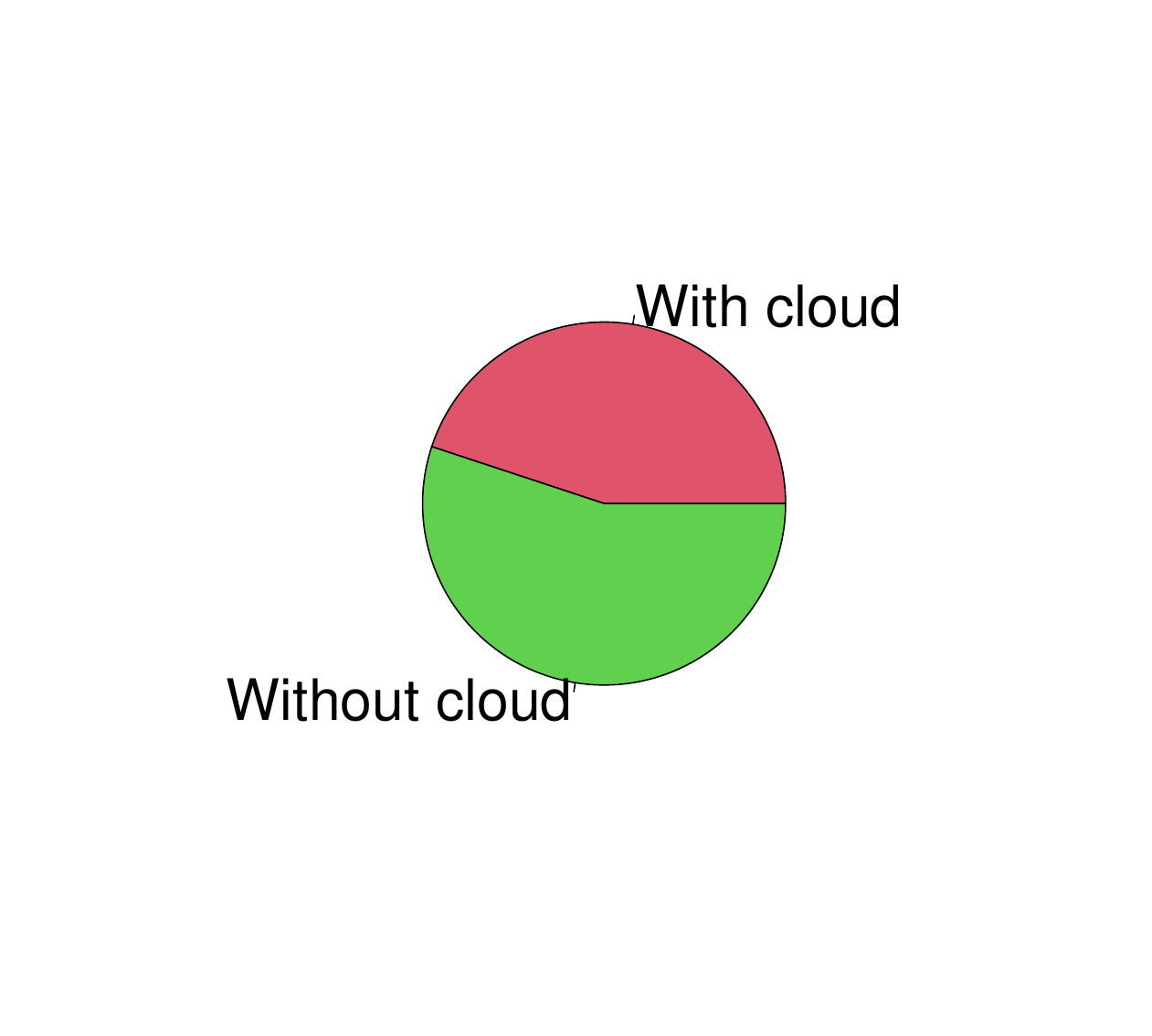}
         \caption{\footnotesize Categorizing by considering cloud or not.}
     \end{subfigure}
     \hfill
     \begin{subfigure}[b]{0.24\linewidth}
         \centering
         \includegraphics[width=1\linewidth]{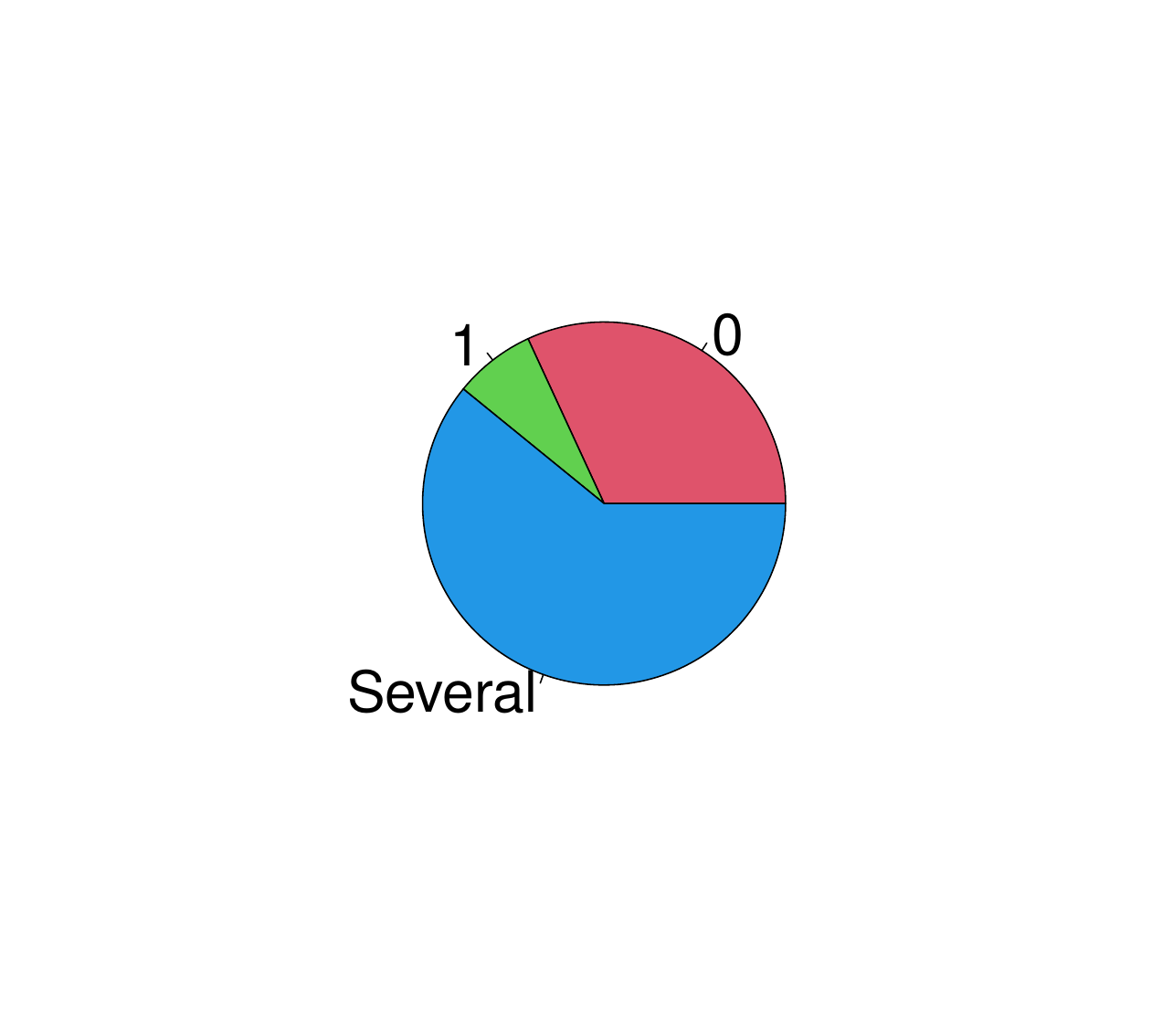}
         \caption{\footnotesize Categorizing by number of robots.}
     \end{subfigure}
     \hfill
     \begin{subfigure}[b]{0.24\linewidth}
         \centering
         \includegraphics[width=1\linewidth]{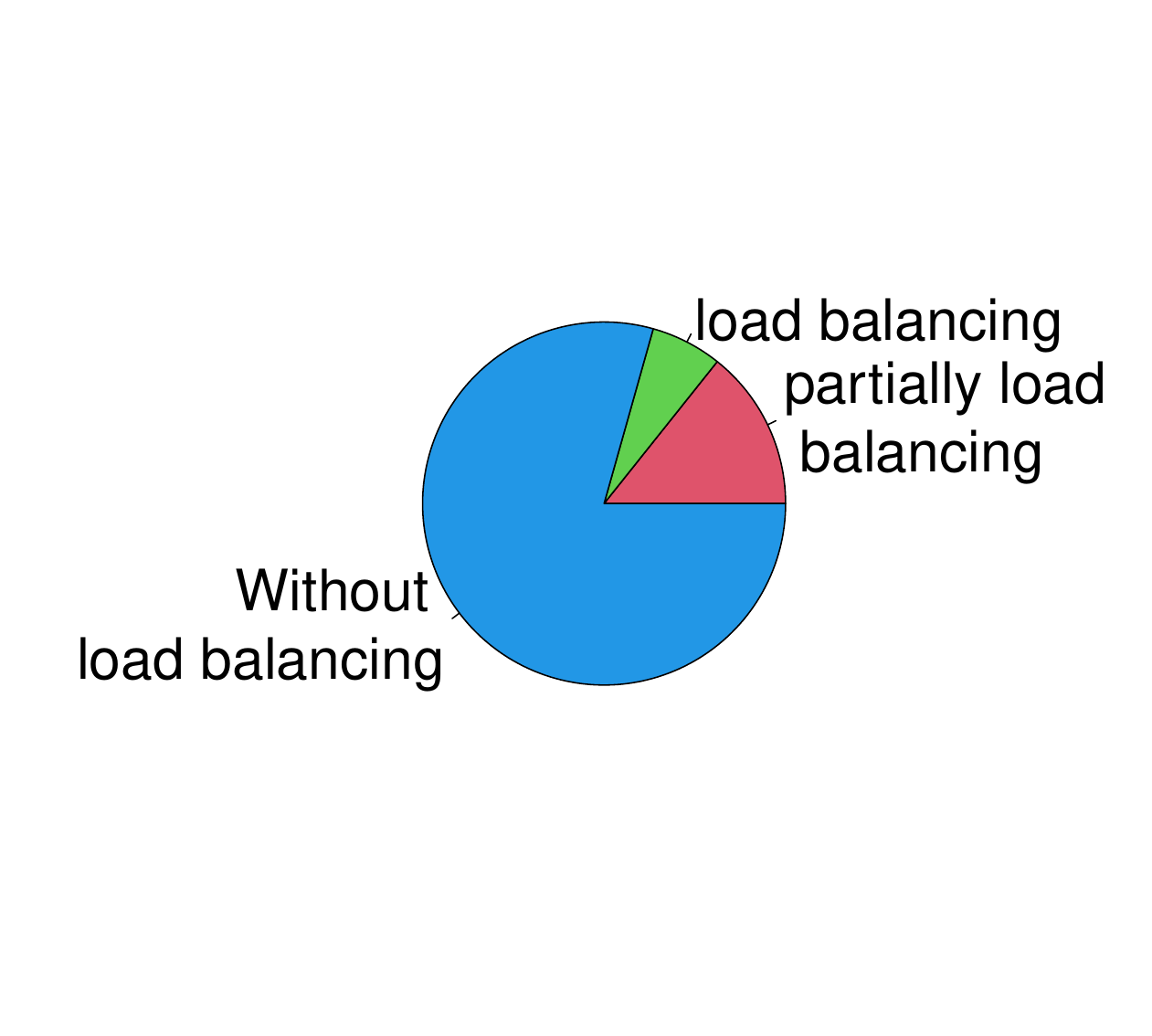}
         \caption{\footnotesize Categorizing by aiming to balance the loads.}
     \end{subfigure}
     \hfill
     \begin{subfigure}[b]{0.24\linewidth}
         \centering
         \includegraphics[width=1\linewidth]{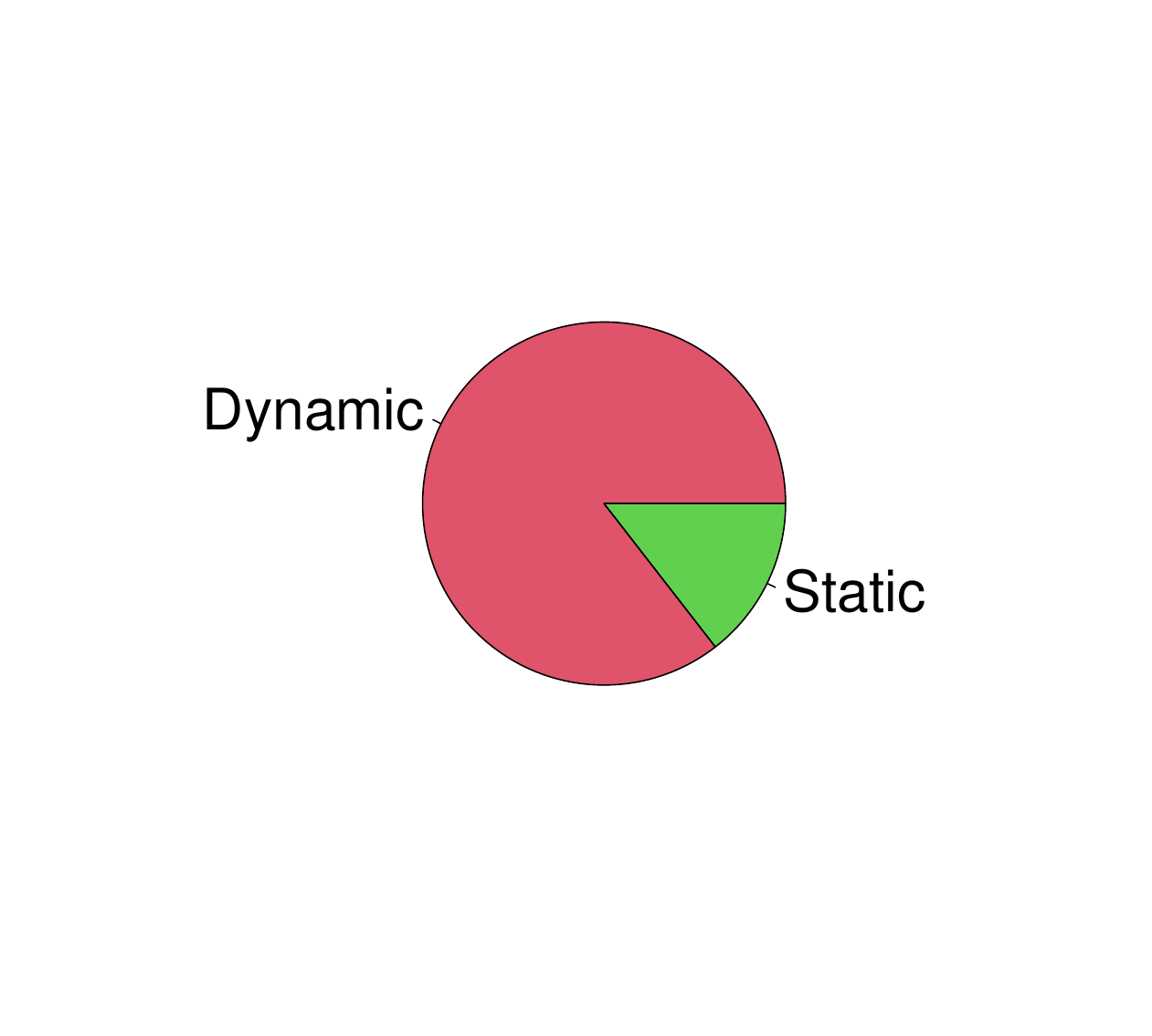}
         \caption{\footnotesize Categorizing by being static or dynamic task assignment.}
     \end{subfigure}
     \hfill
     \begin{subfigure}[b]{0.24\linewidth}
         \centering
         \includegraphics[width=1\linewidth]{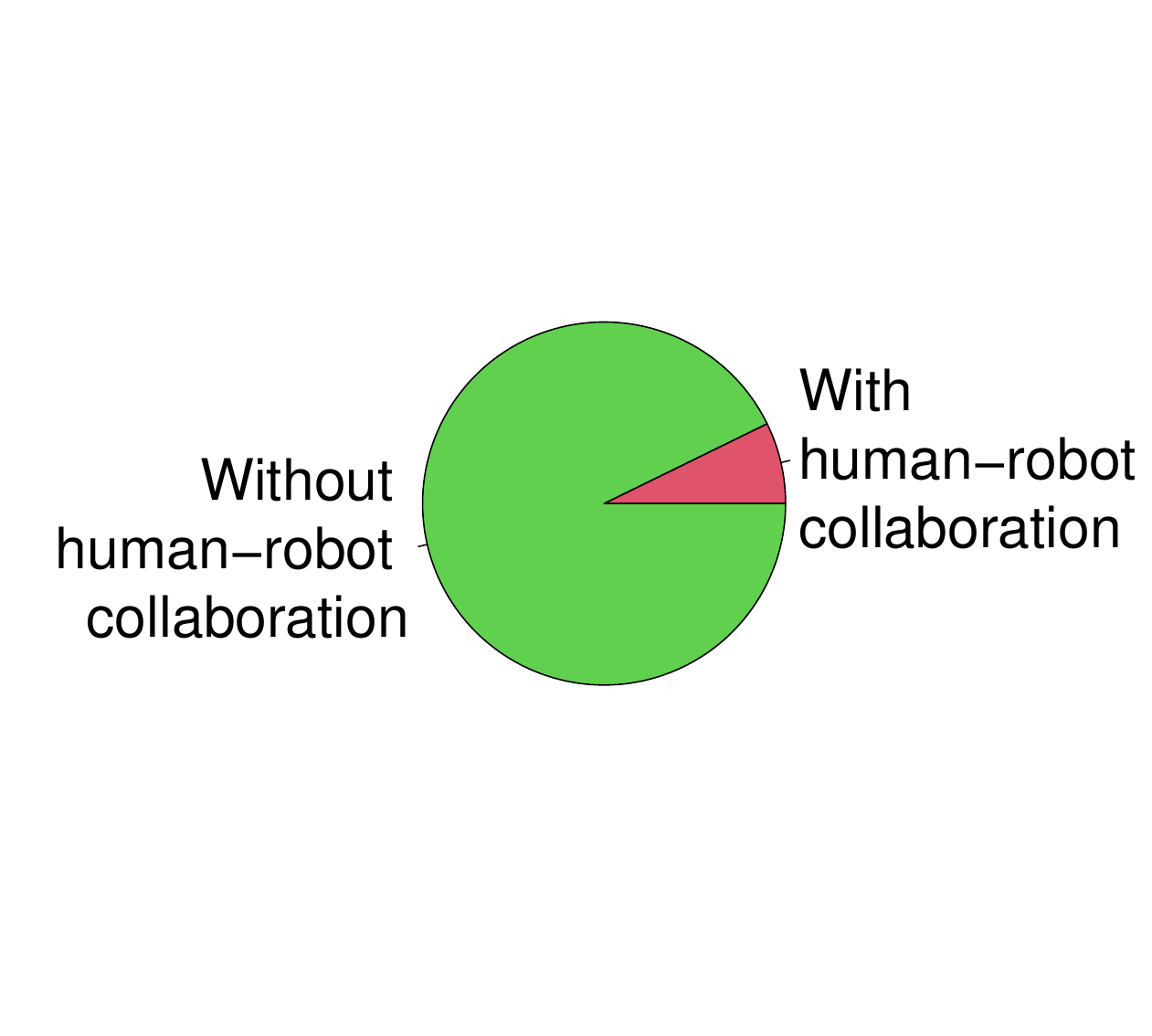}
         \caption{\footnotesize Categorizing by considering human-robot collaboration.}
     \end{subfigure}
     \hfill
     \begin{subfigure}[b]{0.24\linewidth}
         \centering
         \includegraphics[width=1\linewidth]{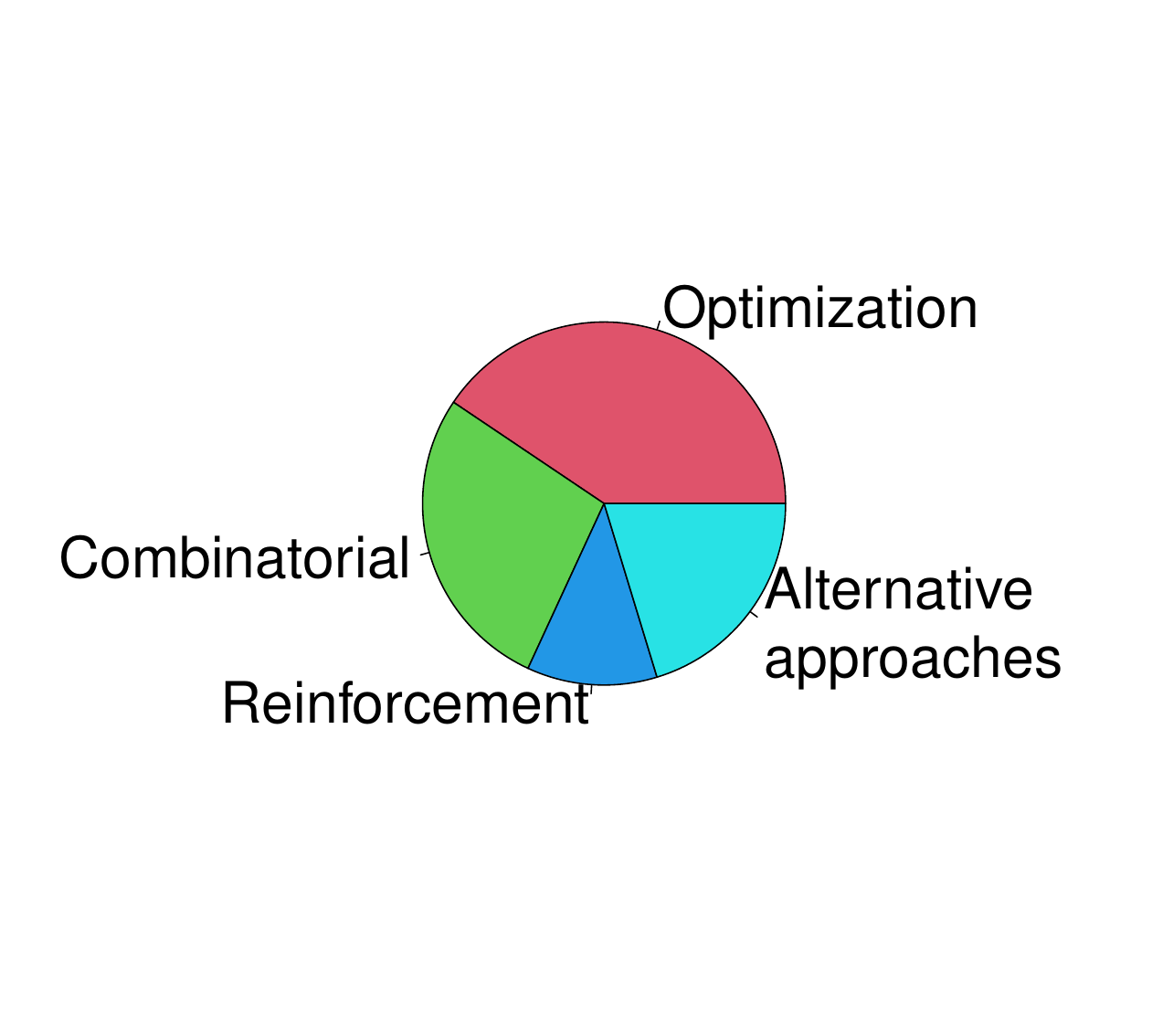}
         \caption{\footnotesize Categorizing by the approach used to solve the problem.}
     \end{subfigure}\hfill
        \caption{Different categorization of all contributions.}
        \label{fig1}
\end{figure*}

We present an overview of the works studied in the form of diagrams organized according to various distinguishing features, as shown in Figure \ref{fig1}. These diagrams provide an overview of the trends in research on task allocation and scheduling. They help to illustrate the evolution of the field and the increasing focus on different architectural and methodological aspects, such as the shift towards edge computing or the emergence of combinatorial and reinforcement learning approaches.

All contributions have been considered according to the following criteria:
\begin{itemize}
\item Year: The year of publication of the developed model.
\item Static/Dynamic: Two types of task assignment: In static allocation, tasks are assigned from a predefined set, while in dynamic allocation, tasks are assigned as they arrive.
\item Load balancing: Whether the scheduling model ensures that all processing units complete their tasks at almost the same time.
\item Cloud infrastructure: Whether the scheduling model is designed for a cloud-based system.
\item Number of robots: The number of robots in the system.
\item Parameters: The parameters used to develop the scheduling model.
\item Main objectives: The main goal of the developed model.
\item Approach used: The method used to solve the task allocation problem.
\item Restrictions: Constraints that should be considered when applying the model to other problems.
\item Problems: Limitations of the model that need to be investigated further.
\item Type of experiments: Whether the model was validated by simulation or real-world tests.
\end{itemize}
We classify the papers according to their solution approach for scheduling and task allocation and then sort them by year of publication, as shown in Figure \ref{fig0}. The solution approaches include:
\begin{itemize}
\item Optimization: Formulation of the problem as an optimization model.
\item Combinatorial: Translating the problem into a graph-theoretic or set-theoretic model.
\item Reinforcement learning: Solving the problem with the help of reinforcement learning techniques.
\item Alternative approaches: Other models, such as language and automata theory, geometric models or novel architectures.
\end{itemize}
While the methods of task allocation and scheduling could also be categorized by system models and the challenges they address, in this paper they are categorized by solution approaches to provide a clearer understanding of the different mathematical frameworks and methods in this area. By organizing the papers in this way, we highlight the different theoretical foundations such as optimization, combinatorics, and reinforcement learning that can be applied in different system models and environments. This approach is particularly valuable for researchers focused on developing robust, generalizable solutions that can be adapted to different real-world challenges of robotic networks. By categorizing task allocation methods according to mathematical approaches, we highlight how these strategies address key challenges in robotic networks, such as resource management and real-time performance. This categorization allows us to focus on the generalizable aspects of the methods that are independent of specific robotic types and can be adapted to different system architectures.

It is important to note that robotic network systems operate in different environments with different resources, task complexity and constraints. For example, unpredictable network latencies, hardware failures or different dependencies between tasks may occur in real systems. As a result, the applicability and validity of certain metrics, such as energy consumption or task completion time, may differ significantly between simulation environments and real-world deployments. Many of the metrics used in our study are tailored to specific aspects of robotic network systems and may not offer universal applicability for all scenarios. This diversity makes it difficult to produce intuitive outcome figures that effectively capture the benefits of the different methods in each scenario. The detailed explanations in the text are intended to complement the summarized data and provide insights into the strengths and limitations of each approach when applied in different real-world contexts.

With this categorization of papers, we aim to provide a comprehensive understanding of how different approaches can address the complex challenges of robotic network systems. This categorization allows researchers to identify the most appropriate methods for their specific applications while highlighting areas where further research is needed.

All contributions are summarized in the supplementary material based on their solution approaches.

\section{Optimization}
In this section we discuss contributions that solve the problem using an optimization approach and apply a standard solver to obtain the result. The works are separately analyzed based on whether the cloud infrastructure is considered or not.
\subsection{Without Cloud}
\cite{Gini:2017} investigated the task allocation of multi-robots, taking into account the time windows constraints of the tasks and their precedence order, and proposed an optimization approach that can be solved by several possible methods, such as market-based, swarm-based, distributed constraint optimization methods as a decentralized approach or branch-and-bound method as a centralized approach to minimize a cost function or maximize a reward function for all robots for completing their tasks. The temporal model is translated into logical expressions, \cite{Allen:1983}, and then the logical expressions are translated into graphs that facilitate their use, \cite{DECHTER:1991}. Dynamic task allocation considers changes in the environment and provides solutions through centralized approaches like branch-and-bound and metaheuristics as well as decentralized methods such as DCOP-based techniques and market-based approaches. 


All robots in the decentralized model should be able to communicate with each other and in the centralized model, the central unit should be able to communicate with all robots. Moreover, communication failures have not been considered, which means that in the centralized model, the central unit has to reschedule all tasks and generate new solutions when new tasks arrive, while in the decentralized model, approximation methods should be used to reduce the computation time. Moreover, robots that can perform multiple tasks simultaneously and multi-robot tasks are not considered, and disjoint temporal models are not considered.

\cite{NUNES:2017} studied the ordering of tasks with time window constraints. They described a general optimization model for the task allocation problem from which all existing methods can be derived by considering a general optimization of a generic function, which can be a cost function, a reward, a distance to the task, etc., and then, depending on the problem, maximize the number of tasks completed, minimize the sum of the total path cost of all robots, minimize the latency, maximize the reward, and minimize the number of robots deployed. 


The time constraints are simple and time-critical tasks cannot be correctly assigned to robots. To find a solution, they investigated various existing proposals according to the main problem by considering the centralized and the distributed solution methods, but there is no clear way to solve problems with multiple objectives.

\cite{LEE:2018} investigated the optimization of task completion time, resource consumption, and communication time. The author proposed a resource-based task allocation method to ensure the efficiency of task allocation over a long period of time. The proposed task allocation method is a market-based approach. And it assumes that occasional recharge and resources are consumable resources available to robots in task allocation, which includes rescheduling of tasks. 


The method does not consider the two parameters recharge time and transfer time to recharge stations, which may change the solution. Moreover, their method is compared with the cases without task rescheduling and without considering resources. To improve performance, additional constraints are introduced into the optimization problem and performance is experimentally compared to methods without considering the additional constraints. However, the improved performance is apparent even without testing.

\cite{Zhou:2018} developed a method that finds the smallest number of robots in a given time limit and assigns tasks to them so that the robots can complete their tasks within the time-limit. The authors proposed a multi-objective approach called Mofint (multi-objective GA with forest individual containing non-intersecting trees), which first finds upper and lower bounds on the number of robots and then determines the optimal completion time by finding an optimal spanning forest. The main problem they addressed is the complete coverage problem (CCP), which is equivalent to finding an optimal spanning forest in graph theory \cite{Pettie:2008, Hell:1985}. The problem is to find the smallest path that avoids obstacles between all points in a given area. First, an area is divided into several regions with respect to obstacles. Each region is considered as a task, and each task is assigned to a robot. Considering the regions as vertices and adjacent regions as edges, a graph is obtained. Given a number of robots, the solution to the coverage problem (minimum completion time) with a given number of robots is to find the optimal spanning forest where the number of tree components equals the number of robots. Since the regions of the area are a partition of the area and each region is assigned to a robot, there is no overlap between the robots. The method for finding the regions is partitioning and homomorphic projection using Morse theory \cite{Milnor:1963}, where the area becomes a union of different bands. Next, they find the upper and lower bounds on the number of robots. Then, from the lower bound (LB) to the upper bound (UB), they test whether the completion time is shorter than the time limit. For the bounds, the LB for a fixed number of regions is equal to the smallest integer greater than or equal to the value of the sum of all weights divided by the time limit. For the UB, they generate a random spanning tree and then use the linear tree partitioning algorithm to determine the number of partitions. In this algorithm, the random spanning tree is partitioned into multiple parts such that the sum of the weights of all parts is less than or equal to the time limit. The objective is then to minimize the weight of the heaviest tree in the spanning forest. To solve the optimization problem, they randomly create a forest with a fixed number of trees between the LB and the UB. And then add random edges to create a LB-tree forest. 

\begin{figure}[th!]
\begin{minipage}{.48\textwidth}
\centering
\includegraphics[width=0.33\linewidth]{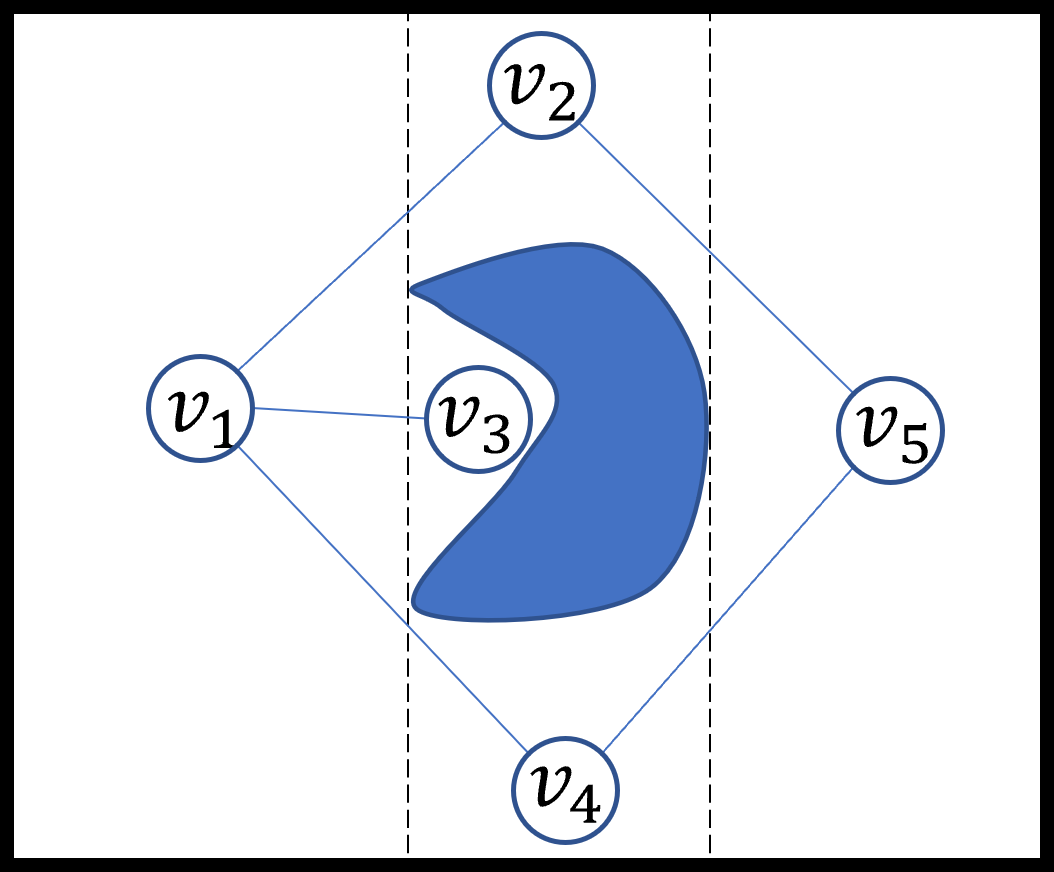}
\caption{Environment abstraction. The imaginary vertical dashed lines are some borders of regions, $v_1$ to $v_5$. The filled object in the middle of the environment is the obstacle.} 
\label{figu12}
\end{minipage}\hfill
\begin{minipage}{.48\textwidth}
\centering
\includegraphics[width=.72\linewidth]{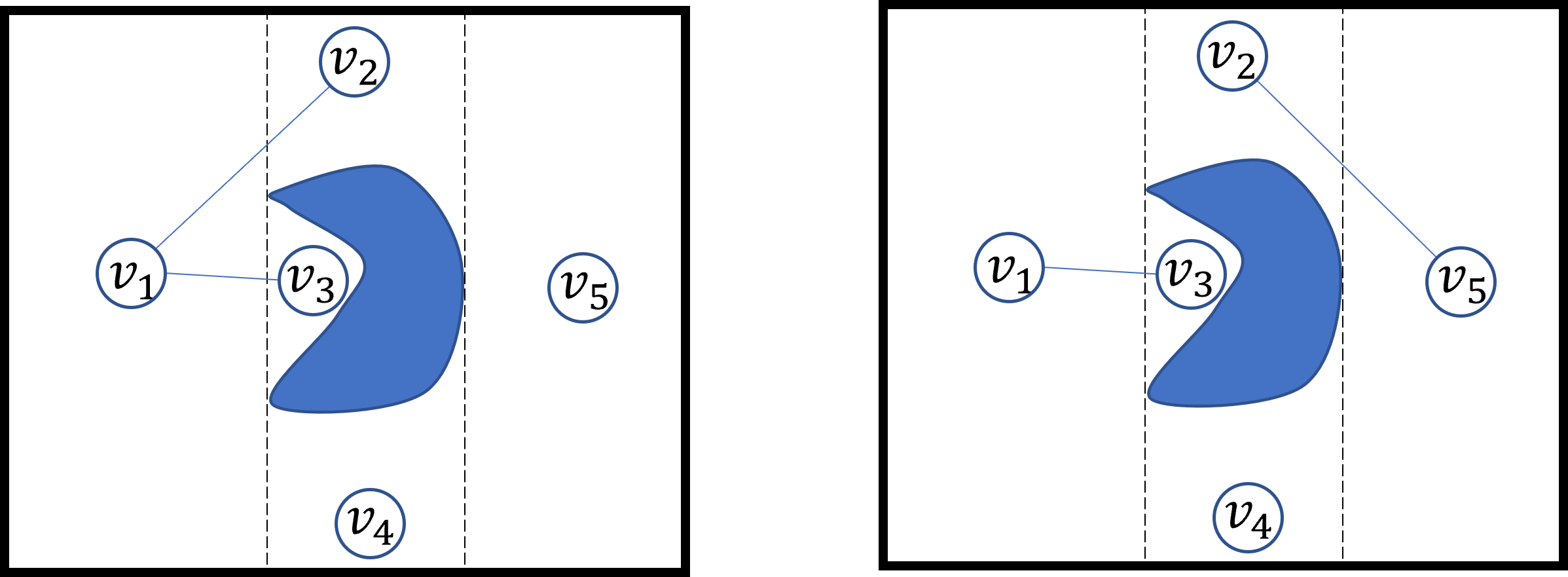}
\caption{Task allocation. Two possible allocation to three robots. Tasks for each robot can form a spanning tree. The partitioning vertices in the left will pass the time limit, $T=2$, but the right one does not pass the time limit.} 
\label{figu13}
\end{minipage}
\end{figure}

The weights of the vertices of the graph of regions are considered as non-negative integers. Since we are looking for the minimum number of robots, we should look for an appropriate partition for this number of robots. Such a partition should be defined in such a way that it does not exceed the time limit. Moreover, the definition of the band size depends on the metric, which is not defined. Next, the UB can be defined simply as the number of vertices determined by the partitioning of the region. Moreover, the partitioning of vertices into a fixed number of parts is not unique. Examples can be found where the region is partitioned into a fixed number of parts such that one part meets the time limit but another part does not, see Figures \ref{figu12} and \ref{figu13}. Therefore, the assignment of each partition of nodes to robots does not necessarily mean that it does or does not respect the time limit. To solve the optimization problem by adding random edges that either create a cycle or reduce the number of trees by one, we need to check for possible cycles in the graph after adding each random edge.

\cite{Dantu:2019} proposed a distributed dynamic task assignment algorithm for swarms of robots under the assumption of communication instability. The problem is translated into an optimization problem for maximizing the utility of all agents. In order to solve the optimization problem, it is necessary to maintain the bundle vector (the list of tasks assigned to the agents, where the tasks are ordered by when they were won), the path vector (the same as the bundle vector, but the order of tasks is the order in which the tasks are completed by the respective agents), the winning agent vector (the vector of values corresponding to the agents and tasks to determine an agent's expectation of which agent placed the highest bid on the task), the winning bid vector (the vector of values corresponding to the agents and tasks to determine an agent's expectation of placing the highest bid among all agents on the task), and the timestamp vector (the vector of values corresponding to the agents and tasks to determine when an agent placed the bid on the task). 


In the case where all tasks have relatively long execution times with respect to the average communication time, communication delays do not lead to significant changes in the allocation. Moreover, the optimization problem can be simplified and transformed into linear programming by transforming the main objective logarithmically. And then the main objective is to minimize the completion time of all tasks instead of solving a more complex objective function.

\cite{Notomista:2019} examined capabilities and energy consumption by minimizing the cost function. In this study, the tasks are prioritized based on the time constraints and capabilities of the robots and the optimization problem (Mixed Integer Quadratic Problem) is solved to find the optimal task allocation. The optimal task allocation is achieved by updating the priorities of the tasks according to time. 


The cost function must belong to class $C^1$ (functions that are continuously differentiable), which is not specified in the paper, otherwise the arguments are not valid. Moreover, for each robot, the task with the highest priority is considered unique. Moreover, all the cost functions in the experimental results and examples are the distance to a certain point in a compact Euclidean subspace is a very simple function, but other cost functions such as a combination of energy consumption, time to complete tasks and some others can be considered which increase the complexity of the optimization problem. In some scenarios, where the cost functions are not of class $C^1$, the method cannot be applied. This means that their proposed method works only for special cost functions. For example, functions such as $|2x^2-1|$ or $|\frac{2}{\pi}\arctan(\frac{\pi}{2}x)|$ are not compatible with the application of their method. Moreover, their method does not take into account environmental changes.

\cite{emam:2020} studied the minimization of the cost function in a dynamic environment where task priorities may change due to environmental changes. The proposed method is similar to that of \cite{Notomista:2019}. The optimal allocation is achieved by updating the states of the robots over time and the priorities are updated with the changes in the environment. 


The cost function is considered to be continuously differentiable and for each robot, the task with the highest priority should be unique, which cannot cover more general scenarios. Moreover, it does not consider the scenario that an environmental disturbance can reduce the cost, e.g., removing an obstacle. Also, the frequency of environment changes must be considered, e.g., repeated placement and removal of obstacles in the same location.

\cite{Rahmanpour:2020} proposed a method for planning the motion and communication strategies for a robotic network to perform all tasks with the goal of minimizing energy consumption and avoiding collisions. For each task, a robot is selected, which is called the leader. The leader plans the movements to a particular location and when the task is completed for the newly arrived task based on the current formation, a new leader is selected. Each robot solves a convex optimization problem to plan a collision-avoiding optimal trajectory and an optimal communication strategy that guarantees network connectivity in the presence of uncertain communication. If we consider an agent (node) called an operations center (OC), it is an interconnected node that enables communication among all nodes. When a task is assigned to the team, a particular robot is selected as the leader. The leader will visit the task location within a certain time, and the other robots will support the leader by passing their information to the OC. To select the leader and configure the other robots, the robots search for the shortest obstacle-free path between OC and the task location after a task at that location is announced by OC. The shortest path is defined by a sequence of straight lines between the two positions in the flat surface of the area, where the line has no intersection with obstacles. This path is divided into several (number of robots) segments of equal length. Now the shortest (obstacle-free) path from the current position of each robot to each spot of the segments is found. This information is transmitted to all the robots, and then the robot is assigned to each spot in such a way that the total length of the path that the robots have to travel to reach that spot is minimal. In this way, we can minimize the movements of the robots to design the path from OC to the position of the task. 

\begin{figure}[tbp]
\begin{minipage}{.48\textwidth}
\centering
\includegraphics[width=0.3\linewidth]{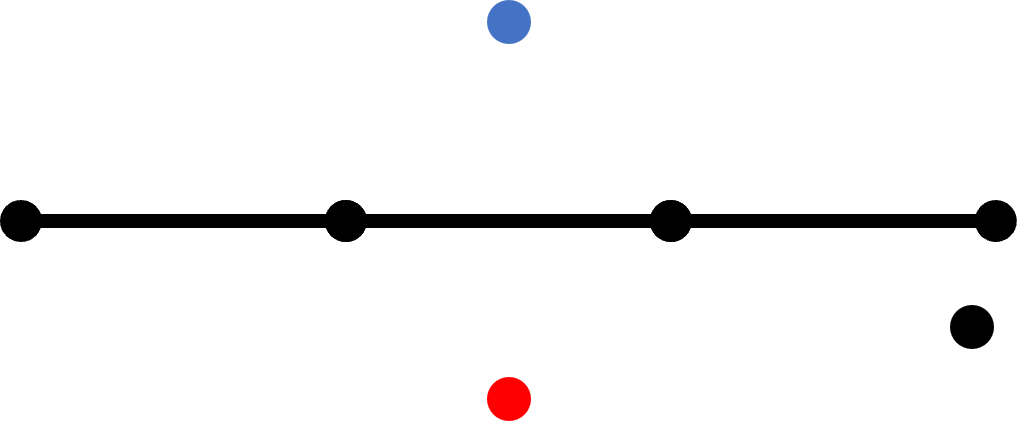}
\caption{Three robots, where two of them (red and blue dots) have identical distance to all the spots $z_{h,v}$. The line is a obstacle free path.}
\label{figu8}
\end{minipage}\hfill
\begin{minipage}{.48\textwidth}
\centering
\includegraphics[width=0.4\linewidth]{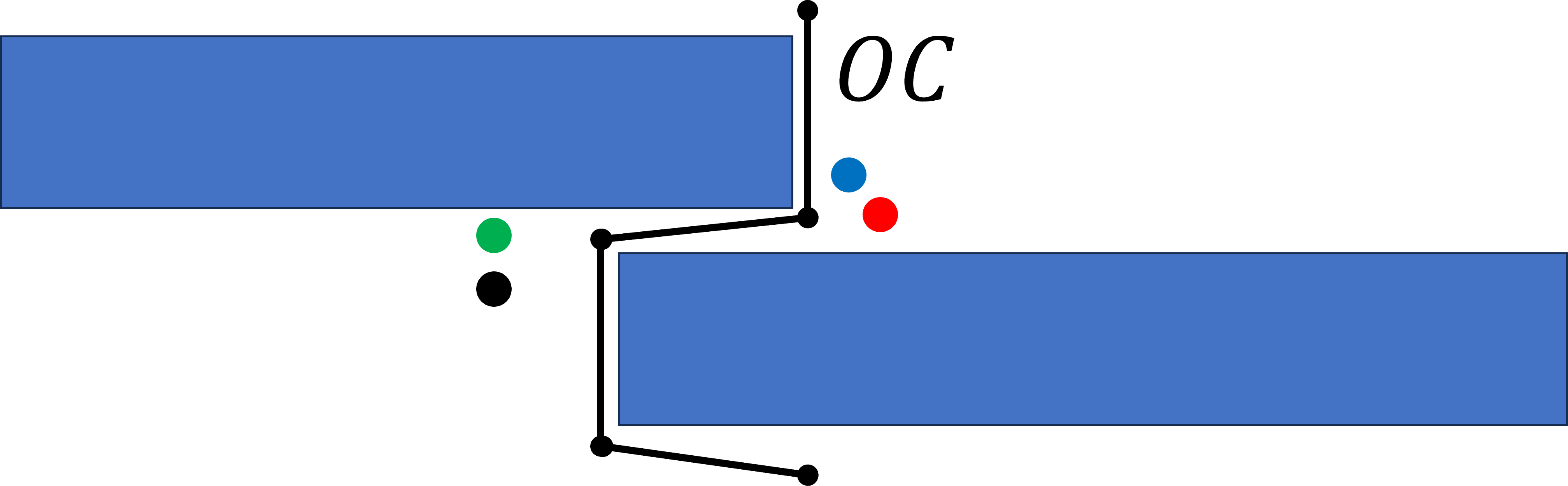}
\caption{An example of the obstacle-free path from the OC to the task, with 4 robots shown in color circles. The line is a obstacle free path.}
\label{figu9}
\end{minipage}
\end{figure}

Dynamic obstacles were not considered. In addition, the method for finding the shortest path should consider the case where there are two robots, both of which have identical minimum distances to the spots, and one is randomly selected. Moreover, the optimal assignment of robots to each segment cannot be found simply by finding the shortest path to each segment, see Figure \ref{figu8}. There are several examples where a robot should be assigned to a spot even though other robots are closer to that spot, see Figure \ref{figu9}. In addition, to find the optimal assignment of robots to spots, we need to solve the problem of minimizing the total paths traveled by all robots to each segment spot. The constraint on the threshold radius around the robots is not considered, which means that the size of the shortest distance between OC and the task for a given number of robots cannot be smaller than twice the sum of the radii around all robots; otherwise, we should use a smaller number of robots.

\cite{Brown:2020} studied the allocation and routing of robots to move objects between stations in such a way that the makespan (the time from start to finish) is minimal. The problem is translated into a problem called precedence constrained multi-agent task assignment and path-finding (PC-TAPF), which is an optimization problem to minimize the makespan. A hierarchical algorithm is proposed to find the optimal makespan. The proposed method jointly considers task dependencies, performs task scheduling, and provides routing that avoids collisions. The main idea is to first solve the problem without considering collisions. Then, the task assignment is passed to the so-called conflict-based search (CBS), which searches for the collision-free set of paths. Traveling Salesman method is used for the first part, and the tree search for the second part. A 2D environment with multiple stations is considered, i.e., places where operations are performed on objects that are regularly placed at fixed positions. The stations contain pick-up and drop-off areas, i.e., places where objects are collected or dropped off before and after performing operations. There are also several robots that move objects from one station to another, with the position of each robot being known at any given time. The stations are the area where the robots cannot move. The position of each object at any point in time is known. It is assumed that a robot can pick up an object if the object and the robot have the same position at the same time. To avoid conflicts, the method looks for possible conflicts in the solution. If there is a conflict, the same is applied for the second time, but in this case, since the number of conflicts is already known, the nodes where there is a conflict are searched and their route is replaced by a conflict-free route. The proposed method consists of a four level algorithm. The sequential next- best assignment search (NBS) tries to find a valid operating schedule for a sequential assignment problem, the conflict-based search (CBS) uses a binary tree search to find an optimal path, incremental slack prioritized search (ISPS) obtains the operating schedule and set of constraints from CBS, slack and collision aware tie-breaking $A^*$ ($A^*_{ SC }$) searches the graph to find the states corresponding to the critical path with the lowest cost.

\begin{figure}[h]\centering
\includegraphics[width=0.2\linewidth]{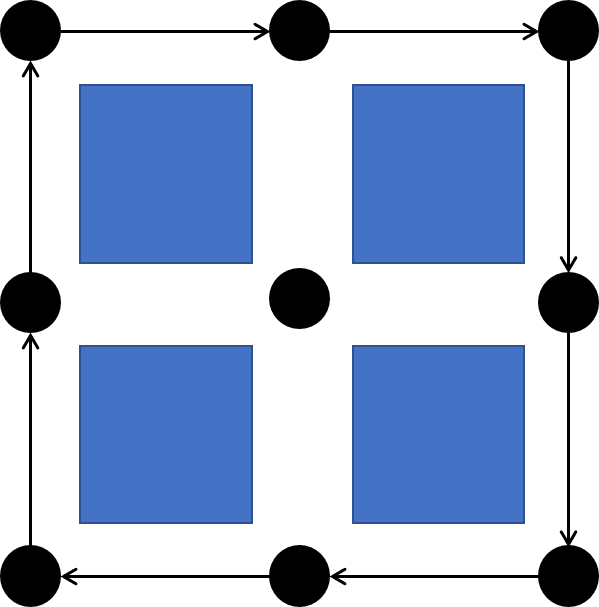}
\caption{Nine robots are planned to move. Eight planned initially with the movement without conflict, but for the middle robot there is always conflict with other robots. The arrows show the direction of movement of the robot at each position.}
\label{figu2}
\end{figure}

Saying that the object and the robot have the same position at the same time is not enough as a condition for an object to move. There should be a condition that specifies which of the robots that do not carry objects should move towards an object to carry it. For example, the concept of total distance between robots and objects can be introduced, which should be minimized for optimal scheduling of objects for the robots. Analogous to the existence of the final station (terminal), there should also be an initial station that has no inputs but produces several outputs (raw materials) that are used as inputs by other stations. Otherwise, these raw materials would have to be placed randomly in space. Moreover, since the proposed method does not impose any constraints on conflict avoidance, we cannot ensure that we will find a conflict-free route the second time. For example, there could be a node whose state in the route is conflict-free, but its state causes conflicts at other nodes, see Figure \ref{figu2}. This means that conflict avoidance should be added to the proposed algorithm in the first place. ISPS shows faster runtimes, in contrast to the expected results of CBS and NBS dependencies. The obtained runtime measurement is valid only if the algorithms are computed independently. However, this contradicts the proposed method which states that the algorithms must be executed with respect to each other. Since the goal is to minimize the makespan, there should be a result that gives the average makespan of each experiment.

\cite{Behrens:2020} described an optimization problem for assigning tasks to robots in a sequence of tasks and motions of robots such that there are no collisions between robots, the predefined set of constraints is satisfied, and overall makespan is minimized. In their method, tasks are divided into confined tasks (a task where a robot's action is limited to a small part of the workspace, e.g., grasping and placing) and extended tasks (a task where a robot's action is limited to a large part of the workspace, e.g., welding along a line). The main focus is on the optimization of the extended tasks. For a given set of tasks that satisfy the set of constraints during execution, each task has multiple starting positions (called degrees of freedom), and the time interval in which a robot can perform each task is measured. The workspace is partitioned to identify regions that a robot occupies during the execution of a task in a sequence of time frames by discretizing the time interval of the task into smaller successive intervals. For solving tasks and motion planning for extended tasks, the problem is translated into a constraint satisfaction problem, which is a type of optimization problem modeled with triples $(X,D,C)$, where $X$ is a set of variables, $D$ is a set of domains where parameters take values, and $C$ is a set of constraints. The solution is to assign values from $D$ to the variables $X$ such that the set of all constraints $C$ is satisfied. To solve the problem, the gradient method and steepest descent by Cauchy \cite{Goldstein:1962} is used (backtracking search method), that is, start with an upper bound on the minimum makespan and add the constraint that the minimum makespan is less than the upper bound. Then, remove the values from the domains that do not satisfy the new constraint, and then decrease the value of the upper bound until the optimal solution is reached for the values of the variables. The optimization model is viewed from three perspectives: task layer, robot layer, and collision-free plan. 

\begin{figure}[th!]\centering
\includegraphics[width=0.7\linewidth]{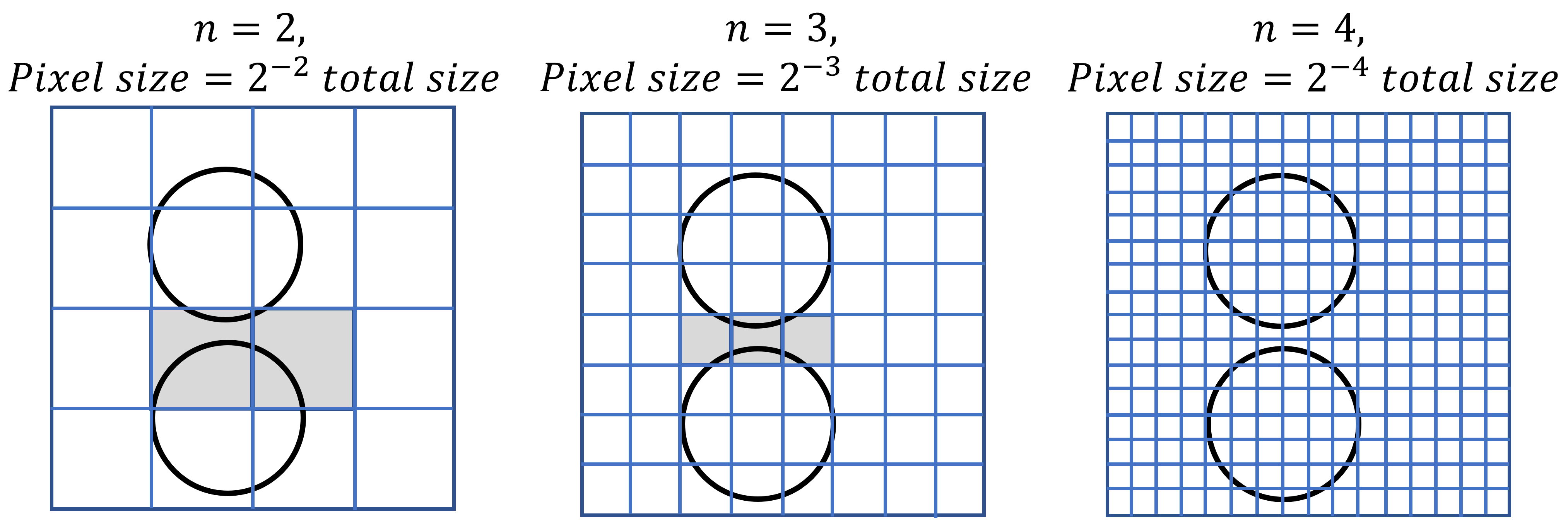}
\caption{Two robots space occupancy. With larger pixel size they have collision but with smaller pixels they do not have collision. The workspace is the large square. Pixel sizes are chosen as proportion of the total size with the factors $2^{-n}$ for $n=2,3,4$. The grey rectangles are pixels considered with robot collisions.} 
\label{figu1}
\end{figure}

The time intervals considered in the paper depend only on the task and the dependence on the robots is not considered, i.e., the robots should be identical. Moreover, the size of the region and the discretization of the time interval are not fixed and can be either a short or a long interval. Depending on the choice of the interval size and the size of the domain, different solutions can be obtained. Moreover, in the backtracking search method, in order to obtain a solution for the first upper bound, a random selection seems to be made in the solution space, so that the values in the domains that satisfy the constraints are selected. But the steps to reduce the upper bounds are not described. And the solution completely depends on the selection and reduction of the upper bounds. Moreover, different upper bounds may lead to different solutions. Also, the robot dependency is skipped but should be considered since the navigation codes, time intervals, constraints, and active components are robot-dependent, and their values may change when switching from one robot to another. In \cite{Behrens:2020}, tasks are translated into ordered visit constraints originally defined in \cite{Behrens:2019}. Here, the tasks are considered confined, so the start and end locations are considered identical, and the robot configuration is the same at the start and end. However, the new modified version includes different locations and different configurations to include extended tasks. Moreover, collisions may occur between components of a single robot, which is not considered. In addition, collision avoidance depends on the size of the regions (voxelization sizes). When the region size is large, the robots can have intersections in the configuration spaces, but when the region size is smaller, they have empty intersections, see Figure \ref{figu1} for a 2D example. This is consistent with the well-known result that there are always infinitely many other real numbers between two distinct real numbers, see \cite{Gaughan:1993}. This means that if the two robots are not connected at any point, there will always be a voxel size where the intersection of their voxalizations is empty.

\cite{Fu:2021} proposed a method to decide whether to assign subtasks to edge devices or offload them to fog nodes, such that the total execution time of all subtasks and the total energy consumption of all edge devices are minimal. In this method, the tasks that need to be executed by edge devices are translated into a DAG generated by the dependencies of the subtasks. Considering the average execution time and energy consumption of all subtasks on each core of the system architecture, the problem is then transformed into an optimization problem whose solution allows optimal scheduling of the subtasks. 


In this model, all edge devices must have a direct communication link with a fog node, all fog nodes have direct communication with each other at the same communication speed, and the initial and final subtasks of each task requested by edge devices should be executed on edge devices. Moreover, it is assumed that the transmission bandwidths between all edge devices and all fog nodes are identical. The proposed method does not minimize the energy consumption, but only considers a threshold for the total energy consumption of all devices. All subtasks should either be offloaded to fog nodes or executed on the same edge device that initiates the request. This makes it impossible to use neighboring edge devices to execute subtasks. The speed of data transmission and energy consumption for executing subtasks are assumed to be linear over time.

\cite{Lippi:2021} examined the allocation of tasks when humans and robots work together, taking into account the change in human performance over time, and proposed a method for minimizing the normalized makespan while maximizing process quality and minimizing the agents' workload. In the proposed method, the problem is transformed into mixed-integer linear programming (MILP). After the solution is found, the parameters are recursively updated after the completion of each task and checked for possible re-allocation. 


In scenarios that require multiple re-allocations, the optimization problem must be solved each time a task is completed. This is a time-consuming process, apart from the time required to update all parameters when the number of tasks assigned to an agent is large. Moreover, one of the criteria for re-allocation of tasks is to change the threshold for the change in cost from the original task allocation, although the threshold is not specified and there are no criteria for how to generate such a threshold. Their model assumes that at most two agents can perform a single task and only one human can supervise the execution of a task. There are scenarios where more than two robots need to execute a task and multiple human agents need to supervise the execution of a task. For these scenarios, the proposed method does not provide a solution. Moreover, the proposed method assumes that the finish time of a task is finite, which is not suitable for some tasks. Finally, some tasks are co-dependent and should be started and finished at the same time, but the formulation does not consider such tasks.

\cite{Bai:2022} solved the problem of distributing objects from their initial positions to their destination using multiple robots within a time window. The robots are assumed to have a limited capacity to transport objects, and the objective is to minimize the total travel time of all robots to transport all objects. The problem is then transformed into a combinatorial optimization problem where the total travel time of all robots is to be minimized. An auction-based distributed algorithm was used to solve the problem, which first finds a feasible solution when information about all objects and robots is available. Tasks are then assigned to robots based on the ratio between the expected cost of the task and the cost of the last feasible solution at each time step. 


It is assumed that the number of robots is very large to ensure the existence of a feasible solution. It is also assumed that the robots have the same capacity, and that robot failures, recharge times, and the energy required by a robot to complete a given task are not considered. In practice, however, the number of robots is limited, and it is difficult to find a feasible solution with a limited number of robots.

\cite{Fang:2023} studied task scheduling for a robotic network of single-task robots performing multi-robot tasks. They proposed a particle swarm optimization method for multi-objective task scheduling in which the task completion time, robot cost, makespan, and workload balance are minimized. To find a suitable solution, they use local particle updates in each step based on the strength of the target to find a neighboring region with best solutions.

The local search strategy may result in finding local optimal solutions. Moreover, the correlation between the targets is not considered since minimizing the makespan and load balancing are equivalent. Moreover, with a very large number of robots and tasks, the search for a solution is very time consuming. It is also assumed that the robots have a stable performance. They have tested the performance of their method with simulations.

\cite{Ye:2023} investigated the task allocation of a multi-station multi-robot welding system, minimizing the completion time of welding a single workpiece at a single station, the time difference between the completion times of adjacent stations, and the path length of robots moving to the station. They designed a multi-objective optimization problem that was solved with an evolutionary algorithm that sorts the stations and robots based on the expected completion time and the paths to the stations. Their goal is to assign tasks to the stations and then instruct the robots to move between the stations to complete the tasks.

They have formulated each goal individually, but the main goal of optimizing all objectives simultaneously has not been considered. In addition, the dependencies between the goals were not taken into account in the formulation. For example, if the distance a robot travels to a station is long, the time to weld the workpiece the robot is tasked with will also be long. In addition, the formulations only contain the times for the completion of the individual workpieces. However, after the completion of each workpiece, the station needs a certain amount of time to remove the completed task so that it is ready for the next task. Also, possible collisions between the robots and the need to redirect them while moving between stations are not accounted for in the formulation, which can affect the performance of the system. They tested the performance of their proposed method using simulations and in the real-world environment.

\cite{Yan:2024} investigated the task allocation and path planning of multi-UAVs using a genetic algorithm modified with crossover and mutation operators to meet the resource requirements of tasks with simultaneous target arrival. They also proposed an unlocking strategy to prevent the UAVs from remaining in an infinite waiting state. It is assumed that the UAVs fly at a fixed altitude and have a constant speed. Their goal is to find a task allocation strategy that regulates the relationship between targets and UAVs while finding a suitable path for the UAVs to their targets, since the repeated change of the UAVs' motion angle increases the computational complexity of the problem. They derived a combinatorial optimization problem that simultaneously minimizes the total flight distance of all UAVs and the maximum flight distance of any UAV. They took into account the limited resources carried by the UAVs, the resource requirements of the targets, the constraints on the energy capacity of the UAVs, the maximum number of UAVs that can be at a target simultaneously, and the possibility that multiple UAVs remain in the waiting condition to be assigned to some targets. In the latter case, the expected resources for each target are determined, sorted in descending order, and the UAVs are assigned to the targets with the highest order first.

To avoid the UAVs being in an infinite waiting state, it may happen that a task cannot be completed before switching to another task. For example, if tasks have deadlines and for a given number of UAVs, the total resources carried by all UAVs are less than the resources required to complete the task in some of the targets in the ordered target list, then the task will be incomplete in the next step, but the task will be ranked lower and its deadline may be exceeded before it is completed. They have tested the performance of their proposed method with simulations.

\subsection{With Cloud}
\cite{wang:2017} proposed a hierarchical auction-based mechanism to find the shortest communication time by removing unnecessary repetitive computation and reducing the communication cost in a robotic network cloud system. In the proposed method, network nodes are prioritized to allocate resources and maximize the overall transmission by managing the nodes' requests. 


The latency and memory usage by the robots are not considered and the architecture topology is considered fixed. Moreover, there is no fair comparison with state-of-the-art methods. All robots need to communicate with the cloud to perform their assigned tasks. The execution time of the tasks and the scenario of execution of simple tasks by robots that do not require communication with the cloud are not considered.

\cite{li:2018} proposed mixed-integer nonlinear programming to minimize the latency considering the dependencies of the algorithms in a robotic network cloud system. 


The memory required in the robots to execute the algorithms is not considered and the communication time is also not fully considered. For example, a robot may be able to execute two independent algorithms individually, but may not have enough memory to execute them in parallel. Furthermore, since all algorithms are requested by robots, the start time should be calculated from the time a robot sends a request to execute an algorithm until it receives the result of that algorithm. In this way, the start and end point of time initiation is the robot and in case of multiple robots, all robots should be the start and end points.

\cite{Chenc:2018} examined the urgent response to sudden demands on the cloud. The author proposed a method to find the minimum number of virtual machines and physical machines, and the minimum distance between virtual machines and physical machines, to ensure resource allocation optimization and timeliness. The cloud data center includes multiple physical machines with different number of CPUs, memory capacities, and disk spaces. Each physical machine runs multiple virtual machines with different number of CPUs, memory capacities, and disk spaces. The process of resource allocation is to place the virtual machines on the physical machines. When an urgent resource request arrives, the proposed method reorders the virtual machine queue based on the priorities, determines the resource capacities of the physical machines hosting the request, and solves the multi-objective model that finds the mapping of the virtual machines to the physical machines that accommodate the request. The priority of resource allocation is the normalized priority of users and urgent degrees of users' resource requests. The virtual machine allocation priority is defined as the weighted average of the normalized user priorities and the normalized urgent degrees of the users' resource demands. To define the multi-objective model, so-called distances (performance vectors) are formed between virtual machines and the physical machines hosting the request (physical host). The matching distance between a virtual machine and a physical machine is defined as the natural distance between the two normalized vectors. The main objective is the sum of all matching distances when mapping virtual machines to physical machines. This means that for a physical machine with known free factors (number of CPUs, disk space, and memory capacities) we need to assign the task to the virtual machine with the largest factor that is closest to the free factors of the physical machine. The next goal is to assign all requested virtual machines to a minimum number of physical machines. Then a genetic algorithm is used to solve the problem. 



The multi-objective model described in \cite{Chenc:2018} can be transformed into an associative commutative matching problem, which has been shown to be NP-complete, as demonstrated in \cite{Benanav:1987}. Moreover, the degree of load imbalance is compared only after the requested virtual machines are assigned to the physical machines, and the virtual machines are assumed to be running under full load. This shows that the best-fit method (a kind of greedy algorithm that selects a physical machine with the most free CPUs to allocate the new virtual machine) is better than their proposed method with the least degree of imbalance. 

\cite{singh:2019} studied load balancing, minimizing energy consumption, maximizing resource utilization, and maximizing security in a cloud data center. The authors proposed a multi-objective optimization approach. To find a solution, they randomly select solutions from the solution space and use the Pareto front for non-dominated solutions to move towards an optimal solution. 


The nature of security is a side-channel attack, different tasks require different energy consumption, and the cost of data transmission is not considered. Moreover, the number of requested tasks and their energy consumption do not play any role in the optimization and the solution depends on the initial allocation.

\cite{Liu:2019} models the total delay for task offloading and constructs an optimal task allocation based on it to minimize the total delay. In this paper, the data arrival model is called a Markov-modulated on-off process, switched discrete-time Markov-modulated Bernoulli process \cite{Trivedi:2016}, where the success probability changes with respect to a Markov chain. The data delivery to the Fog node is considered independent and identically distributed (i.i.d.), and the data offloading to the edge node or cloud center is a two-hop connection, roadside units (RSUs) are used for offloading the data to the cloud and edge nodes. If the computation tasks cannot be completed in the required delay time, the tasks are offloaded to other nodes (nearby Fog nodes, the Edge node, or the Cloud node). Assuming that the channel gain follows the exponential distribution with parameter $1$, compute the outage and transmission probabilities and then define the outgoing process using the min-plus algebra, which allows combining the two hops for offloading the data to the cloud node or the edge node with a single parameter. Then formulate the delay process and find the steady-state distribution for each hop. Finally, define data arrival and service (super) martingales. The optimization problems that depend on where the tasks are offloaded are extracted and solutions are provided. 


The problem is defined only for discrete time periods. And the data is split into multiple partitions and transmitted to other nodes for processing. When the partial data is received by the node, the node starts processing. The authors have not provided any explanation for cases where the input data needs to be collected from other nodes to start processing partial data, or when nodes have the condition that there is a minimum data size for processing partial data, and the optimal partitioning of data does not satisfy this condition. Moreover, the departure process is originally described as an infimum-plus algebra, which should really be called min-plus because it is not well defined otherwise. The defined threshold is locally defined and should be extended to a global threshold for all incoming data if it is greater than any value of the service. For a task only one source node is considered, and for multiple source nodes, we have to use Medium Access Control (MAC) protocol which is not an easy task because for a single source node the data shift is considered independent of the source due to the formulation but for multiple source nodes it depends on the source due to MAC protocol, bandwidth usage is shared and two source nodes can be adjacent and are in the set of fog nodes of each other. The delay from the RSU to the edge node is assumed to be constant. However, when a vehicle moves, the data may be transmitted to the edge node through different RSUs. Therefore, the distance from the RSU to the edge node may change, so the propagation delay may not be constant. The wording suggests that the RSUs are always closer to the Edge than to the cloud, which does not include the case where a vehicle moves close to the cloud server, so the opposite is true.

\cite{Geng:2020} developed a method for simultaneously minimizing resource utilization, time, and cost and performing load balancing. The authors describe a multi-objective optimization, formulate all the objectives and use the hybrid angle strategy \cite{Tseng:2006} to find the optimal solution. And for load balancing, the objective of overloading was carried out. 


The method hybrid angle strategy combines ant colony, genetic algorithm and local search method to find an optimal solution, but it needs much computation time. Moreover, the method can only be applied to discrete optimization problems. If we apply the method to a continuous and non-convex problem by discretization, depending on the discretization, the solution is either weak or it is extremely time consuming to find a solution due to the large population size.

\cite{Dang:2021} proposed a scheduling method in which the fog layer handles all the tasks by distributing and balancing the loads among all the fog nodes and reducing the delay. The authors translate the problem into an optimization problem that minimizes the service delivery delay (the time interval between when the fog receives a request and when the IoT node that sent the request receives a response). 


Consider a system that contains three layers: IoT, fog, and cloud layers to provide IoT services. Each fog node has information about its neighboring nodes, and the new task first reaches a fog node that is closest to the IoT. In this case, if the task is received by a fog node that has better specifications (meaning that minimum requirements to be able to perform the task) compared to all its neighbors, the task is assigned to it. PSO (particle swarm optimization) is used to find the suboptimal solution of the optimization problem. In this case, a schedule solution with expected delays is predicted and each time a new solution with a lower delay is found, it is updated. If the difference of average delays in two consecutive steps does not change more than a certain threshold, the optimization process is stopped. However, suppose that there is a fog node that has better specifications than the original fog node (the node that receives the request for the task), but this node is not the neighbor of the original node. In this case, the task is assigned to the initial fog node, but there is a fog node where assigning the task to it can actually improve the performance. So, the proposed method can find the optimal assignment solution only when all the fog nodes are adjacent to each other. Moreover, in most existing methods, the tasks are distributed to the nodes that have the required resources and reliable communication. For task orders, two orders are considered, namely, one task should be completed before another task starts, and two tasks should start simultaneously; other orders such as tasks to be completed simultaneously, tasks to be executed simultaneously, etc. are not considered. Moreover, the simulation is performed for a request rate (number of requests sent by the IoT per second) between $0.01$ and $0.05$, which is a very low number; in a busy network, the values are higher. Moreover, the communication between the cloud and the fog nodes is very high, which leads to extreme delays compared to the AFP method, where the cloud is used as the central unit for scheduling tasks on the fog nodes.

\cite{Chen:2021} investigated dynamic task scheduling in robotic network cloud systems. This study considers whether a task can be performed by a single robot or a group of robots, or whether it should be moved to the cloud, depending on the characteristics of the task. Their goal is to optimize the quality of service by optimizing the latency, energy consumption and cost considering the specifications of the architecture and the tasks. They developed a mixed-integer optimization problem and solve it using a heuristic method. 

The developed model does not take into account the interdependencies between latency, energy consumption and costs. Moreover, it is necessary to determine in advance which task should be performed by a single robot or a group of robots or moved to the cloud. The performance of the proposed method was tested using simulations.

\cite{Casini:2021} proposed a technique for splitting and partitioning tasks to improve the dynamic scheduling of tasks on multiprocessor systems (e.g. robotic network cloud systems). Then they proposed the technique of load balancing and limited workload redistribution to increase the acceptance of future workloads. Their goal is to schedule more tasks and complete them faster, taking into account the time windows of the tasks and minimizing the latency of the tasks. For task splitting and partitioning, they used the algorithm proposed in \cite{Burns:2012}. They translated the problem into an optimization problem and then obtained a solution by simplifying the problem, setting a lower bound on the latencies of the tasks and solving the dual problem. For load balancing, they either heuristically find a suitable processor for the task or split it into several serial subtasks and consider them as a stream of tasks. 

The communication time is not considered and it is assumed that the processors have a constant performance and that there are no delays in the execution of the tasks, which increases the latency. Also, splitting the tasks into smaller subtasks leads to additional delays in starting the successive subtasks. In some cases, the delay is caused by the execution of a subtask by another processor. This processor must communicate with the processor responsible for executing the previous subtask and also with the processor responsible for executing the next subtask. The performance of the proposed method was tested using simulations.

\cite{Ours:2022net} developed a task assignment algorithm for robotic network cloud systems that achieves optimal performance for a given task set by simultaneously minimizing the memory usage by all robots and the completion time of all tasks by the robots. The problem is transformed into a multivariate optimization problem and the solution can be obtained using a branch-and-bound algorithm. 


The method can only be applied to systems where the complete information about the algorithms, such as the size of the processing, input, and output memory and the space complexity of the algorithms, as well as the average execution time of the algorithms on each processing unit and the average communication time between the processing units, is given. The cases when a robot does not work properly or the architecture of the robotic network cloud system is dynamically changed are not considered in the study.

\cite{Yin:2024} developed a task scheduling mechanism for cloud-edge computing for production lines that minimizes service latency and energy consumption. They translate the problem into a multi-objective optimization problem and solve it using PSO and gravitational search algorithm (GSA), which helps to consider the neighbors of the previous solutions in the last step. In their formulation, they assumed that the tasks are independent, but in the real world, most tasks are interdependent, which means that a task can only be performed if all other tasks on which it depends have been completed first. They tested their method with simulations.

All contributions that formulate the problem as an optimization problem are summarized in Tables \ref{tabo1} and \ref{tabo2}. Optimization methods are effective in minimizing resource usage in well-defined environments, but often struggle with computational complexity in dynamic settings. Therefore, hybrid approaches that combine optimization with real-time heuristics or machine learning can help reduce complexity and adapt to changing conditions.

\section{Combinatorial}
In this section we discuss contributions that solve the problem using a combinatorial approach. The works are separately analyzed based on whether the cloud infrastructure is considered or not.
\subsection{Without Cloud}
\cite{Chopra:2017} proposed an extension of the Hungarian method, \cite{Burkard:2012}, by considering a distributed version of the method that allows a team of robots to cooperatively compute the optimal solution to a linear objective function without requiring a coordinator or shared memory.


There is no fair comparison with the state-of-the-art methods and in case there are two suitable matches between robots and tasks with the same cost, the proposed method does not explain which is the decision criterion to find the most suitable match. Moreover, the solution is obtained by successively finding a match between robots and tasks. However, depending on the match, the convergence may be very slow, and a load balancing procedure is required to avoid the case where a single robot is assigned multiple tasks, but some robots have no tasks. In a more realistic scenario, not all robots are capable of performing all tasks, and the method requires categorizing robots by tasks they can perform.

\cite{Chen:2019} studied a rescue mission that maximizes the number of rescued, minimizes the average waiting time, and minimizes the total path cost. In this method, tasks are clustered and robots select tasks from their preferred clusters after task assignment is completed. A proportional selection strategy is used to avoid local optimum. The market-based approach is used to find an optimal assignment and it allows adding or removing robots as well as removing, modifying and adding tasks, which allows rescheduling of tasks. 


Clustering and proportional selection depend on the initial metric used, and different metrics may lead to different results. For example, if the metrics used for clustering and/or proportional selection are small or the number of robots is large, all robots spend most of their time validating their optimal scheduling and communicating with each other, which may result in some of the tasks being removed due to their deadlines, since robots only have information about their own scheduled tasks.

\cite{Lu:2019} studied a method for task scheduling for heterogeneous computing. The authors proposed Lookahead to prioritizing tasks and selecting processors based on a prediction cost matrix with an algorithm called PPTS.


The comparison is made only with PEFT, \cite{Arabnejad:2014}, which uses the optimistic cost table to obtain the priority of each task and assigns each task to a processor, and HEFT, \cite{Hariri:2002}, which sorts tasks in decreasing order of priority and assigns the highest priority to the processor with the smallest earliest completion time. The method is not compared with IPEFT, \cite{Zhou:2017}, the improved version of PEFT, which uses the pessimistic cost table to obtain the priority of each task and uses the critical node cost table to assign each task to a processor, which provides a better comparison with state-of-the-art methods. Moreover, the metrics used to compare different methods are not suitable to compare them. The metrics defined over each method measure different properties, and the PPTS method has additional properties to measure (computation time successor of a task) than the other methods. So to compare them, we should restrict the metrics to the common properties that all metrics measure and then compare them. Otherwise, the comparison is not valid. General terms for comparing different metrics can be found in \cite{willard:2004}.

\cite{WANG:2020} studied minimizing the maximum travel times of collaborating robots, i.e., minimizing the completion time of the longest task. The authors translated the optimization problem into a matching problem with minimum edge weight sums. 


They use matching instead of perfect matching, which exponentially increases the solution space. Deterministic task scheduling is used, and the method fails to schedule the model with uncertainties such as robot failures, delays, and other uncertainties. Moreover, transfer robots that are not assigned to any task are considered $\Idle$ since the problem is deterministic. If these robots move to the centroid of all remaining tasks instead of being $\Idle$, the completion time of all subtasks may be faster. Otherwise, it might appear that only certain robots are used to transfer tasks because of their proximity.

\cite{Orr:2020} has studied the scheduling of tasks with duplication. Task duplication means scheduling a copy of a task on different processors. An advantage of task duplication is that it can improve performance due to the reduction in communication cost. The authors have proposed a task scheduling method with communication delay where tasks are in a precedence relationship, communication cost is incurred between processors, and the task can be scheduled on multiple processors. Their goal is to minimize the overall completion time. The tasks are translated into a DAG, and the graph of the architecture is extracted. First, the tasks are partitioned, and the tasks in each partition are assigned to a single processor (using branch-and-bound search to find an optimal solution). A duplicable task is defined as a task with an out-degree of at least 2, or a task that has a descendant with an out-degree of at least 2. It is then shown that duplicating the set of all duplicable tasks can reduce the completion time. Duplicable tasks are duplicated on the processors to which their ascending tasks are assigned to see whether or not this improves completion time. 


If the DAG of tasks is complex and most tasks have an out-degree of at least 2, the computation becomes exponential, and all duplicable tasks should be tested by trial and error to determine which tasks should be duplicated. Furthermore, the set of duplicable tasks only provides the set of tasks whose duplication can improve the completion time, but it does not specify exactly which tasks should be duplicated. Moreover, in the case of non-duplicable tasks, the problem becomes a static allocation, where the complexity of finding an optimal solution increases exponentially for a large number of tasks.

\cite{Hari:2020} proposed a method to find the target tasks of all robots in such a way that the maximum mission time is minimized and the scheduling constraints for human operators are met. A method called task assignment, sequencing, and scheduling is developed for a team of human operators and robots, where the robots travel to the tasks and work on the tasks together with the human operators. The goal corresponds to the generalized traveling salesman problem, which is to find a sequence of tasks for each robot and schedule them for the human operators such that each task is visited exactly once by some robots. Each task can be performed jointly by only one robot and one human operator, and each human operator can work on at most one task, and each task can be scheduled for at most one human operator. The mission time of a robot is calculated as the sum of the travel time, the waiting time, and the processing time of the target tasks of that robot. The waiting time comes into play when the number of human operators is less than the number of robots and some robots have to wait until a human operator is available. A $\alpha$ approximation solution is proposed as the initial solution, where the $\alpha$ ratio is a constant factor such that the cost of the approximated solution is at most $\alpha$ times the optimal cost. Then other methods like Branch and Cut are used to obtain a better result. The proposed algorithm consists of three steps: changing the travel cost of a robot between two tasks by adding half of the processing times of the two tasks, assigning the goal order for each robot according to the algorithms in \cite{Frederickson:1976}, first applying Crane's algorithm to find the $1$-tour (a matching algorithm to find the edges with the minimum travel time, then another algorithm to expand to a cycle, again another algorithm in that cycle to replace some paths with the shortest ones if there are any, and then another algorithm to split the $1$-tour into disjoint $k$-tours), and then apply the $k$-splittour algorithm to split it into $k$-disjoint tours for all robots, and a greedy heuristic implementation to create feasible schedules for the robots to go to their respective task sequence in the same order and either wait until a human operator is free or process the task with an assigned human operator. 

\begin{figure}[tbp]\centering
\includegraphics[width=0.25\linewidth]{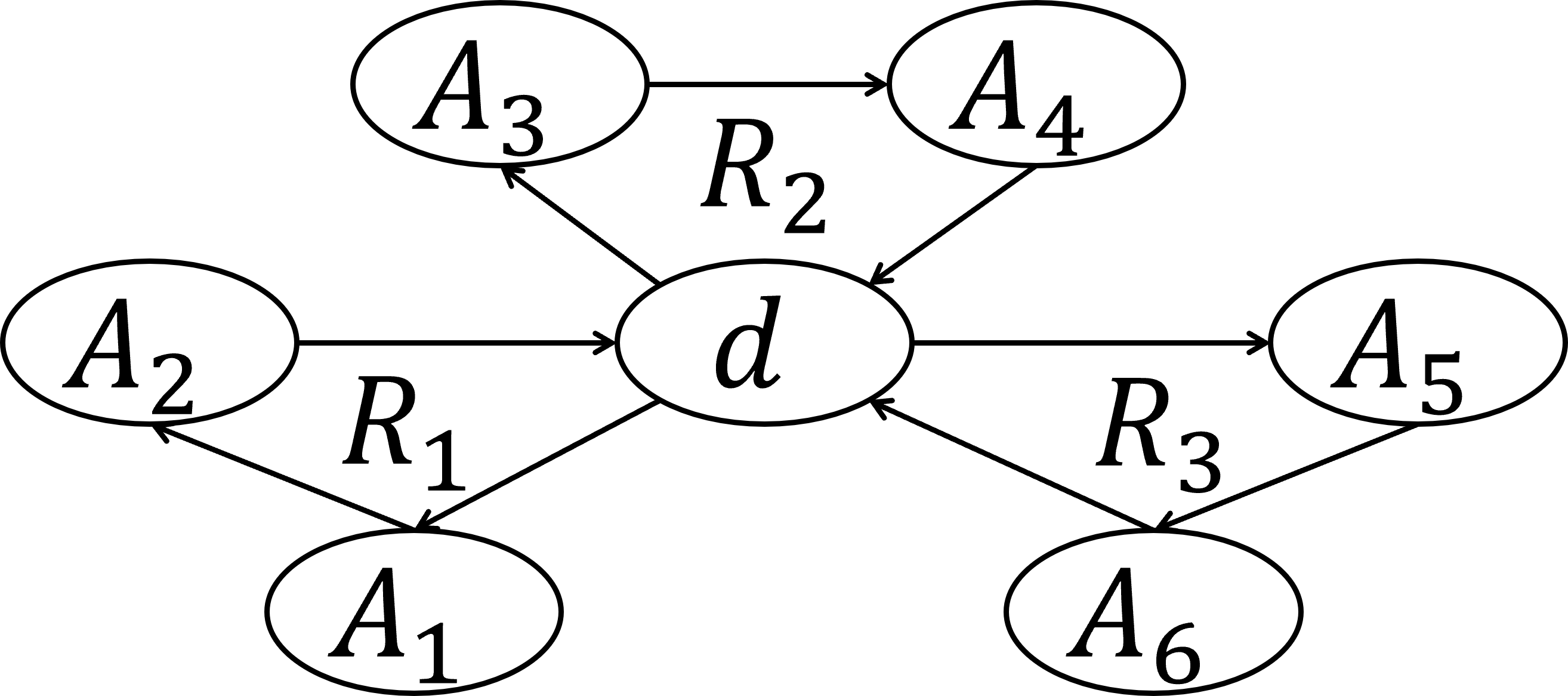}
\caption{Example of tours (3-cycles) of three robots with identical processing and travel times. $d$ is the initial point of all robots.}
\label{figu3}
\includegraphics[width=0.8\linewidth]{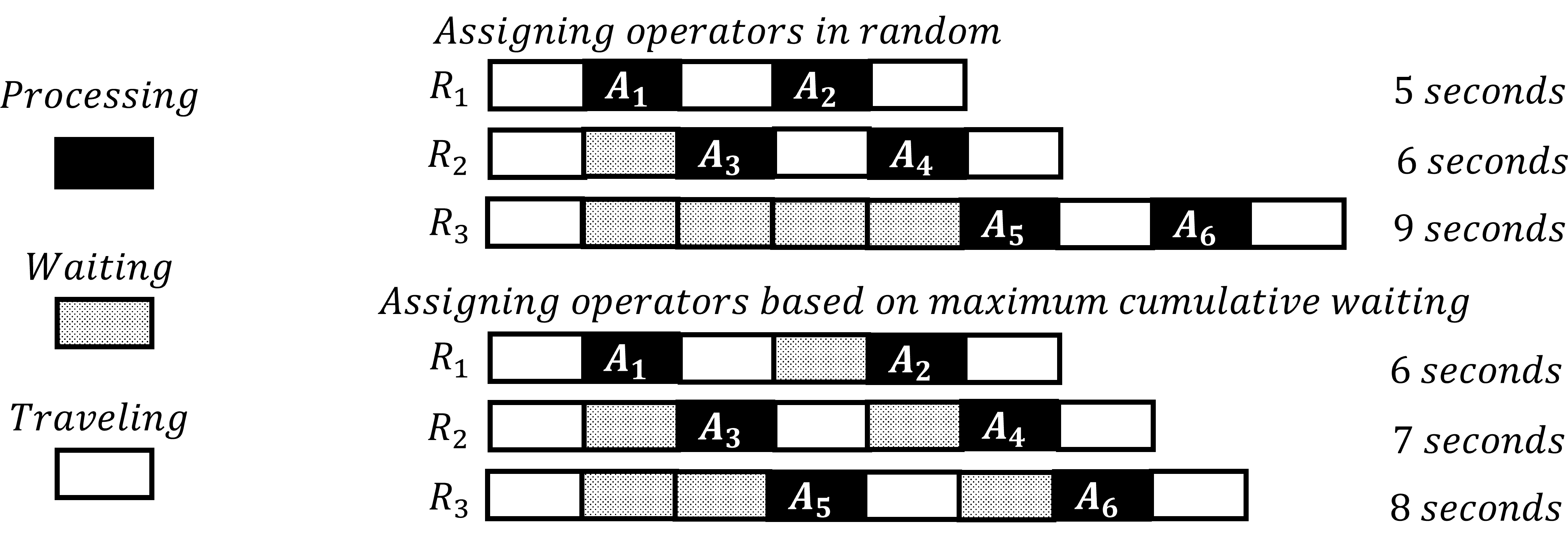}
\caption{Assigning operators to the waiting robots in random, as proposed in \cite{Hari:2020}, against assigning to a waitng robot with maximum cumulative waiting.} 
\label{figu4}
\end{figure}

The authors did not consider minimizing $\Idle$ times (the time an operator is without scheduled tasks). Also, the ratio $\alpha$ is considered as a constant in the interval $2\leq\alpha < 3.5$, which is a relatively large value. In this case, if a human operator is available for more than two waiting robots, the human operator is randomly assigned to one of the robots, since we have no control over the waiting time of each robot and a robot may wait much longer before moving to the next task due to the random selection, see Figures \ref{figu3} and \ref{figu4}. Moreover, the number of conditions that need to be tested grows exponentially. And we only need some constraints on the edges of the graph to reduce the computation time. Therefore, the computation time of the proposed method is exponential in general, which can be found polynomially only for some limited cases, but not in general. Moreover, since the travel time is the time from the current position of the robot to the next task, they must mention that the travel time from one task to the next is determined by the order of the tasks; otherwise, we can consider the time that the robot returns to its initial position before moving to the next task, and thus the travel time increases. Instead of randomly assigning available human operators to the waiting robots, it is possible to obtain a better result by assigning the available human operators to the waiting robots with the maximum cumulative waiting time.

\cite{Malencia:2021} has studied a fair redundant assignment of agents to tasks that improves task performance. The proposed method consists of translating tasks and agents into a bipartite graph whose edges are weighted by the cost of each task assigned to an agent. The redundant assignment problem is about which task should be assigned an additional resource. The problem is transformed into an optimization problem, and an attempt is made to solve it. The solution is near-optimal in the sense that some of the constraints of the optimization problem are relaxed to find solutions faster. It is assumed that there are multiple robots, that each robot can do all tasks, and that multiple robots can be assigned to the same task. A redundant assignment is fair if it optimizes the worst-case task cost (i.e., the expected worst-case task cost). Consequently, fairness becomes a minimax or maximin problem. Since minimax problems are combinatorial and finding an optimal solution is infeasible for most redundant assignment problems, supermodularity properties are used to find a near-optimal solution that can be solved with a greedy algorithm. This means that adding an element (robot) to a solution set (superset) increases the cost. Since supermodularity corresponds to an increasing (decreasing) negative (positive) function on its domain, which is positive or negative depending on the definition of the cost function, we can use the bisection algorithm to find the solution. The threshold assignment algorithm, \cite{Glover:1986}, is used to obtain the solution. The edges are assigned values of 1 and 0, respectively, if the edge is considered as an assignment or not. The method of identifying the values of edges uses a threshold function. This threshold function is a criterion for the selection of edges. It can be the cost of the shortest path, the cost of edges, etc. 

\begin{figure}[thp]\centering
\includegraphics[width=1\linewidth]{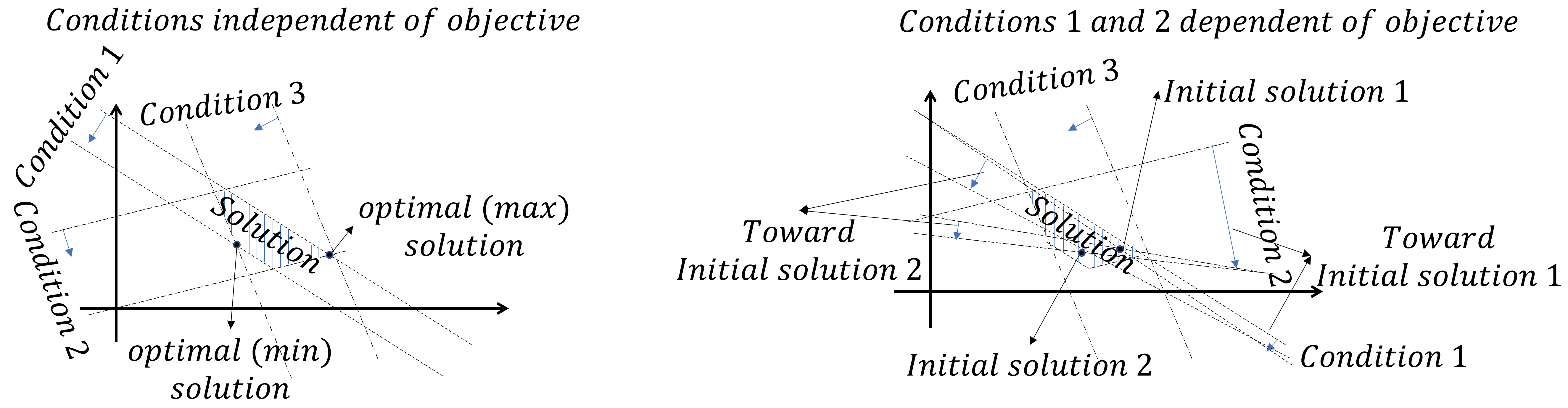}
\caption{Intuition for solving a linear optimization problem with constraints independent of the main objective and with two constraints dependent on the main objective. On the left-hand side, the constraints move linearly toward the optimal solution. On the right-hand side, the constraints move jointly toward an initial solution depending on the initial solution; this movement is not necessarily linear.}
\label{figu7}
\end{figure}

The optimization problem requires a constraint on the assumption that tasks should be assigned to at least one agent. And it is not specified which thresholds are used and which criterion terminates the application of the recursive thresholding algorithm. Moreover, the randomly generated bipartite graph is not sufficient for the simulation used. It must be noted that all task nodes have a degree of at least $1$; otherwise, the generated graph cannot be a suitable graph since some tasks cannot be assigned to any agent. In the proposed algorithm, an assignment is initialized and the feasible solutions replace the initial solution every time instead of being added to the initial solution. The next simulation uses the normal distribution, which is not validated. To find a near-optimal solution, the bounds of the relaxing parameter and the main objective must be independent. Independence should not be a necessary condition to obtain a solution. However, it is beneficial because generating the relaxing parameter independent of the main objective unifies the constraints of the optimization problem, and all constraints can be treated independently. This means that the change in the subspace generated by each constraint is independent of the subspaces generated by all other constraints. This makes finding the solution easier, see Figure \ref{figu7}.

\cite{Sahni:2021} proposed a mathematical model for joint offloading of multiple tasks that takes into account the dependencies between subtasks and schedules network flows to minimize task completion time. Network flow dependencies are the problem in offloading tasks to multiple devices, leading to competition for bandwidth usage. In the proposed model, the problem is translated into an optimization problem and a solution called Joint Dependent Task Offloading and Flow Scheduling Heuristic (JDOFH) is proposed. The method considers the DAG of tasks and the start time of network flows. When a new task arrives in the system, the task is offloaded by multi-hop communication with other units based on their resource availability to find the best node (processing unit) that can perform the task optimally. Thus, the underutilized resources can be found to distribute the tasks in a better way. However, this leads to an additional problem due to bandwidth constraints that affect the performance of the network. The task offloading problem with task dependencies that minimizes the total completion time of all tasks is studied, i.e., the problem of offloading tasks together (tasks consist of multiple dependent subtasks) and scheduling network flows to transfer data between dependent subtasks. For a given task, decide on which device to execute each subtask of the task and find (schedule) the start time of the subtasks. The architecture and set of tasks are translated into a simple graph with communication links and a DAG with dependency links between subtasks. The main idea is to use the set of execution flows and decide the scheduling of each subtask based on the order of execution flows. Subtasks in the same co-subtask stage are prioritized. Their priorities are recursively defined by the ranking metric for tasks in each co-subtask stage. The method is compared with 
\begin{itemize}
\item Local Execution (LE): tasks are executed on the device that generates the task; 
\item Remote Execution (RE): tasks are considered as a single unit and executed on the device with the lowest task completion time; 
\item Separate Task Offloading and Network Flow Scheduling (SOFS): subtasks are offloaded to devices using the algorithm HEFT, where in the algorithm HEFT subtasks are sorted in descending order of priority, and the highest priority is assigned to the processor with the smallest earliest completion time, \cite{Hariri:2002}. Then the network flow is prioritized by the earliest deadline first; 
\item Joint Scheduling Based at Task Release Time (ALT): it is similar to JDOFH and optimizes task offloading and network flow scheduling together. The only difference is that subtasks at the same level have no rank (order), so subtasks at the same level have no priority to figure out which one should be assigned first.
\end{itemize} 

It is assumed that no two flows are allowed to pass through a link at the same time. However, under this assumption, the full capacity of the bandwidth cannot be used. For example, if two tasks need to be transmitted from one device to another, which together are smaller than the bandwidth capacity, their simultaneous transmission has more advantages than their separate transmission. Moreover, various notions such as the rank of subtasks of a task, the cosubtask, the total data load on a device, and the average processing speed of all devices are not defined. The priorities of subtasks are defined based on the maximum processing speed, but not all subtasks can be scheduled on the processor with the highest speed, which means that the defined metric is not well-defined. Even if we take the average, the order is not well-defined because scheduling subtasks on the processor with the lowest or the highest speed requires a different order due to the different completion times of the subtasks. Moreover, in the experiments for LE, without testing, the best performance is shown when the arrival rate of tasks on all devices is identical and all devices are the same, or for the scenario where the devices are different but the arrival rate of tasks for all devices is the same and proportional to the processing speeds of the devices, because in these cases we do not need to consider the communication time. Moreover, task completion in RE may take more time compared to JDOFH because tasks are considered as single units and are completely assigned to one device instead of assigning subtasks to multiple devices. Also, because SOFS uses the HEFT algorithm first, communication is not fully considered and network flow is not minimized because the devices selected to perform subtasks may take more time to communicate because they are farther away from the device that generates the task. This means that without testing, we can say that the SOFS method has poor performance when the number of tasks increases. Moreover, ALT is similar to JDOFH with fewer constraints. Therefore, the completion time of tasks that apply ALT is always greater than or equal to the completion time of tasks that apply JDOFH.

\cite{Fusaro:2021} proposed a method to minimize the cost of allocating tasks to a human-robot team by translating the problem into a behavioral tree, dividing the tasks into a series of parallel tasks. Given a set of tasks that an agent can perform, the MILP problems are solved recursively to minimize the cost. 


In the proposed method, it is necessary to know the actual cost of performing the tasks by humans. And the solution space grows exponentially with the number of agents and the number of parallel tasks. Moreover, multiple human agents cannot perform the same task, and if a human agent have to perform the same task multiple times, the performance of human agents needs to be monitored, which is not considered. Moreover, there are no clear constraints to distinguish between human and robot agents. And there are no comparisons with the existing methods.

\cite{Jin:2022} study the competitive behavior of multi-robot coordination using the $k$-winer-take-all (kWTA) algorithm and analyze the behavior of the $k$ largest competitors. They investigated the convergence of the kWTA algorithm with adaptive gain activation function in a dynamic task competition scenario of multi-robot systems with communication including uncertainties and disturbances. Their model describes a distributed kWTA with time delay. The problem is then translated into an optimization problem to minimize the competition rate (difference between the square of the outputs and the success rate of the inputs). As time progresses, the maximum delay, the winners, and the outputs are updated via communication links. All winning robots continue to operate, but all losing robots are shut down immediately. Instead of actual or expected time delay, an upper bound was used where convergence of kWTA is achieved when the delay is less than the upper bound.

\begin{figure}[htp]
\centering\includegraphics[width=.65\linewidth]{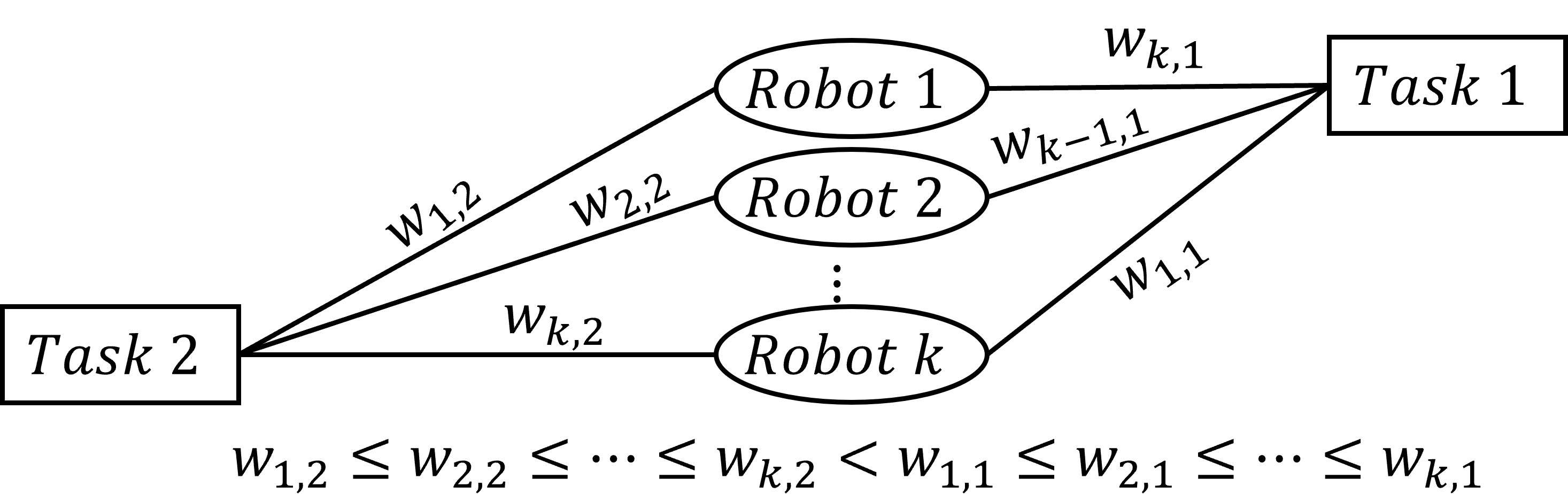} 
\caption{\label{figwin} An example where all robots are winners for two tasks with $k$ robots. The $k$ winning robots move towards the first task, and if one of the robots is moved towards the first task, the winning rate for the second task decreases, and vice versa. Even if several identical robots have the same distance to a task, there should be a tie-breaker to determine the winning robot.} 
\end{figure}

Such delays may depend on communication instabilities and robot malfunctions where the time delay may exceed the upper bound. For multi-task robot systems, the competitive scenario applies only to a single task. For multi-task robot systems with multiple tasks, there should be a parameter (tie-breaker) that prevents the scenario of at least two robots winning the same task. For a system with a finite number of robots, there should be a dynamic for the value of $k$ such that the number of robots is less than or equal to the initial $k$ at which all robots are winners, otherwise moving the winning robots toward one task may decrease the winning rate for other tasks, which in turn may increase the energy consumption of the robots and the total distance traveled. However, if $k$ strictly decreases with time, the number of winning robots that can complete the task decreases, see Figure \ref{figwin}.

\cite{Ours:2023} investigated static task allocation for robotic network cloud systems, which minimizes the total processing time of all tasks by duplicating tasks. They proposed a combinatorial graph-theoretic approach based on the precedence order of tasks that recursively determines which tasks should be duplicated and to which node of the architecture the duplicated task should be assigned. Their goal is to minimize the time in which the system performs all tasks such that it is faster than if there are no duplicates. Then they proposed the optimization, \cite{Ours:2022net}, to be solved for all nodes of the architecture using the branch-and-bound algorithm to determine the solution to the task duplication problem faster. They used simulations to find out how the proposed method works.

Although the time complexity of the proposed method is polynomial, it is still very time consuming as the number of nodes and the number of tasks increase. This means that time is needed to determine the optimal task allocation with duplication before deploying the architecture, and this time is much larger compared to solving the task allocation without duplication. In addition, the memory usage of the robots was not considered, since duplicating tasks for the robots increases their memory usage, which increases their cost and at the same time may decrease the performance of the robots.

\subsection{With Cloud}
\cite{Fan:2019} proposed a method to reduce the makespan and maximize the resource utilization using a directed acyclic graph and predicting the completion time of a task. They first generate the graph of all tasks, then randomly pair tasks with processors based on the earliest start time of the tasks and determine the execution time of the tasks. Based on the generated execution time, they then determine the makespan of the entire graph and then select the best pairs that minimize the makespan. 


As the number of tasks increases, the complexity of the algorithm increases exponentially. And the precedence order defined by the graph of tasks is ignored in the algorithm.

\cite{Yu:2020} proposed a method to balance the waiting time for scheduling clusters, a model to find the dependency correlation measure to find the similarities between tasks by their data dependencies. A clustering method is used to clusters tasks to reduce the number of tasks and the waiting time for scheduling. Then the next task can be started only after the previous tasks are completed. Load imbalance occurs when two parallel jobs at the same level are not completed at the same time. A dependency imbalance occurs when a late task needs the results of two tasks in a different order. Then, the late task must wait until both jobs are completed before it can be started. The clustering graph is defined as an extension of DAG by replacing the nodes with two sets of nodes containing tasks and a virtual node describing the delay time for extracting tasks in each job. Parallel tasks or sequential tasks based on dependencies in the extended graph are clustered as a single job. Then, all virtual nodes in a cluster are replaced with a single node representing the clustering delay, which is the difference between the execution of the job and the actual execution of all tasks in the job. Each cluster is then assigned to a single processor. Runtime imbalance and dependency imbalance are measured and minimized to find an optimal clustering method. It is concluded that the process execution time is least when the number of clusters is equal to the number of processors. In the proposed method, the number of clusters is known in advance and finding the maximum dependency correlations is the most important part of the algorithm. 

\begin{figure}[tbp]
\centering
\includegraphics[width=0.6\linewidth]{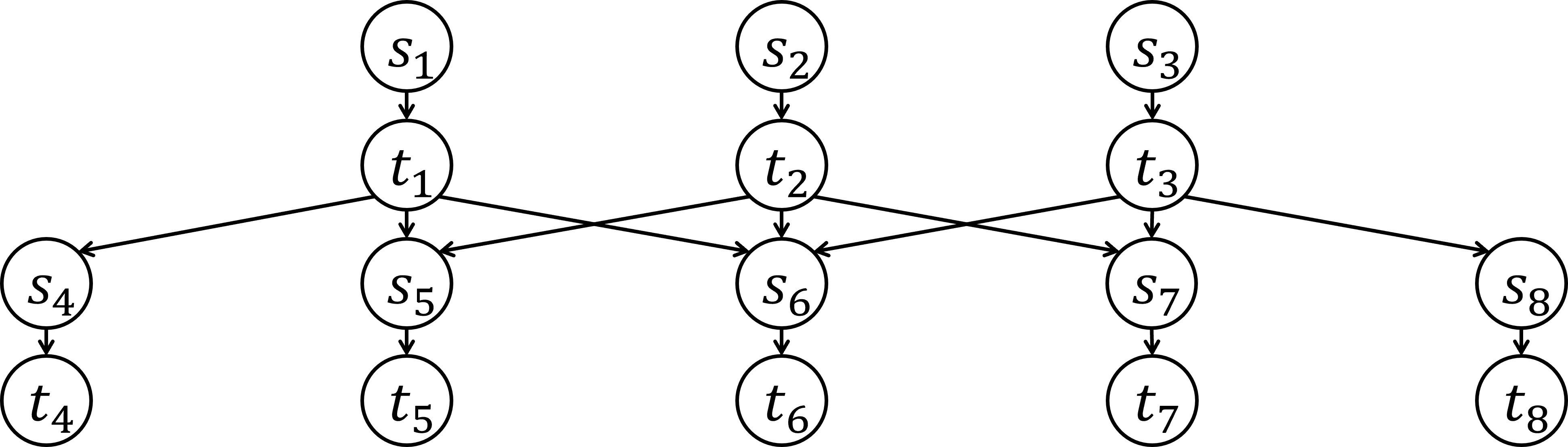}
\caption{An example of $DAG$ with $(t_1,t_2)$, $(t_2,t_3)$, and $(t_1,t_3)$ have maximum dependency correlation.}
\label{figu10}
\includegraphics[width=0.4\linewidth]{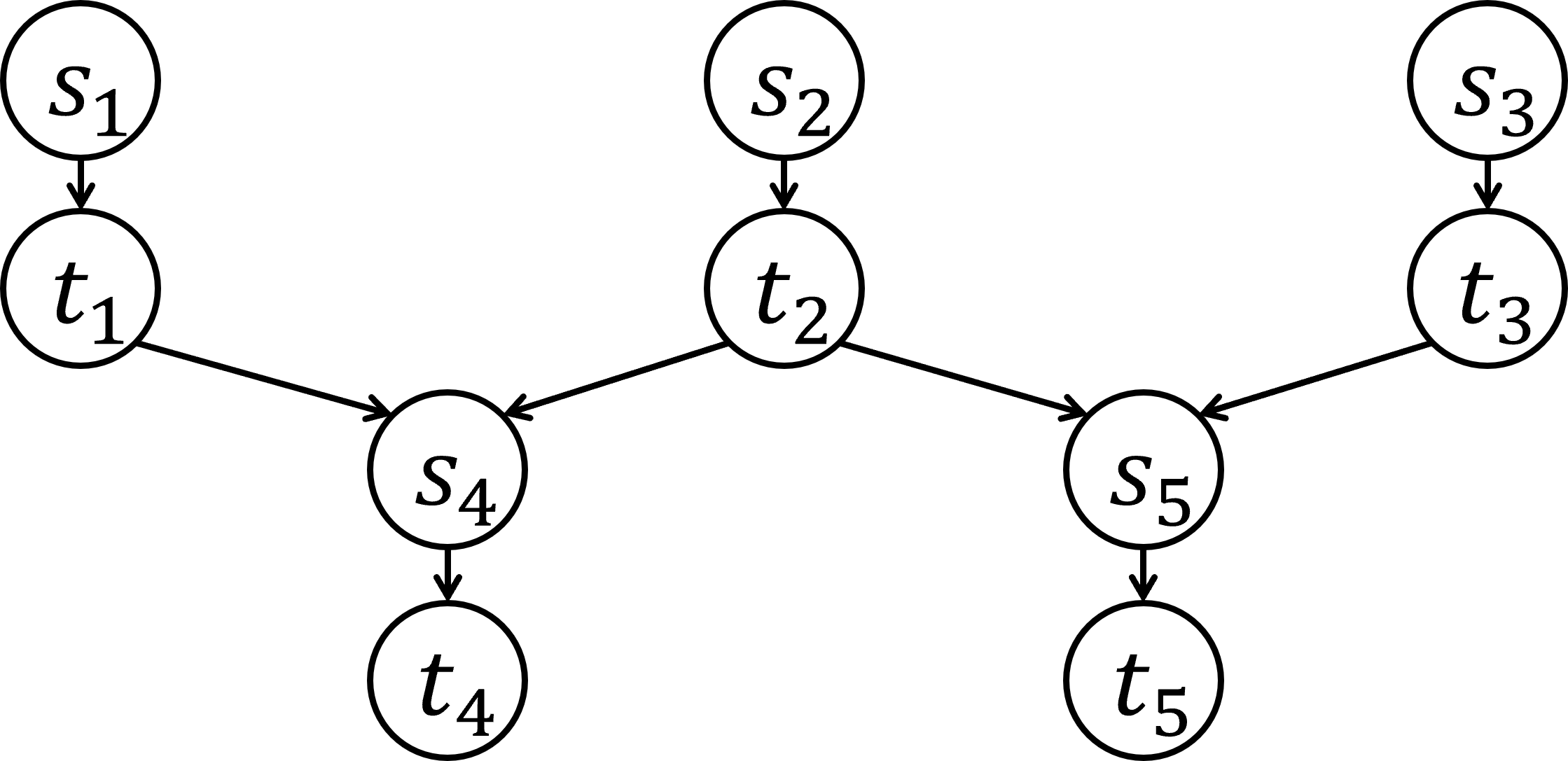}
\caption{An example of $DAG$ with $(t_1,t_2)$ and $(t_2,t_3)$ have maximum dependency correlation.}
\label{figu11}
\end{figure}

For the cases where the execution times of the tasks are assumed to be identical, see Figures \ref{figu10} and \ref{figu11}, the method does not determine which two tasks should be considered for the first cluster. It also concludes that the number of clusters is independent of the graph, which does not seem to be correct. This is because if we consider, for example, $30$ tasks in a sequence and say $5$ processors, then the only cluster with the minimum execution time is chosen (the communication time between processors is omitted in this case). In this case, the optimal performance can be obtained by considering a single processor to avoid the delay caused by the communication between processors. It cannot be said with certainty that $5$ clusters are the optimal number of clusters. Suppose that there are $k$ sets of tasks, where each set consists of serial tasks and the tasks in the different sets are parallel. Let us further assume that $k_1 > k$ are identical processors and the total execution time of each task of any set on each processor is constant, then according to the conclusion, $k_1$ clusters are required, which means that by the pigeonhole principle some tasks from different sets should be assigned to different clusters, which means extra communication time. However, if we consider $k$ clusters, the extra delays due to communication of data are eliminated, so we have a shorter execution time. This contradicts their conclusion.

\cite{Zheng:2020} proposed a priority-based (PB) scheduling method to maximize the parallelism of ready tasks that need to be scheduled for a distributed computing system by considering the precedence order of tasks. It gives priority to nodes based on execution time and their interdependencies. Define the priorities of nodes based on four terms Direct Quotient (DQ) (it is the number of tasks ready after the current node is completed and the nodes with the highest DQ should be executed first), Level Quotient (LQ) (it is the maximum length from this node to a sink node, and the node with the highest level should be executed first for a given set of ready tasks), Export Quotient (EQ) and Import Quotient (IQ) (they are defined recursively). IQ is only used to calculate EQ and EQ is used to determine the importance of executing a task for unlocking other tasks. Then the priority of the two tasks can be determined by first comparing their DQ and LQ. And if they are equal or give opposite results, their EQ is used as tie-breaker. If all the values EQ, DQ, and LQ are equal for both the tasks, the winner is selected randomly. Every time a task is scheduled, the set of ready tasks changes, so we need to update the values of DQ, LQ, and EQ. 


There can be multiple sink and source tasks. And since we need to update the values of DQ, LQ, and EQ every time a task is scheduled, determining the priorities of tasks can be a time-consuming process. Moreover, both source and sink tasks are used in the formulation, but the source tasks should be placed at the $0$ level, and all of them can be considered as sink nodes. Moreover, the exponential increase in the runtime of the method compared to Internet Computing Optimal (ICO), \cite{Cordasco:2007} and Area Oriented (AO), \cite{Cordasco:2010} shows that it is better not to use PB when there is a possibility that a complete bipartite subgraph of a DAG exists (which is the case when the number of nodes of a DAG is very high). Moreover, the proposed method has an advantage over other methods only for DAGs with a very small number of nodes. For a large number of nodes, it is better to use other methods because of the runtime of the proposed method. Moreover, comparing the average makespan reduction without describing the makespan of the method does not tell us anything about the goodness of the method. The method can only improve the makespan because the resulting makespan can be very high compared to the optimal task assignment. So the average makespan reduction will be very high, but at the same time the generated makespan can still be very large.

\cite{Ours:2020} explored algorithms assignment to the robot, the fog, and the cloud that simultaneously minimize the maximum memory required by the robot and the total time required to execute all algorithms. In the proposed method, the set of all tasks is decomposed into the set of all algorithms and translated into a DAG. The algebra of memory and time is defined based on the precedence order of algorithms to describe the average execution time and memory consumption of the robot. Then, the branch-and-bound algorithm is applied to find the optimal allocation. 


To apply the method, complete information about the algorithms is required, such as the size of the processing, input, and output memory and the space complexity of the algorithms, as well as the average execution time of the algorithms on each processing unit and the average communication time between the processing units. Moreover, the method cannot be applied to a system with multiple robots.

\cite{Minjia:2021} proposed a method to find the optimal task allocation in the cloud that minimizes the completion time with load balancing of processors. The authors, recursively categorize the tasks and then use the clustering scheduling method that minimizes the worse scheduling length. Then recursively return to the original DAG by assigning each node to an appropriate processor to maximize the gain. For clustering, the critical path is used, and the nodes in the critical path are added to clusters considering the scheduling length. 


The optimal performance of their approach with minimal scheduling time is when the original DAG can be split into multiple disjoint classes (only once) of DAGs with at most a single edge connecting each class to another. And there is no argument that categorizing tasks reduces scheduling time compared to scheduling the entire DAG. Moreover, one of the main goals of categorizing tasks into different classes is to minimize scheduling time. However, in scenarios where all tasks are executed in parallel or serially, the scheduling time is higher than the usual scheduling methods because more time is required for clustering and categorizing tasks. In addition, the method should be tested and compared with at least one other scheduling method.

\cite{Pu:2021} proposed a method to find the optimal task allocation that minimizes the task completion time considering the price of using the cloud infrastructure. In the proposed method, tasks are distributed among multiple resources considering the budget constraint. To minimize the completion time, some of the tasks are duplicated to resources that are in the $\Idle$ state in the same time window so that the completion time of their immediate successors is reduced. 


Duplication of tasks is done regardless of how much duplication of a task can improve performance. Also, the budget must be large enough to apply the algorithm. Moreover, in \cite{Pu:2021}, the sub-budget has a different scale than the budget. To obtain the total weights, the terms with different scales must be added, which is not allowed.

\cite{Shafiq:2021} studied minimizing the makespan and maximizing the resource utilization in the cloud, where tasks are assigned to virtual machines based on their arrival time, and the loads on virtual machines are migrated to another available virtual machine when the deadlines of tasks are violated. 


If the deadline of all tasks is very long, all tasks can be assigned to a single virtual machine. The proposed method provides optimal performance for load balancing when the length of tasks is similar and the deadlines are short. The method performs poorly compared to greedy algorithms when the number of tasks is large.

\cite{Bharti:2022} explores offloading computational tasks to the cloud or to the fog of a collaborative robot architecture when performing machine learning tasks given the constraints of robots. The study investigates offloading tasks considering reliability, security, and privacy with a method called CoRoL, which uses split learning that enables offloading without revealing the data and reduces the amount of offloaded data by splitting it. Tasks are considered machine learning tasks (ML), and when a task is assigned to a node in the architecture, a portion of the dataset is moved to that node to be trained with the learning model and expected accuracy. If a node achieves the expected accuracy after training, the task is assigned to that node. The problem is then translated into an MILP, and depending on CPU load, energy consumption, and available memory, the node to which a task should be assigned is selected. 


The tasks are considered as ML tasks only, but in case of some other tasks that should be offloaded to other nodes to minimize energy consumption, the splitting method does not work. Another issue is that the method does not recognize when a task should be better offloaded to other nodes.

All contributions that solve the problem using a combinatorial approach are summarized in Tables \ref{tabc1} and \ref{tabc2}. Combinatorial methods are very efficient for solving certain types of problems, but often have scalability issues when applied to larger, more complex systems. One possible solution is to combine combinatorial techniques with metaheuristic approaches such as genetic algorithms or simulated annealing, which can help to explore larger solution spaces more efficiently.

\section{Reinforcement Learning}
In this section we discuss contributions that solve the problem using a reinforcement learning approach. The works are separately analyzed based on whether the cloud infrastructure is considered or not.
\subsection{Without Cloud}
\cite{Bian:2019} addressed the problem of unknown resource demand of a task before the arrival of the task and the fact that the resource demand may not be compatible with the scheduled resource, which is called multi-resource fairness. The authors proposed an online task scheduling tool, called FairTS, that learns to shorten the average task slowdown while maintaining multi-resource fairness. The average task slowdown is the average of the time differences between the arrival of the task and the completion of the task proportional to its length (the length of a task is the execution time of the task when it is assigned to its requested resource). The average slowdown of tasks allows us to handle tasks with high computational demands. At each time step, new tasks arrive to be executed by the fog and resources are allocated to each task. In the fog, there are several types of resources with different capacities. For each task, the arrival time, the start time, the finish time, the bandwidth demand for uploading, and the computation demand with a certain number of resources required for each resource type are known. The bandwidth capacity limit between the end-users and the fog is also known. Dwell time is defined as the time between the arrival of a task and its completion, which is divided into different types of delays, waiting delay (the time between the arrival of a task and its start time), transmission delay (the time required to transmit the task to the fog system), and execution delay (the time required when the task is split into multiple subtasks and assigned to multiple resources). Assume that a fixed set of tasks arrives in the system. The first objective is to minimize the task slowdown, and the second objective is to minimize the variance over the dominant shares of the tasks for the set of all maximum shares of each resource in all tasks. The problem is converted into an optimization problem, and then Deep Reinforcement Learning is used to solve the problem. The policy gradient method (PG) is used 
and the current state is given as input and the action probability vector as output. At each time point, the agent observes the final state of the environment and applies an action to it. This changes the state of the environment to the next state. This action generates some rewards which are collected by the agent. Based on the collected rewards, the agent then uses a policy to decide how to maximize the expected cumulative rewards in the long run. The fog is the environment, the availability of resources and the resource demand of tasks at a given time are states, and the resource allocation is an action. 


However, the simultaneous minimization of the two objectives is the goal of all load balancing methods, but no corresponding load balancing studies have been conducted. It has been shown that the performance of the proposed method is within the performances of random and greedy (the method SET \cite{Havill:2003}, which is a method that assigns tasks to resources with the shortest execution time). The reward formula ignores the arrival time and the finish time of a task. However, the finish time can only be determined after resources have been allocated to the task. The task arrival time should be taken into account because the tasks that arrive earlier should be executed earlier. In the experimental simulation, it is assumed that $\beta=0$ (a constant parameter that weights the importance of shortening the task slowdown and resource fairness), which contradicts the initial assumption that $\beta > 0$. Considering the Poisson distribution with the parameter equal to $0.8$, this means that the probability of getting at least $6$ tasks is about $0.0002$, and with $5$ CPUs on the fog, we are more likely to have a smaller number of tasks than the number of CPUs. This avoids the more general problem of assigning more than two tasks to the same CPU.

\cite{Cui:2023} developed a task scheduling method for edge-enabled collaborative robotic networks to maximize the number of completed tasks. They proposed a decentralized multi-agent method using deep reinforcement learning with a partially observable Markov decision policy to determine the assignment of users to groups of collaborative robots and edge devices considering the deadline for the tasks, and then assign the communication and computational resources of the edge devices to the robots to offload the computational tasks. In this model, the edge device is used for scheduling tasks and forwarding them to the robots. In the proposed method, the reward function for each agent has been defined based on the global performance of the overall system. They consider the interdependence between the assigned tasks and assign them to the same group of collaborative robots.

If some of the tasks in the stream are interdependent (e.g. if tasks are to be completed at the same time), then all these tasks should be assigned to the same group of robots, but since they are parallel, they may be distributed between two groups of robots. In the worst case, this also means that all tasks should be assigned to a single group of robots. Furthermore, the communication delay is not fully considered as the communication between the users and the edge devices as well as between the edge devices and the robots is not included in the formulation. Also, the method requires additional optimization for the optimal distribution of tasks among the edge devices to avoid additional communication between the edge devices. They tested the performance of the proposed method using simulations.

\subsection{With Cloud}
\cite{Wang:2019} proposed a method for adaptive resource allocation. It collects knowledge from the environment, incorporates adaptive policies for dealing with environmental changes, and makes a series of decisions. The method aims to minimize service time by considering routing and computation delays as service time and maintaining a balance in terms of computational power and resources. The authors consider the graph of the architecture. When a request is made for an application, it is transmitted to the node that hosts the application. The service time is measured as the sum of the routing delay of the request and the data processing delay. Processing delay and routing delay are formulated. The variance of network load and variance of computing load are minimized to balance the loads. The main objective is to minimize the average of the sum of processing delay and routing delay from one node to another for all node pairs. Then, the collected information is fed into a Deep-Q network, which is somewhat similar to Bayesian statistics (finding the best prior that maximizes the total reward) that adjusts the model after each new observation. At each time step, the agent observes the environment (state). Based on a certain policy, it performs an action and receives the reward for that action. The action moves the state to the next state and so on. At each time step, the policy is updated based on the reward, and the goal is to maximize the total reward. This is similar to Bayesian statistics: observing the new data at each time step is the state, the action is the description of the distribution model, the policy is the prior, then the prior is replaced by the posterior, which can be viewed as updating the policy. The rewards can be considered as the negative mean square errors of the data from the model. For the total rewards, the weighted cumulation of the rewards is considered, the action-value function is defined as the expectation of the cumulative reward when an action is performed in a state according to the policy, where the policy is a translation of the outcome of states by performing actions into a probability distribution, actions are updated by the action-value function at a certain learning rate. 


It is assumed that only a single node hosts an application, so in case of malfunction of this node, the whole system is not able to execute the applications hosted by this node. Moreover, in the formulation of routing delay, a fixed value for routing capacity is considered and all paths between two nodes are considered. By finding and using the shortest paths, the routing delay is minimized and the routing capacity does not imply the available capacity for data transmission between nodes. 

\cite{mostafavi:2020} proposed a scheduling method to minimize the makespan and response time and increase resource efficiency. The scheduling method is based on reinforcement learning. At each time step, the size of the occupied buffer and the total length of virtual machine tasks are considered independent to use Bayes' theorem, and the $Q$-values are estimated. 


The $Q$-value function described is not mathematically well-defined, the sum of rewards must converge in the long run, otherwise the $Q$-value function cannot be a function. Moreover, it leads to a reduction of the state-space dimension, which is just a rescaling of the components. If the number of iterations is smaller than the smallest buffer capacity, the $Q$-value can be poorly estimated.

\cite{Ding:2020} has proposed a method to find an optimal solution for distributing tasks to servers that minimizes the cost to the user. In the proposed method, the task allocation problem is translated into a reinforcement learning problem where the reward function is the negative average user cost. To reduce the solution space, the tasks are divided into a single server and coalitions. Each task of a single server type is assigned to a single server, while the tasks of coalition types are proportionally distributed among the servers in the coalition to which they are assigned. 


The solution space is infinite, even with the coalitional reinforcement learning method. Moreover, the main algorithm presented does not check the constraints of tasks and available resources. Some tasks can be executed by some servers, but not all, which introduces additional difficulties in task classification. Also, the experimental results are compared with a result that is not focused on minimizing the user cost.

\cite{Liu:2023} proposed algorithms based on deep $Q$-networks and dueling deep $Q$-networks for scheduling decentralized cloud robotic systems. Their goal is to find an optimized task scheduling that considers the parameters of lifetime, specifications, quality level, reliability, and performance of robots before task assignment and the parameters of total price and total completion time of robots after task assignment. In this way, they maximize the overall quality of the service and minimize the overall performance of the service. They compared their model with random assignment, task assignment to the robots with the earliest start time, and task assignment using round robin scheduling, where resources are evenly distributed among jobs, see \cite{Moseley:2022}.

They tested their proposed model with simulated data with tasks from uniform distributions with the same number of tasks and robots. However, the statistical significance of the proposed method was not investigated as there are several more realistic scenarios, e.g., systems where most tasks are long and short tasks are rare, or for the scenario where most tasks are short and long tasks are rare. Furthermore, the index used for task assignment is a linear regression, where the mean of the values of all components is used to find the solution, removing the existing correlations between components, e.g. the relationship between larger computational resources and faster computation. This means that by better identifying the parameter space, we can better extract the component values and thus perform better scheduling. In addition, the ability of robots to perform tasks is also important, as not all robots are able to perform all tasks.

\cite{Chen:2023} investigated task scheduling in cloud manufacturing, i.e., when a user sends a request to the cloud, manufacturing services are provided in the cloud based on the user's needs. This requires the optimization of resource usage and load balancing. They proposed a deep reinforcement learning approach for task scheduling that takes into account the dynamics of the environment, such as when service providers are offline or factory resources are down, as well as the interdependence between scheduled tasks and the trade-off between resource utilization and load balancing. They have introduced a strategy called Maximum Posteriori, which considers the scheduling solution with the task assignment with the highest probability in the posterior resulting from the Markov decision process. The main goal is to minimize the total time and cost of resource usage and balance the loads. They have conducted experiments with simulations and collected data from the real world.

The given policy can solve the scheduling problem, but if the posterior distribution is randomly shifted to the left or right, the solution will not be optimized. Moreover, balancing the loads is equivalent to minimizing the makespan (minimizing the total time consumed), so the two objectives are completely interdependent on each other.

\cite{Li:2023} investigated task scheduling for time-critical tasks in robotic network cloud systems considering energy consumption and total completion time. They proposed a Deep Reinforcement Learning approach with Proximal Policy Optimization that considers traffic volume and adapts to environmental changes to minimize completion time with task prioritization, minimize energy consumption based on distance, and maximize resource utilization. To minimize energy consumption, they focused primarily on data transmission via a selective communication path between devices. This path involves communication between devices with the highest bandwidth and the least interference during data transmission.

Changes in the environment can lead to a loss of communication or a greater delay in two consecutive time frames, as the devices are in motion and this can increase the distances to each other or to the edge devices. This means that the optimal communication path for data transmission is selected in one time frame, while the communication path may be completely different in the next time frame. This can lead to additional delays if large amounts of data need to be transferred. The method was tested in practice with the help of simulations and experiments.

All contributions that solve the problem using a reinforcement learning approach are summarized in Table \ref{tabr1}. Reinforcement learning methods are becoming increasingly popular due to their ability to adapt to changes in real time. However, as they rely on large data sets, their applicability may be limited in environments with a lack of sufficient data. One possible solution is the integration of transfer learning or model-based reinforcement learning, which can reduce data requirements by using pre-trained models or simulating environments.

\section{Alternative approaches}
In this section we discuss contributions that solve the problem using any other approaches such as Bayes theory, automata theory, geometrical approach, and developing a new architecture. The works are separately analyzed based on whether the cloud infrastructure is considered or not.
\subsection{Without Cloud}
\cite{Schillinger:2018} proposed a centralized dynamic task allocation of a robotic network based on modeling the problem as a non-deterministic finite automaton, where the task is decomposed into several subtasks and assigned to different robots by minimizing the cost depending on the energy level of the robots. 


There is no comparison with the state-of-the-art methods. The proposed method does not consider the communication time between devices and does not take advantage of the cloud infrastructure to further reduce the cost.

\cite{Gombolay:2018} investigates task assignment with the goal of minimizing the makespan and/or energy consumption by considering latency and spatial proximity to tasks. The authors propose a method for assigning tasks to a collaborative human-robot system where all temporal and spatial proximity constraints are satisfied and the makespan is minimized. First, the algorithm obtains the information about the agents and their capabilities, the set of all subtasks and their precedence order, and at least one of the parameters cutoff and timeout (the cutoff is a given upper bound on the makespan and the timeout is the duration of searching for a solution). Then, the algorithm (similar to the branch-and-bound algorithm) searches for the first task assignment method where the makespan is less than the cutoff value or the runtime is greater than the timeout value. 


The algorithm recursively performs agent assignment and task sequencer, where the former determines which agent performs which task, and the latter provides a task assignment method for a collaborative human-robot system that satisfies all temporal and spatial proximity constraints, but without minimizing the makespan. However, the completion of the algorithm does not guarantee the optimality of the obtained solution and depends on the initial cutoff and timeout values. Moreover, there is no specific method for selecting the initial cutoff and timeout values.

\cite{kim:2019} proposed a method to minimize energy consumption by combining multiple allocation methods: Task classification, processor allocation, queue ordering, task migration, DVFS, and task stealing. The optimal allocation with minimum energy consumption is achieved by classifying tasks as short and long with the average execution time of all tasks by allocating short tasks to slow processors and long tasks to fast processors. In each stream, the shortest task is executed first. It then balances the load by moving the longest remaining task to a fast processor, and reducing the frequency if the task can complete before its deadline. It also allows task stealing. 


To balance the method of assigning tasks to processors, the information about the arrival of tasks must be known in advance. The order of tasks is not considered and tasks can be executed independently. Moreover, load balancing and task stealing contradict each other. And task stealing may result in dropping a task that could be completed before its deadline. Their method does not work when the time windows of tasks are connected.

\cite{stylus:2020} investigates the minimization of path planning costs in a robotic network. The authors assume that the workspace is divided into several disjoint regions of interest. Then, the dynamics of each robot is translated into a weighted transition system and then extended to all robots as a component-wise direct product, where the path that each robot can take is an infinite path of states with transition relations. Finite paths can be extracted from the infinite path, since the sequence of the infinite path (infinite words) has finitely many states and there should be an idempotent subword that can be obtained by the canonical form of the infinite word, \cite{Simon:1975, Pach:2018}, so the optimal path planning can be extracted. It is a fast method to find the path planning of robots when the number of robots is very large. 

\begin{figure}[tbp]\centering
\centering
\includegraphics[width=0.2\linewidth]{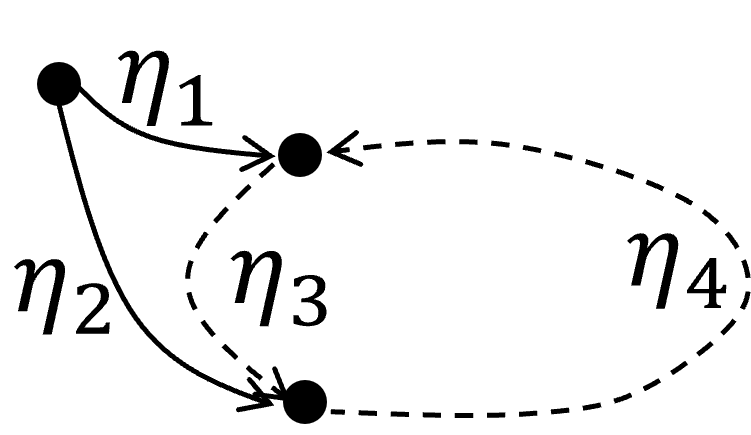}\hspace{10pt}
\includegraphics[width=0.2\linewidth]{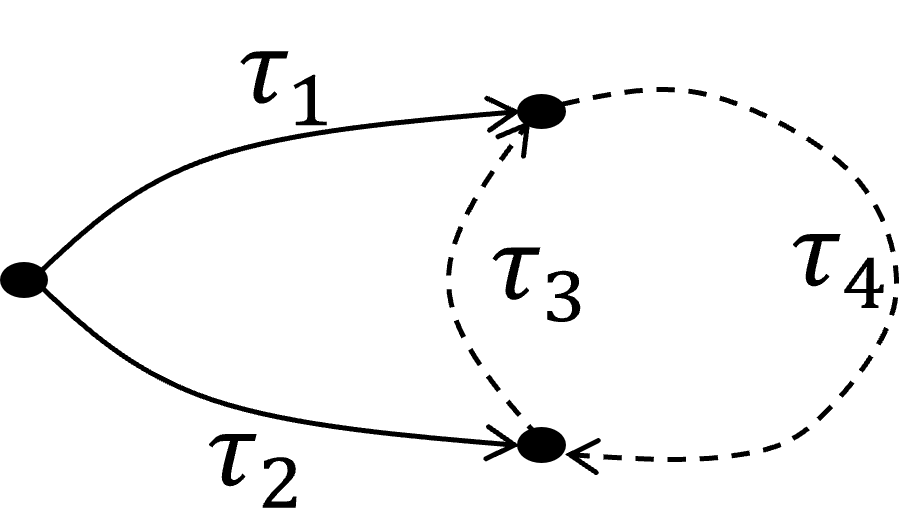}
\caption{Two different prefix with different costs (left). Two different prefix with the same costs (right). $\eta_i$ and $\tau_j$ are pseudowords (infinite words) corresponding to paths.}\label{figu6}
\end{figure}

In some scenarios, it is difficult to find the canonical form of the infinite word. Moreover, to generate a canonical form for an infinite word, random samples of states are used, and depending on the sample, there may be different prefixes with the same cost, see Figures \ref{figu6}, and the method does not describe how to choose a suitable prefix when the costs are the same.

\cite{Pupa:2021a} proposed a method for scheduling tasks that minimizes the nominal execution time, time span, and waiting time, and maximizes the job quality metric while taking into account task priority constraints. Task scheduling (dynamic scheduling) is performed by real-time monitoring of human activities. The proposed method consists of two layers: In the first layer, optimal task schedules are generated for both the human and the robot: Tasks are scheduled based on minimizing the three parameters (nominal execution time, time span and waiting time) and maximizing the job quality metric considering the priority constraints for the tasks. In the second shift, based on the first shift schedule and human monitoring of the actual task execution time, rescheduling is performed to further minimize the time span and waiting time. The first layer is called the task assignment layer, and the second layer is called the dynamic scheduling layer. The human monitoring block is used to track the actual execution time of the assigned tasks for the human. This information is used to adjust scheduling according to human behavior for optimal scheduling. The communication interface block is used to identify the cases where the robot is unable to perform a task and a human decides to perform the task instead of the robot. It is assumed that all tasks should be performed by the collaboration of a single human and a single robot. The goal is to perform multiple jobs, and each job consists of multiple tasks. To assign each task to either the human or the robot, a job quality metric is first estimated based on the last observation of the human operator's state and then updated by monitoring the human operator. Then, the tasks with the most recent metric are sent to the dynamic planner to schedule the next task of the robot and the human operator. The human and the robot then communicate with the planner to find out whether the tasks need to be rescheduled or not. 


The evaluation of the metric is unnecessary as it is updated immediately after the evaluation. Moreover, monitoring by the human operator is only checked when a new task arrives in the system. However, monitoring should also be done when tasks are rescheduled, otherwise, the metric remains the same and tasks are assigned to the same agent with the same metric. Time pressure, physical risk, and job autonomy are also important factors in job quality. The last two are conceptual terms that should be converted into a numerical metric. The proposed formulas for job quality are defined as the weighted sum of the cost and average cost to the human agent over the duration of performing a task. Now, if we assume that task reallocation is beneficial if it reduces human labor time, then it is most beneficial if all tasks can be done entirely by robots. But at the same time, it means the unemployment of the human employee. Moreover, the salary also depends on the working time, so the working time must be increased. All these mean that the cost of the human agent must be a $U$-shaped function, which is not described in this paper. Moreover, the dynamic planner is completely determined by the human agent. It is assumed that the human agent can correctly select the tasks to be done, which is not always the case. And the monitoring function is used to test whether the human or the robot can complete all the tasks assigned to them. This result is then communicated to transfer some of the tasks from the human to the robot and vice versa. Since the levels and schedule define two orders of tasks and both orders of tasks should be maintained, moving tasks to higher levels cannot be easily considered.

\cite{Sa:2021} investigated the distribution of tasks for human-robot collaboration and describes a method for generating a value representing a human who should or should not trust a robot to perform a task. In the proposed model, task requirements and agent capabilities are translated into values between zero and one. The trust function is defined as a success probability of completing the task and a belief probability, which plays the role of updating the trust. The trust becomes a sigmoidal function, and the belief is considered as a uniform distribution. 


The natural trust has no proper and simple formulation and needs to be discretized to find its value. Moreover, the probability of success is not a probability function, but a fitted sigmoidal function with two additional parameters, where two parameters are positive. Also, the number of parameters in the proposed method is large, and the penalization of parameters shows that other state-of-the-art methods have better performance because they have fewer parameters.

\cite{Ours:2022ral} is an extension of the work \cite{Pupa:2021a} by proposing a method that overcomes some of the limitations of \cite{Pupa:2021a}. The authors investigated task allocation when a single human and a single robot collaborate to perform a set of jobs. A method was developed to minimize the $\idle$ time of each agent and the completion time of the jobs, taking into account the task execution time of the currently executing job, the precedence order of the tasks, and the compatibility of the tasks performed by the agents. When a job arrives in the system, it is considered as a directed acyclic graph whose nodes represent the set of tasks required to complete the job and whose edges represent the precedence order of tasks. At the same time, for each task, the weights of the quality metric are determined as a metric to describe the agent that is better suited to perform each task. This results in the priority of the tasks that should be assigned to the agents. Then, the agents are monitored in real time to detect performance delays and possible rescheduling of some tasks while the agents are executing the tasks. To reduce the completion time of a job, the total $\idle$ times of both agents are reduced by prioritizing the tasks that release more tasks upon completion. The method performs all necessary steps and evaluations automatically, without the need for a human agent to help with scheduling or rescheduling. 


It does not consider tasks with time-window constraints, which is an important factor when prioritizing tasks. In addition, the method cannot be applied to cases with multiple humans and robots.

\cite{Wu:2023} investigated secure task allocation for intelligent transportation systems. Users and machines send their requests without revealing their sensitive information, and task allocation is then performed using an encrypted matchmaking strategy. A user sends the linear integer secrets instead of requesting a specific task. The matching strategy is then determined using $\mathrm{AND/OR}$ operators. They translate the matching strategy into a bilinear homomorphic image of their direct product of $p$-groups, where $p$ is a large prime number. Then they compare whether the generated elements in the image have the same order as the elements of the pre-image to determine whether the task and the users match.

Since the pre-image groups are cyclic groups with the order of the prime number $p$, all non-trivial elements have the order $p$. Decoding the information can therefore be done simply by selecting two dissimilar elements, at least one of which is non-trivial, so that the entire $p$-group can be generated by replacing the feeds. They tested the performance of their method using simulations.

\subsection{With Cloud}
\cite{Zhang:2018} investigated the optimization of task scheduling performance in clouds and used two stages for task assignment. First, tasks are classified using the Bayes classifier, which classifies tasks based on historical scheduling data. Then, a match is found between the tasks and the most appropriate virtual machine, and the tasks are scheduled into the $\Idle$ time slot of the selected virtual machine. 


The proposed method is inherently greedy and does not consider switching tasks between virtual machines, which affects the performance. Experimental results are obtained by comparing the method with min-min and max-min methods. It is easy to see that the comparisons are redundant, as these two methods are not optimal for a large number of tasks. This means that all results involving some kind of optimization have better performance than min-min and max-min.

\cite{malik:2019} proposed a method for minimizing energy consumption for data transmission in a cloud center, where the data transmission is done through multiple volunteer devices instead of the Internet. 


The volunteer devices have limited amount of energy. The overall energy consumption for data transmission is minimized, but the cost of energy consumption is distributed among the volunteer devices. Moreover, load balancing is required to avoid using only some of the volunteer devices for data transmission.

\cite{botta:2019} proposed an architecture consisting of robots, fog, cloud, and dew, where dew nodes are intermediate nodes between fog and robots, based on the concept of microservices provided by end-user devices. In this, local processing devices near the robot are used to store data or for computation to minimize the computation time and storage space of robots by adding a new layer (dew) to the architecture, where dew nodes are smart things such as mobile devices and laptops near the robot with different capabilities. 


The authors did not describe the constraints and criteria for assigning a task to each node of any layer as the number of layers increases. The complexity of the model increases as minimizing the energy consumption limited to the end-user devices becomes more complex.

\cite{Ours:2020h} studied task dependencies, average execution time, communication time, communication instability, compatibility, and robot, fog, and cloud capability, where each parameter is measured for all processing units and translated into a hypervolume to be maximized. The set of all tasks is decomposed into the set of all algorithms and translated into a DAG. Compatibility, communication, and capabilities of the robots, the fog, and the cloud are considered as parameters and the tasks are translated into numerical values with respect to each robot and based on a parameter. Thus, a subspace of a hyperspace is created for the parameter. Then, the robots are translated into a subspace of a hyperspace with respect to each task based on the parameter. The intersection of the constructed hyperspaces for each robot and task creates a hypervolume to be maximized. The hypervolume with the maximum size results in the optimal assignment of tasks to robots, optimizing the performance with respect to the parameter. 


Defining the numerical values that relate robots and tasks for a given parameter is not a simple task. Moreover, some tasks, such as the energy consumption of robots to perform tasks, are continuous, which makes it difficult to find a suitable continuous function. Therefore, it needs to be discretized, which may reduce the accuracy.

\cite{Wang:2023} proposed an architecture for collaborative cloud-edge systems (cyber-physical machine tools, CPMT) with task offloading process to balance the loads, minimize the delay and maximize the throughput. The proposed architecture consists of three layers: the physical layer, the cyber layer and the application layer. The physical layer contains all physical components. The cyber layer is a virtualization of the physical layer that helps to predict, evaluate and optimize the performance of the system based on data operations. The application layer includes all services provided to users. During system deployment operations, data is collected, such as data transmission, individual device performance, completion time and so on, for further analysis. This information is used in the cyber layer, which uses it to determine the optimal architecture for task completion through a combination of a deep neural network, expert experience and domain knowledge. In the search for the optimal architecture, the strategies of fastest response and load balancing were considered separately.

Since the communication delay is random, device failures may occur, and this information cannot be collected and used in the model during the deployment phase, so the model may fail in the long run. Therefore, the model must continuously collect information and update its status in time and when performing tasks to better handle unexpected scenarios that may occur in the long run. In addition, load balancing and delay minimization are considered separately. To test their method, they conducted real-world experiments.

\cite{Hao:2023} considers a robotic network cloud system in which edge devices utilize green energy, and their goal is to find an optimal task scheduling that mainly utilizes the supplied green energy while maintaining the operating state of the device without depleting the battery that stores the generated green energy, taking into account the prediction of green energy. In this way, the required energy can be maintained over the next few days. The model collects information such as the battery volume, the daily energy required to keep the device running, the prediction of the energy that will be collected in the future, the energy consumption when sending and receiving data, the computing frequency of the device, the number of tasks, the number of instructions for the tasks, their deadline, their execution time and so on. Then they used a weather forecast model to predict the battery's energy level for the next day. Then they transformed the problem into an optimization problem and proposed a genetic algorithm to find an optimized task scheduling method that minimizes the total time and consumption of non-green energy. 

The proposed method assumes that the number of incoming tasks remains constant, and no inaccuracies such as predictions, equipment failures, and communication delays were considered. The performance of the proposed method has been tested using simulations.

All contributions that solve the problem using approaches other than optimization, combinatorial, and reinforcement learning are summarized in Tables \ref{tabot1} and \ref{tabot2}. Problem-specific approaches are very efficient within their defined framework, but their rigidity limits their adaptability to dynamic or unforeseen conditions. A promising direction for future research is the development of hybrid models that combine the precision of problem-specific methods with the flexibility of generalizable approaches such as machine learning or adaptive algorithms to improve their robustness in changing environments.

\section{Conclusion}
In this paper, we have reviewed the task allocation and scheduling strategies and associated metrics suitable for robotic network systems, including cloud, fog, and edge computing environments. We have created seven summary tables (in the supplementary material) that summarize the main points of each approach. The different approaches have been developed with different optimization goals, such as minimizing completion time, energy consumption, communication overhead and computation time, or maximizing resource efficiency and the number of completed tasks.

We have found that many studies focus on task scheduling without considering certain constraints such as deadlines or task execution dependencies, while other studies prioritize certain goals, which often leads to trade-offs. For example, increasing the number of completed tasks typically increases energy consumption and computational costs. We have also reviewed new architectures that aim to offload computations to nearby processing units, such as fog or cloud, which can lead to additional communication delays.

The current trend suggests that combinatorial optimization is one of the most effective approaches, especially as faster processing units reduce the need for cloud infrastructures, allowing more tasks to be processed locally on the edge. In addition, we observed the increasing use of reinforcement learning approaches to solve task allocation and scheduling problems. Nevertheless, there is still a need for a unified theory that integrates different structural properties of tasks and architectures while adapting to specific problems in robotic networks, especially those operating in noisy or unpredictable environments.

Some important open problems remain:
\begin{itemize}
\item The development of unified frameworks that address the trade-offs between different optimization goals (e.g., completion time vs. energy consumption).
\item Investigate whether a global optimizer can be developed that can handle a wide range of task allocation and scheduling challenges.
\item Develop adaptive algorithms that can switch between optimization and reinforcement learning methods depending on the problem context to take advantage of both methods.
\item Develop more robust models that work effectively in less controlled, real-world scenarios with dynamic environments and hardware constraints.
\end{itemize}
Practical challenges such as hardware limitations, dynamic and unpredictable environments, and real-time processing requirements are essential to consider when moving from theory to real-world deployment. Although this paper focuses primarily on theoretical methods, it is important to understand how these methods can be applied to real-world robotic systems. Developing scalable solutions that can operate under real-time conditions and hardware constraints remains a key challenge for future work. Further research is needed to close the gap between theoretical approaches and their practical implementation in large-scale, real-world robotic networks. Future research should also focus on developing hybrid models that combine the strengths of cloud, fog and edge computing while ensuring adaptability to real-time conditions and unreliable environments. In particular, methods that can seamlessly switch between centralized and decentralized processing depending on the needs of the system will be crucial for the next generation of robotic networks.

\section*{Acknowledgments}
This work was partially supported by operation Centro-01-0145-FEDER-000019 - C4 - Centro de Compet\^{e}ncias em Cloud Computing, co-financed by the European Regional Development Fund (ERDF) through the Programa Operacional Regional do Centro (Centro 2020), in the scope of the Sistema de Apoio \`{a} Investiga\c{c}\~{a}o Cientif\'{i}ca e Tecnol\'{o}gica - Programas Integrados de IC\&DT. This work is supported by NOVA LINCS ref. UIDB/04516/2020 (\url{https://doi.org/10.54499/UIDB/04516/2020}) and ref. UIDP/04516/2020 (\url{https://doi.org/10.54499/UIDP/04516/2020}) with the financial support of FCT.IP. It was also partially supported by the Computer Science and Communication Research Centre (CIIC), Escola Superior de Tecnologia e Gest\~{a}o (UIDB/04524/2020), with financial support from FCT-Funda\c{c}\~{a}o para a Ci\^{e}ncia e a Tecnologia under reference CEECINST/00060/2021/CP2902/CT0009 (\url{https://doi.org/10.54499/CEECINST/00060/2021/CP2902/CT0009}). Part of this work was conducted while Saeid Alirezazadeh was affiliated with C4-Cloud Computing Competence Center, Universidade da Beira Interior, Covilh\~{a}, Portugal. He is currently with the Computer Science and Communication Research Centre (CIIC), Escola Superior de Tecnologia e Gest\~{a}o of Instituto Polit\'{e}cnico de Leiria, Portugal.

\bibliographystyle{unsrt}
\bibliography{sample}

\newpage
\setcounter{page}{1}
\section*{Summary Tables}\label{summary}

\subsection*{Optimization}

All contributions that formulate the problem as an optimization problem are summarized in Tables \ref{tabo1} to \ref{tabo4}.

\begin{table*}[b!]
\tiny
\caption{Papers on task allocation and scheduling in robotic network systems using an optimization approach for solution.}\label{tabo1}
\begin{center}
\resizebox{\textwidth}{!}{\begin{tabular}{p{0.01\textwidth}p{0.01\textwidth}p{0.03\textwidth}p{0.02\textwidth}p{0.02\textwidth}p{0.03\textwidth}p{0.1\textwidth}p{0.1\textwidth}p{0.12\textwidth}p{0.12\textwidth}p{0.12\textwidth}p{0.05\textwidth}}
Paper&Year&Static/ Dynamic&Load balancing&Cloud Infrastructure&Number of robots&Parameters&Main objectives&Approach Used&Restrictions&Problems&Type of Experiments\\
\hline
\cite{Gini:2017}&2017&Dynamic&No&No&Several&Temporal constraints&Minimize a cost function or maximize a reward function for all robots for completing their tasks&Branch and Bounds or metaheuristics for the centralized model and Distributed Constraint (DCOP)-Based Methods or Market Based Methods for the decentralized model&Stable communication between all robots in both models&Requires rescheduling and new solutions upon arrival of new tasks and failure of communication. Requires an approximation method to reduce computation time. Multi-task robots and multi-robot tasks are not included and disjoint temporal models are not considered&-\\
\hline
\cite{NUNES:2017}&2017&Dynamic&No&No&Several&Time windows, and tasks' order&Developing a generalized optimization model for task allocation problem that all the existing methods can be extracted from it&A generic function optimization problem&Time constraints are simple, time-critical tasks cannot be correctly assigned to robots&Solution method varies from one objective to another and solution of multiple objective problems is not easy to obtain&-\\
\hline
\cite{wang:2017}&2017&Dynamic&No&Yes&Several&Communication time&Minimizing communication and computation costs&Hierarchical auction-based &Latency and memory usage by robots are not considered. No fair comparison with state-of-the-art methods. Robots require to communicate with the cloud to perform all their assigned tasks&Dynamic change of architecture topology and including the case where a robot do not require communication with the cloud are not considered.&Simulation and real-world\\
\hline
\cite{LEE:2018}&2018&Dynamic&No&No&Several&Task completion time, resource consumption, communication time&Minimize task completion time, resource consumption, and communication time&Market-based approach&Recharge time and transfer time to charging stations may change the solution&Compared with the cases without task rescheduling and without considering resources, without testing and by design the improvement of the performance can be observed&Simulation\\
\hline
\cite{li:2018}&2018&Static&No&Yes&Several&Execution time, transmission time, dependencies of algorithms&Minimize overall time for performing tasks by the robots&Mixed-integer nonlinear programming&Does not fully account the communication time&Memory requirement by a robots to perform algorithms is not included. Time initiation should be when a robot send a request.&Simulations\\
\hline
\cite{Zhou:2018}&2018&Dynamic&No&No&Several&Number of robots, time limit&Find the smallest number of robots in a given time limit and assign tasks to them so that the robots can complete their tasks within the time-limit&A multi-objective optimization problem and a genetic algorithm is used to find a solution&weights of are non-negative integers. Partitions should not exceed the time limit and number of parts equal to the number of robots. The metric for defing bands is not defined.&The partitioning of vertices into a fixed number of parts is not unique. When solving the optimization problem, it should include a test for possible cycles after adding each random edge.&Simulation and real-world\\
\hline
\cite{Chenc:2018}&2018&Dynamic&Partially&Yes&-&Sudden demands require urgent responses&Minimizing number of virtual machines and physical machines with minimum distance between virtual machines and physical machines to ensure resource allocation optimization and timeliness&A multi-objective optimization problem and a genetic algorithm is used to find a solution&The multi-objective model is an NP-complete problem. Load unbalance degrees are compare after allocating virtual machines on physical machines. It is assumed that the virtual machines are running under full load&Some terms are not defined. The second objective is a constant minimization.&Simulation\\
\hline
\cite{Notomista:2019}&2019&Static&No&No&Several&Capabilities and energy consumptions&Minimizing the cost function&Mixed Integer Quadratic Problem and updating the priorities of the tasks by time is used for solution&The cost function must be of class $C^1$ and the highest priority should be unique&proposed method works only for special cost functions and environmental changes is not considered&Real-world and simulation\\
\hline
\cite{Dantu:2019}&2019&Dynamic&No&No&Several&Communication instability&Maximize the sum of the agents' utilities, where the agent's utility is defined as a function of the tasks assigned to the agent in the completion order&A multi-objective optimization problem solved by an auction-based strategy&Tasks with relatively very long execution times with respect to the average communication time are not considered&Logarithmically transforming the main objective translate the problem to linear programming. The term bundle has the mathematical meaning and is not an appropriate term to be used in this context.&Simulation\\
\hline
\cite{singh:2019}&2019&Dynamic&Yes&Yes&-&Energy consupmtion, resource utilization, Security&Load balancing, minimizing the energy consumption, maximizing resource utilization, and maximize the security&A multi-objective optimization and Pareto front for solution&Different energy consumption for different tasks and data transfer cost are not considered&Number of requested tasks and their energy consumption are not considered&Simulation\\
\hline
\end{tabular}}
\end{center}
\end{table*}
\begin{table*}[b!]
\tiny
\caption{Papers on task allocation and scheduling in robotic network systems using an optimization approach for solution.}\label{tabo2}
\begin{center}
\resizebox{\textwidth}{!}{\begin{tabular}{p{0.01\textwidth}p{0.01\textwidth}p{0.03\textwidth}p{0.02\textwidth}p{0.02\textwidth}p{0.03\textwidth}p{0.1\textwidth}p{0.1\textwidth}p{0.12\textwidth}p{0.12\textwidth}p{0.12\textwidth}p{0.05\textwidth}}
Paper&Year&Static/ Dynamic&Load balancing&Cloud Infrastructure&Number of robots&Parameters&Main objectives&Approach Used&Restrictions&Problems&Type of Experiments\\
\hline
\cite{Liu:2019}&2019&Dynamic&No&Yes&Several&Delay for task offloading&Model the total delay for task offloading, construct an optimal task allocation minimizing the total delay&Optimization problem is described by a probabilistic model of the total delay and considering the data arrival and service as (super) martingales&No explanation for cases: input data must be collected from other nodes, a minimum data size for processing, and the optimal partitioning&The departure process should be min-plus. The defined threshold should be a global threshold. For a task, only one source node is considered. Propogation delay is assumed to be constant. RSUs are considered closer to edges than cloud.&Simulation\\
\hline
\cite{emam:2020}&2020&Dynamic&No&No&Several&Energy consumptions and capabilities&Minimizing the cost function in dynamic environment&Similar to \cite{Notomista:2019}, updates the states of the robots by time and updates priorities by environmental changes&Cost function is continuously differentiable. The highest priority task is unique&Environmental disturbance may reduce the cost and requires the frequency of the environmental change&Real-world\\
\hline
\cite{Geng:2020}&2020&Dynamic&Partially&Yes&-&time, cost, resource utilization&Load balancing and minimizing the resource utilization, time, and cost&A multi-objective optimization and use the hybrid angle strategy for solution, the objective of overloading made for load balancing&Combines ant colony, genetic algorithm, and local search to find an optimal solution which requires high computational time&Cannot be applied to a continuous and non-convex problem with a large population size&Simulation\\
\hline
\cite{Brown:2020}&2020&Dynamic&No&No&Several&task dependencies, task scheduling, route-planning, collision avoidance, makespan&Minimizing makespan and avoiding collisions&Multi-objective optimization problem with Traveling Salesman, tree search, and $A^*$ searches are used for solution.&The condition for moving an object that is the object and the robot have the same position is not sufficient. There should be an initial station.&To assure the conflict-free route for the second time, aonflict avoidance should be added. By construction, the runtime of ISPS must be higher than that of NBS all the time but, in the results, it is smaller than the runtime of NBS most of the time.&Simulation\\
\hline
\cite{Behrens:2020}&2020&Dynamic&No&No&2&Collision avoidance, scheduling, makespan, motion planning, average execution time&Minimizing makespan and avoiding collisions&A multi-objective optimization problem with backtracking search method to find solutions&Robots are identical. Region size and time intervals are not described.&In backtracking search the first upper bound is by a random selection on the solution space, and the steps to reduce upper bounds are not described. Different upper bounds yield different solutions. Robot dependency is not included. Possible collisions between components of a robot. Collision avoidance depends on voxelization sizes.&Simulation\\
\hline
\cite{Rahmanpour:2020}&2020&Dynamic&No&No&Several&Movement and communication&Motion planning and communication strategies minimizing energy consumption and collision avoidance.&Convex optimization problem&Dynamic obstacles. Requires tie-breaker for robots with identical distances to a spot. True optimal assignment of robots is not always by finding the shortest path.&Requires to solve the minimization problem of the total paths traveled by all robots to find the optimal assignment. Requires a constraint on the threshold radius around the robots.&Simulation\\
\hline
\cite{Dang:2021}&2021&Dynamic&Partially&Yes&Several&Response time, available resource, task order&Minimizing the service provisioning delay&An optimization problem solved by particle swarm optimization to find the suboptimal solution&All the fog nodes should be adjacent.&Task orders such as tasks to be completed at the same time or tasks to be executed at the same time are not considered. In the experiments the request rate is very small and the communication between the cloud and the fogs is very high, causing delays.&Simulation\\
\hline
\cite{Fu:2021}&2021&Static&No&No&Several&task dependencies, average execution time, communication time, energy consumption&Minimizing the total execution time and the total energy consumption of all edge devices&An optimization problem with a solution provides by optimal subtask scheduling&All edge and fog nodes generate a complete graph with the identical communication speed. The initial subtask of each task required to be executed on edge devices.&Provides a threshold instead of minimizing the energy consumption. Does not allow ofloading subtasks to neighbor edges. Data transmission speed and energy consumption are assumed to be linear.&Simulation\\
\hline
\end{tabular}}
\end{center}
\end{table*}
\begin{table*}[b!]
\tiny
\caption{Papers on task allocation and scheduling in robotic network systems using an optimization approach for solution.}\label{tabo3}
\begin{center}
\resizebox{\textwidth}{!}{\begin{tabular}{p{0.01\textwidth}p{0.01\textwidth}p{0.03\textwidth}p{0.02\textwidth}p{0.02\textwidth}p{0.03\textwidth}p{0.1\textwidth}p{0.1\textwidth}p{0.12\textwidth}p{0.12\textwidth}p{0.12\textwidth}p{0.05\textwidth}}
Paper&Year&Static/ Dynamic&Load balancing&Cloud Infrastructure&Number of robots&Parameters&Main objectives&Approach Used&Restrictions&Problems&Type of Experiments\\
\hline
\cite{Lippi:2021}&2021&Dyanamic&Partially&No&Several&Makespan, execution time, quality indes, supervision, workload, re-allocation&Minimizing normalized makespan while maximizing process quality and minimizing agents workload&A mixed-integer linear programming where solutions are recursively updated after completion of each task&Solution for the case require multiple re-allocations is time-consuming process, apart from the parameters' update times when the number of tasks assigned to an agent is large&The threshold for the cost is not described. The method does not work for multi-robot tasks. It does not include the tasks that should be started and finished at the same time.&Simulation\\
\hline
\cite{Chen:2021}&2021&Dyanamic&No&Yes&Several&Execution time, latency, energy consumption, characteristics of tasks and robots, grouping of robots&Optimization of quality of service by optimizing latency, energy consumption and costs, taking into account the specifications of the architecture and tasks&Mixed-integer linear programming, where solutions are determined using a heuristic method&It must be determined in advance which task is to be performed by a single robot or a group of robots or moved to the cloud.&The developed model does not take into account the dependencies between latency, energy consumption and costs.&Simulation\\
\hline
\cite{Casini:2021}&2021&Dyanamic&Yes&Yes&Several&Execution time, latency, task partitioning and splitting, load balancing.&Minimizing latency, load balancing and maximizing the number of completed tasks taking into account the time windows of the tasks.&Optimization problem and estimated solution using the lower bound for latency and solving the dual problem&The performance of the processors always remains the same and there is no communication time between the processors&The processors may not always have the same performance if they perform a task at different times, and the communication delay and communication affect the latency. Splitting tasks also increases latency because it takes time to switch from one task to another&Simulation\\
\hline
\cite{Ours:2022net}&2022&Static&No&Yes&Several&Task dependencies, average execution time, communication time, memory usage&Simultaneously minimizing all the robots'memory usage and the total time to execute all algorithms&Combinatorial graph theory and multivariate combinatorial optimization, and the solution is achieved with algebraic norms and branch-and-bound algorithm.&Require complete information about the algorithms and the robotic network cloud system architecture.&Cannot be applied to the system when a robot fails and the architecture changes dynamically.&Simulation on real-world data\\
\hline
\cite{Bai:2022}&2022&Dynamic&No&No&Several&Travel distance, robots' carrier capacity&Minimizing the total travel time of all robots&Combinatorial optimization method solved using an auction-based algorithm given an initial feasible solution.&Disturbances in robots, energy consumption, and recharge time are not considered.&The number of robots is considered large to obtain an initial solution, while in a real scenario the number of robots is usually limited.&Simulation\\
\hline
\cite{Fang:2023}&2023&Dyanamic&Yes&No&Several&Robots capacity, resource usage, start and finish time of a task&Task scheduling for a robotic network with single task robots and multi-robot tasks&Optimizing several objectives such as completion time, makespan, robot costs and balancing the workloads using PSO method.&Robots should have stable performance. The search for feasible solutions for a large system is very time consuming and can lead to a local optimal solution &The correlation between the objectives is not considered. Makespan minimization is equivalent to balancing the loads.&Simulation\\
\hline
\cite{Ye:2023}&2023&Dynamic&No&No&Several&Completion time, travel time&Minimizing the completion time of welding a single workpiece at a single station, the time difference between the completion times of adjacent stations, and the path length of robots moving to the station.&Multi-objective optimization problem solved with an evolutionary algorithm that sorts the stations and robots based on the expected completion time and the paths to the stations&Robots do not collide with any obstacles and have constant velocities, after welding each workpiece the completed task is immediately removed and the next task can be replaced.&The dependencies between the objectives were not considered in the formulation. The formulations only contain the times for the completion of the individual workpieces. However, after the completion of each workpiece, the station needs a certain time to remove the completed task so that it is ready for the next task. Possible collisions between the robots and the need to redirect them as they move between stations are not considered in the formulation, which can affect the performance of the system.&Simulations and real-world\\
\hline
\cite{Yin:2024}&2024&Dyanamic&No&Yes&-&Task priority, energy consumption, transmission time, time delay&Minimizing latency and energy consumption.&Multi objective optimization solved with PSO and GSA method&Tasks are independent.&In the real world, most tasks are interdependent and the performance of one task may depend on the performance of several other tasks.&Simulation\\
\hline
\end{tabular}}
\end{center}
\end{table*}
\begin{table*}[b!]
\tiny
\caption{Papers on task allocation and scheduling in robotic network systems using an optimization approach for solution.}\label{tabo4}
\begin{center}
\resizebox{\textwidth}{!}{\begin{tabular}{p{0.01\textwidth}p{0.01\textwidth}p{0.03\textwidth}p{0.02\textwidth}p{0.02\textwidth}p{0.03\textwidth}p{0.1\textwidth}p{0.1\textwidth}p{0.12\textwidth}p{0.12\textwidth}p{0.12\textwidth}p{0.05\textwidth}}
Paper&Year&Static/ Dynamic&Load balancing&Cloud Infrastructure&Number of robots&Parameters&Main objectives&Approach Used&Restrictions&Problems&Type of Experiments\\
\hline
\cite{Yan:2024}&2024&Dyanamic&No&No&Several&Energy consumption, resource requirements of the target, UAV resource limitation, path planning, collision avoidance.&Minimizes the total flight distance of all UAVs and the maximum flight distance of each UAV&Combinatorial optimization problem solved with a modified genetic algorithm, avoiding the UAVs to be in an infinite waiting state by ordering the targets based on their respective required resources.&UAVs must have a fixed altitude and a constant speed&In some scenarios, a task cannot be completed before switching to another task. For example, if tasks have deadlines and, for a given number of UAVs, the total resources carried by all UAVs are less than the resources required to complete the tasks in some of the targets in the ordered target list, then the task will be incomplete in the next step, but the task will be ranked lower and its deadline may be exceeded before it is completed.&Simulation\\
\hline
\end{tabular}}
\end{center}
\end{table*}

\subsection*{Combinatorial}
All contributions that solve the problem using a combinatorial approach are summarized in Tables \ref{tabc1} and \ref{tabc2}.

\begin{table*}[b]
\tiny
\caption{Papers on task allocation and scheduling in robotic network systems using a combinatorial approach for solution.}\label{tabc1}
\begin{center}
\resizebox{\textwidth}{!}{\begin{tabular}{p{0.01\textwidth}p{0.01\textwidth}p{0.03\textwidth}p{0.02\textwidth}p{0.02\textwidth}p{0.03\textwidth}p{0.1\textwidth}p{0.1\textwidth}p{0.12\textwidth}p{0.12\textwidth}p{0.12\textwidth}p{0.05\textwidth}}
Paper&Year&Static/ Dynamic&Load balancing&Cloud Infrastructure&Number of robots&Parameters&Main objectives&Approach Used&Restrictions&Problems&Type of Experiments\\
\hline
\cite{Chopra:2017}&2017&Dynamic&No&No&Several&Total distance traveled, cost function&Minimizing the cost function&Extension of the Hungarian method \cite{Burkard:2012}&No fair comparison with state-of-the-art methods, No tie-breaker to find the most suitable match&Depending on the matching the convergence rate can be very slow. It requires a load balancing approach to avoid assigning most of the tasks to a single robot. Assumes all robots capable of performing all tasks.&Simulation and real-world\\
\hline
\cite{Fan:2019}&2019&Static&Partially&Yes&-&Makespan, resource utilization, task completion time&reducing makespan and maximizing resource utilization&Combinatorial graph theory and solution obtained by matching&The complexity of the algorithm increases exponentially by increasing the number of tasks&The algorithm does not consider the precedence order between tasks&Simulation\\
\hline
\cite{Chen:2019}&2019&Dynamic&No&No&Several&Search, rescue&Maximizing the number of rescued, minimizing the average waiting time, minimizing the total path cost&Combinatorial optimization method by clustering tasks, a proportional selection strategy to avoid a local optimum, and a market base approach to find a solution&Clustering and proportional selection depend on the initial metric&Small metric for clustering and/or proportional selection, or for large number of robots may result in removing some tasks because of their deadlines&Simulation\\
\hline
\cite{Lu:2019}&2019&Dynamic&Partially&No&-&Scheduling length, makespan, lookahead, task prioritizing&Minimizing the scheduling length&Combinatorial optimization method solved by introducing lookahead in task prioritization and processor selection&Compared only with PEFT and HEFT, which are simple models. Requires the comparison with IPEFT&Metrics are defined over each method measure different properties&Real-world\\
\hline
\cite{WANG:2020}&2020&Dynamic&No&No&Several&Travel time, completion time&Minimize the maximum travel times of collaborating robots&A matching problem in a combinatorial graph optimization&Solution space of matching is exponentially larger then perfect matching&Fails to schedule the model with uncertainties. Only a certain robots may be used to transfer tasks due to their proximity&Simulation\\
\hline
\cite{Orr:2020}&2020&Static&No&No&-&precedence relation and communication cost&Task duplication to minimize the overall completion time.&Combinatorial graph theory using branch-and-bound search to find an optimal solution.&DAGs with most tasks with out-degree of at least 2 are computationally expensive.&It does not identify exactly which tasks should be duplicated.&Real-world\\
\hline
\cite{Zheng:2020}&2020&Dynamic&No&Yes&-&DAG, exact execution time, availability, parallelism&Maximizing parallelism of ready tasks&Combinatorial graph theory&existence of multiple sink and source tasks. Finding the priorities of tasks can be time-consuming.&Experiments show the proposed method has an advantage over other methods only for $DAG$s with a very small number of nodes. Average makespan reductions are compared without describing the makespans.&Real-world\\
\hline
\cite{Yu:2020}&2020&Dynamic&Partially&Yes&-&Task clustering runtime, load balancing, dependency balancing&Balancing the queueing time for scheduling clusters and find the dependency correlation measure to find the similarities between tasks by their data dependencies.&Combinatorial graph theory using task clustering&Number of clusters is known and finding the maximum dependency correlations is the goal. Tasks with identical execution times are considered for the first cluster.&The conclusion that the number of clusters is independent of the graph is incorrect.&Simulation\\
\hline
\cite{Hari:2020}&2020&Dynamic&No&No&Several&Misson time, travel time, waiting time, processing time, scheduling constraints&minimizing the maximum mission time and satisfying the scheduling constraints for human operators.&Combinatorial graph theory with a greedy heuristic for solution&Requires minimization of $\Idle$ times. The ratio $\alpha$ is a relatively large value. Computation time of the proposed method is exponential in general.&The travel time should be modified as from one task to the next, determined by order of the tasks. Random assignment of available human operator to a robot may cause a robot to be $\Idle$ for a long time&Simulation\\
\hline
\cite{Ours:2020}&2021&Static&No&Yes&Single&task dependencies, average execution time, communication time, memory usage&Simultaneously minimizing the robot's memory usage and the total time to execute all algorithms&Combinatorial graph theory and solution is obtained by algebraic norms and using branch-and-bound algorithm&Requires complete information about the algorithms&Cannot be applied to the system with multiple robots&Simulation on real-world data\\
\hline
\cite{Malencia:2021}&2021&Dynamic&No&No&Several&Redundant assignment, task performance, task cost& Improving performace by a fair redundant assignment of agents to tasks&Combinatorial graph optimization with the near-optimal solution obtainged by relaxing some of the constraints, supermodularity, and applying a greedy algorithm.&The thresholds and the criterion to terminate the recursive thresholding algorithm are not described. In the experiments generate random bipartite graphs without considering that task nodes must be with degrees at least $1$.&Each time, feasible solutions replace the initial solution instead of being added. The bound of the relaxing parameter and the main objective must be independent as a necessary condition to easier obtan a solution.&Simulation and real-world\\
\hline
\end{tabular}}
\end{center}
\end{table*}
\begin{table*}[b]
\tiny
\caption{Papers on task allocation and scheduling in robotic network systems using a combinatorial approach for solution.}\label{tabc2}
\begin{center}
\resizebox{\textwidth}{!}{\begin{tabular}{p{0.01\textwidth}p{0.01\textwidth}p{0.03\textwidth}p{0.02\textwidth}p{0.02\textwidth}p{0.03\textwidth}p{0.1\textwidth}p{0.1\textwidth}p{0.12\textwidth}p{0.12\textwidth}p{0.12\textwidth}p{0.05\textwidth}}
Paper&Year&Static/ Dynamic&Load balancing&Cloud Infrastructure&Number of robots&Parameters&Main objectives&Approach Used&Restrictions&Problems&Type of Experiments\\
\hline
\cite{Sahni:2021}&2021&Dynamic&No&No&Several&task dependencies, task scheduling, collaborative edge, network flow dependencies, task completion time&Minimizing task completion time taking into account dependencies between subtasks and schedules network flows&A combinatorial optimization problem with solution by heuristic search&It does not use the full capacity of the bandwidth for data transmission.&Several notions are not defined. The metric describing priorities of the subtasks is not well-defined. Experiments superflous as by construction of the proposed method has better performance than the methods LE, RE, SOFS, and ALT.&Simulation on real-world data\\
\hline
\cite{Minjia:2021}&2021&Static&Yes&Yes&-&Makespan, task arrival, communication time, processing time, bandwidth&Minimizing completion time with load balancing&Combinatorial graph theory approach with clustering methd for solution&Optimal performance is when the DAG can be splited onceinto DAGs with at most a single edge connecting each class to another.&For serial or parallel tasks, the scheduling time is higher than the usual scheduling methods because of clustering time. Requires comparison with state-of-the-art methods.&-\\
\hline
\cite{Fusaro:2021}&2021&Dyanamic&No&No&Several&Cost, task dependencies, agent capability, human-robot team&Minimizing the cost&Combinatorial graph theory with clustering to obtain allocation solution and a mixed-integer linear programming to minimize the cost.&Actual cost by human should be known. The solution space grows exponentially.&Requires monitoring the human agents. Human and robot are not distinguished. There are no comparisons with state-of-the-art methods.&Simulation\\
\hline
\cite{Pu:2021}&2021&Static&Yes&Yes&-&Budget, communication time, execution time, task duplication&Minimizing the completion time of the tasks taking into accound the price of using cloud&Combinatorial optimization with solutions by duplicating task&The duplication is regardless of improving the performance. The budget must be large.&The sub-budget and budget have a different scales. The total weights is obtained by adding terms with different scales.&Simulation and real-world\\
\hline
\cite{Shafiq:2021}&2021&Dyanamic&Yes&Yes&-&Makespan, task arrival, deadline, and completion time, resource utilization&Minimizing makespan and maximizing resource utilization&Combinatorial optimization with a greedy heuristic for solution&Tasks with long deadline can be assigned to a same virtual machine. Optimal load balancing is only for similar tasks' lengths and short deadlines.&For large number of tasks the method has poor performance compared to algorithms with greedy natures.&Simulation\\
\hline
\cite{Bharti:2022}&2022&Dynamic&No&Yes&Several&Energy consumption, reliablity, memory usage, cpu load&Maximizing successful task executions by all nodes&Combinatorial optimization method by splitting the dataset with expected accuracy, called CoRoL&The method only works for offloading machine learning tasks.&It does not identify when a task should be offloaded to other nodes.&Simulation\\
\hline
\cite{Jin:2022}&2022&Dynamic&No&No&Several&Time delay, competition rate, communication time&Minimizing the competition rate and over time the maximum delay, the winners, and theoutputs are updated using communication links.&Recursive optimization problem by time, with fewer robots shut down at each time step&The value of $k$ is fixed and should be strictly decreasing by time. It is assumed that there is only one task and all the robots compete to complete this task. In the case where several identical robots compete for the same task and have the same winning rate at any time, there should be a tie-breaker.&The method does not work for the case where there are several tasks where the robots compete for at least one of them. Several other factors such as the energy consumption of the robots and the movement distance of all robots must also be minimized.&Simulation\\
\hline
\cite{Ours:2023}&2023&Static&No&Yes&Several&Average execution time, communication time, task duplication.&Minimizing the competition time of all tasks by duplicating task to other nodes.&Two approaches are used: A combinatorial graph-theoretic approach based on the precedence order of tasks that recursively determines which tasks should be duplicated and to which node of the architecture the duplicated task should be assigned, and the optimization-based approach, \cite{Ours:2022net}, with branch-and-bound solution that should be solved for each node of the architecture&The time complexity of the method is polynomial and takes time to find a solution.&Task duplication increases the memory utilization by all robots. If some of the duplicated tasks are assigned to the robots, they decrease the efficiency of the robots and increase their cost.&Simulation\\
\hline
\end{tabular}}
\end{center}
\end{table*}

\subsection*{Reinforcement Learning}
All contributions that solve the problem using a reinforcement learning approach are summarized in Table \ref{tabr1}.

\begin{table*}[h!]
\tiny
\caption{Papers on task allocation and scheduling in robotic network systems using a reinforcement learning approach for solution.}\label{tabr1}
\begin{center}
\resizebox{\textwidth}{!}{\begin{tabular}{p{0.01\textwidth}p{0.01\textwidth}p{0.03\textwidth}p{0.02\textwidth}p{0.02\textwidth}p{0.03\textwidth}p{0.1\textwidth}p{0.1\textwidth}p{0.12\textwidth}p{0.12\textwidth}p{0.12\textwidth}p{0.05\textwidth}}
Paper&Year&Static/ Dynamic&Load balancing&Cloud Infrastructure&Number of robots&Parameters&Main objectives&Approach Used&Restrictions&Problems&Type of Experiments\\
\hline
\cite{Bian:2019}&2019&Dynamic&Yes&No&-&Task resource demands, average task slowdown&Task scheduling while learning to shorten average task slowdown and maintaining multi-resource fairness.&Online task scheduling using reinforcement learning&Objectives are the goal of all load balancing methods. Experimental results show that the performance of the method is within the performances of random and greedy.&Reward formula ignores arrival time. For simulation, assumed $\beta=0$ contradicting initial assumption $\beta>0$. Considering Poisson distribution with the parameter $0.8$ means $0.0002$ is the probability of at least $6$ tasks arrives and with $5$ CPUs in fog avoids assigning more than two tasks to a CPU.&Simulation\\
\hline
\cite{Wang:2019}&2019&Dynamic&Yes&Yes&Several&makespan and balance the resource usage, service time&Minimizing service time, maintaining balance in terms of computing and resources.&Deep reinforement learning gaining knowledge from environment by adaptive policies&Only a single node hosts an application.&The available capacity for data transmission cannot be assured. 
&Simulation\\
\hline
\cite{mostafavi:2020}&2020&Dynamic&No&Yes&-&Response time and makespan, resource efficiency&Minimizing the makespan and response time and increasing resource efficiency&Reinforcement learning&Size of occupied buffer and total task length of virtual machines are assumed to be independent and the $Q$-values are estimated&$Q$-value function is not well-defined. Poor estimation of the $Q$-value when the number of iterations is smaller than the smallest buffer capacity&Simulation\\
\hline
\cite{Ding:2020}&2020&Dynamic&No&Yes&-&User costs, tasks deadline, server's cost per unit use&Optimal solution for distributing tasks across servers minimizing user cost.&Reinforcement learning with the reward function defined as the negative average cost of the users&Infinite solution space.&The constraints of the tasks and the available resources should be checked. Experimental results are compared with methods that are not designed to minimize the user cost.&Simulation\\
\hline
\cite{Liu:2023}&2023&Dyanamic&No&Yes&Several&Lifetime, specifications, quality grade, and reliability and performance of robots, total price and total completion time of robots&Maximizing total service quality and minimizing total service performance&decentralized scheduling based on dueling deep $Q$-networks&All robots are able to perform tasks.&The task assignment index is linear, more examples with different task length are needed to determine the statistical significance of the proposed method.&Simulation\\
\hline
\cite{Chen:2023}&2023&Dynamics&Yes&Yes&-&Dynamics of the environment, inter-correlation between the scheduled tasks, and the trade-off between resource utilization and load balancing&maximizing resource utilization, minimizing total time, and load balancing&deep reinforcement learning approach with maximum posteriori policy&For the defined policy to solve the scheduling problem, the solution is not optimized if the posterior distributions are randomly skewed to the left and right&Load balancing minimizes the makespan (the total time is minimized)&Simulation on a real-world data\\
\hline
\cite{Li:2023}&2023&Dyanamic&No&Yes&Several&Data transmission, energy consumption, environmental changes&Minimizing completion time and energy consumption&Deep reinforcement learning approach with proximal policy optimization that considers traffic volume and adapts to environmental changes&Data transmission should occur within each time frame&Due to vehicle movement, the optimal data transmission path may change in successive time frames, resulting in additional delays in the transmission of big data&Simulation and real-world\\
\hline
\cite{Cui:2023}&2023&Dyanamic&No&No&Several&Edge devices, collaborative robots, task completion rate, communication delay, task deadline&Maximizing task completion rate&Decentralized multi-agent method of deep reinforcement learning with a partially observable Markov decision policy to determine the assignment of users to groups of collaborative robots and edge devices given the task deadline.&Communication delays between edge devices and between users and edge devices are negligible&Tasks that should be completed at the same time are assigned to the same groups of robots. Additional optimization is required for assigning tasks to edge devices to minimize the additional communication delays between edge devices.&Simulation\\
\hline
\end{tabular}}
\end{center}
\end{table*}

\subsection*{Alternative approaches}

All contributions that solve the problem using approaches other than optimization, combinatorial, and reinforcement learning are summarized in Tables \ref{tabot1} and \ref{tabot2}.

\begin{table*}[b]
\tiny
\caption{Papers on task allocation and scheduling in robotic network systems using approaches other than optimization, combinatorial, and reinforcement learning for solution.}\label{tabot1}
\begin{center}
\resizebox{\textwidth}{!}{\begin{tabular}{p{0.01\textwidth}p{0.01\textwidth}p{0.03\textwidth}p{0.02\textwidth}p{0.02\textwidth}p{0.03\textwidth}p{0.1\textwidth}p{0.1\textwidth}p{0.12\textwidth}p{0.12\textwidth}p{0.12\textwidth}p{0.05\textwidth}}
Paper&Year&Static/ Dynamic&Load balancing&Cloud Infrastructure&Number of robots&Parameters&Main objectives&Approach Used&Restrictions&Problems&Type of Experiments\\
\hline
\cite{Zhang:2018}&2018&Dynamic&No&Yes&-&Task time window, task execution cost&Maintaining the deadline of a task and minimizing the total cost&Uses two stages for task assignment: classification using Bayes classifier and then finding a match between the tasks and the most appropriate virtual machine. Tasks are scheduled into the $\Idle$ time slot of the matching virtual machine&The method is greedy and does not consider task switching between virtual machines&The method is compared only with min-min and max-min&Simulation\\
\hline
\cite{Schillinger:2018}&2018&Dynamic&No&No&Several&Planning, cost and energy levels&Minimizing the cost depending on the energy level of the robots&Centralized dynamic task allocation based on modeling the problem as a non-deterministic finite automaton&No comparison with the state-of-the-art methods&Communication times are not included and does not take advantage of cloud for further cost reduction&Simulation\\
\hline
\cite{Gombolay:2018}&2018&Dynamic&Partially&No&Several&Distance, completion time&Minimizing makespan and/or energy consumption considering the latency and spatial proximity to tasks&An algorithm similar to the branch-and-bound searches for the first task assignment by recursively performs agent assignment and task sequencer&The optimallity of a solution cannot be guaranteed as it depends on the initial cutoff and timeout&It is not explained how to select the initial cutoff and timeout&Simulation and real-world\\
\hline
\cite{Arunarani:2019}&2019&Dynamic&-&Yes&-&Scheduling in cloud computing&Review papers within years 2005-2018 for scheduling in cloud computing &All articles with the word ''scheduling'' in the title or keyword, published within 2005-2018, from scientific journals including IEEE, Elsevier, Springer, and other international journals are considered&Reviewed papers on cloud compunig.&Application of scheduling method on robotic networks are not considered.&-\\
\hline
\cite{botta:2019}&2019&Dynamic&No&Yes&Several&Computation and storage&Minimize computational time and storage space&An architecture consisting of robots, fog, cloud, and dew&Additional layer is introduced to the architecture&Missing the description of constraints for assigning tasks to nodes, the complexity of the model increases&Real-world\\
\hline
\cite{kim:2019}&2019&Dynamic&Partially&No&-&Energy consumption, task classification, completion time, time window constraints, task migration, dynamic voltage-frequency scaling (DVFS)&Minimizing energy consumption&Classifying task as short and long and allocating short oness to slow and long ones to fast processors then balancing the loads.&Task arrival information should be known&Without task precedence order, load balancing and task stealing contradicting each other, task stealing can lead to dropping a task before its deadline.&Simulation\\
\hline
\cite{malik:2019}&2019&Dynamic&No&Yes&-&Data transfer, energy consumption, resource utilization&Minimize energy consumption for data transfer&Data transmission instead of internet is made through multiple volunteer devices&The volunteer devices have limited energy level&The cost of energy consumption is shared among all volunteers. Load balancing is required to avoid using only some of the volunteer devices.&Simulation\\
\hline
\cite{Rizk:2019}&2019&Dynamic&-&Partially&Several&Scheduling in multiagent systems&Review papers on task allocation and scheduling in multiagent systems&Papers that consider multiagent systems are considered&Reviewed papers on multiagent systems that focus mainly on task decomposition and the degree of human agent intervention.&Some areas such as the use of clouds, human-robot collaboration, and the use of machine learning techniques are not considered.&-\\
\hline
\cite{stylus:2020}&2020&Dynamic&No&No&Several&Distance, cost&Minimize the costs for path planning&The canonical form of an infinite word in a finite state automata&Finding the canonical form of the infinite word is not a simple task&Canonical forms are obtained by random sampling may generate different prefixes with the same cost.& Simulation\\
\hline
\cite{Ours:2020h}&2020&Dynamic&No&Yes&Several&Task dependencies, average execution time, communication time, communication instability, compatibility, and capability of robots, fog, and cloud&Minimizing execution time, communication time and considering the capabilities of processing units.&Geometrical approach. Parameters are measured for all processing units and translated into a hypervolume to be maximized&The numerical values relating robots and tasks for a given parameter is not simple.&May require discretization for continuous parameters which may reduce the accuracy&Simulation\\
\hline
\end{tabular}}
\end{center}
\end{table*}
\begin{table*}[b]
\tiny
\caption{Papers on task allocation and scheduling in robotic network systems using approaches other than optimization, combinatorial, and reinforcement learning for solution.}\label{tabot2}
\begin{center}
\resizebox{\textwidth}{!}{\begin{tabular}{p{0.01\textwidth}p{0.01\textwidth}p{0.03\textwidth}p{0.02\textwidth}p{0.02\textwidth}p{0.03\textwidth}p{0.1\textwidth}p{0.1\textwidth}p{0.12\textwidth}p{0.12\textwidth}p{0.12\textwidth}p{0.05\textwidth}}
Paper&Year&Static/ Dynamic&Load balancing&Cloud Infrastructure&Number of robots&Parameters&Main objectives&Approach Used&Restrictions&Problems&Type of Experiments\\
\hline
\cite{Pupa:2021a}&2021&Dynamic&No&No&Single&Task execution time, human task execution variability, job quality of the human&Minimizing the nominal execution time, makespan, and waiting time, and maximizes the job quality metric.&Dynamic scheduling with real-time human monitoring, then rescheduling to minimize makespan and delay.&Metric estimation is redundant. Human operator monitoring should also be applied when tasks are rescheduled.&Obtaining the job quality metric is not described properly. The dynamic scheduler is assumed to be completely and correctly determined by the human agent. Moving tasks to higher levels must maintain the two orders (levels and schedule) that cannot be easily considered&Real-world\\
\hline
\cite{Sa:2021}&2021&Dyanamic&No&No&Single&Trust in robots to perform a task, predicting human success or fail for performing a task, natural trust, artificial trust&Produce a value representing a human trust on a robot to perform a task or not.&Probabilistic approach with updating over time&The natural trust does not have a numerical value.&The probability of success is a fitted sigmoid two additional positive parameters not a probability function. Number of used parameters is large.&Simulation on a real-world data\\
\hline
\cite{Shafiq:2021}&2021&Dyanamic&Yes&Yes&-&Makespan, task arrival, deadline, and completion time, resource utilization&Minimizing makespan and maximizing resource utilization&Combinatorial optimization with a greedy heuristic for solution&Tasks with long deadline can be assigned to a same virtual machine. Optimal load balancing is only for similar tasks' lengths and short deadlines.&For large number of tasks the method has poor performance compared to algorithms with greedy natures.&Simulation\\
\hline
\cite{Bharti:2022}&2022&Dynamic&No&Yes&Several&Energy consumption, reliablity, memory usage, cpu load&Maximizing successful task executions by all nodes&Combinatorial optimization method by splitting the dataset with expected accuracy, called CoRoL&The method only works for offloading machine learning tasks.&It does not identify when a task should be offloaded to other nodes.&Simulation\\
\hline
\cite{Dawarka:2022}&2022&-&-&Yes&Several&Architectures of cloud robotics systems&Review papers on creating a cloud robotics system&Papers that consider cloud robotics are examined&Review papers on cloud robotics systems that focus mainly on the architecture of the system.&Various areas such as task allocation and scheduling and collaboration between humans and robots are not considered.&-\\ 
\hline
\cite{Ours:2022ral}&2022&Dynamic&No&No&Single&$\idle$ time of agents, completion time of the job, execution time of tasks, precedence order of tasks, compatibility of tasks&Minimizing the completion time of the job by minimizing both agents' $\idle$ times.&Dynamic scheduling by prioritizing tasks with real-time human monitoring and then rescheduling to minimize completion time and delay.&Cannot be applied to a system with multiple humans and/or robots.&Tasks may have time window constraints that should be considered as a factor in evaluating their priority value.&Simulation\\
\hline
\cite{Wang:2023}&2023&Dyanamic&Yes&Yes&Several&Bandwidth, communication delay, completion time&Design of architecture with balanced load and minimized communication delay&Deep neural network, expert experience and domain knowledge used to find the optimal architecture based on the data collected when the system deployed&Requires stable communication without device failure. Load balancing and delay minimization are considered separately.&Since communication delays are random and there is a possibility of equipment failure, and since this information cannot be collected during the deployment phase, the model may fail in the long run.&real-world\\
\hline
\cite{Hao:2023}&2023&Dyanamic&No&Yes&-&the battery volume, the daily energy requirement of the device to keep it running, the prediction of the collected energy in the future, the energy consumption when sending and receiving data, the computing frequency of the device, the number of tasks, the number of instructions of the tasks, their deadline, their execution time&Maximizing the use of green energy while maintaining the working state of all devices in the future days and minimizing the completion time of all tasks.&Transforming the problem into an optimization problem and solving it with a genetic algorithm&The number of tasks on each day is fixed and constant and there is no inaccuracy in the weather forecast.&Communication delays and device failures can lead to more energy being consumed than expected for the current day, depleting the battery and leaving insufficient energy for the coming days.&Simulation\\
\hline
\cite{Wu:2023}&2023&Dynamic&No&No&Several&Task requirements, user requirements, encryption, matching, security&The matching of users and tasks is done by translating their requirements without revealing the sensitive information&The group-theoretic approach is used by translating the task and user requirements into encrypted integer lists with the size of a large prime number $p$ and then matching is done by comparing the homomorphic image of the direct product of two groups.&The number of requirements requests should be large since $p$ is a large value and therefore a large bandwidth is required for data transmission when users send multiple specific requests that can only be executed by some of the machines.& Since the defined groups are cyclic with the order of the prime number $p$, decoding the information can therefore be done simply by replacing the feeds with two unequal elements, at least one of which becomes non-trivial and generates the entire groups.&Simulations\\
\hline
\end{tabular}}
\end{center}
\end{table*}

\end{document}